\documentclass  [10pt]{article}

\usepackage{Sty/fmt}
\usepackage{Sty/mcr}
\usepackage{Sty/pkg}
\usepackage{Sty/thmE}
\usepackage{Sty/algE}
\usepackage{arydshln}

\usepackage[top=3cm, bottom=3cm, left=4cm, right=4cm]{geometry}

\usepackage{Sty/matsui}
\usepackage{comment}
\usepackage{ulem}

\begin{document}


\title{Adaptive Defective Area Identification in Material Surface Using Active Transfer Learning-based \\ Level Set Estimation}
\date{}

\author{
Shota Hozumi$^{1, \dagger}$, \and
Kentaro Kutsukake$^{2, 3, \dagger}$, \and
Kota Matsui$^{4, *}$, \and
Syunya Kusakawa$^{1}$, \and
Toru Ujihara$^{3, 5}$, \and
Ichiro Takeuchi$^{6, 2, *}$
}

\maketitle



\vspace{12pt}

\noindent
$^{1}$~Department of Computer Science, Nagoya Institute of Technology, Gokiso, Showa-ku, Nagoya 466-8555, Japan

\noindent
$^{2}$~RIKEN Center for Advanced Intelligence Project, Nihonbashi 1-chome, 1-4-1 Nihonbashi, Chuo-ku, Tokyo, 103-0027, Japan

\noindent
$^{3}$~Institute of Materials and Systems for Sustainability, Nagoya University, Furo-cho, Chikusa-ku, Nagoya 464-8603, Japan.

\noindent
$^{4}$~Department of Biostatistics, Nagoya University, 65 Tsurumai-cho, Showa-ku, Nagoya 466-8550, Japan

\noindent
$^{5}$~Department of Materials Process Engineering, Nagoya University, Furo-cho, Chikusa-ku, Nagoya 464-8603, Japan

\noindent
$^{6}$~Department of Mechanical Systems Engineering, Nagoya University, Furo-cho, Chikusa-ku, Nagoya 464-8603, Japan

\noindent
$^{\dagger}$~These authors contributed equally to this study. 

\vspace{12pt}

\noindent
Correspondence and requests for material should be addressed to K.M. (E-mail: matsui.k@med.nagoya-u.ac.jp) and I.T. (E-mail:  ichiro.takeuchi@mae.nagoya-u.ac.jp).

\clearpage

\begin{abstract}
 %
In material characterization, identifying defective areas on a material surface is fundamental.
The conventional approach involves measuring the relevant physical properties point-by-point at the predetermined mesh grid points on the surface and determining the area at which the property does not reach the desired level.
To identify defective areas more efficiently, we propose adaptive mapping methods in which measurement resources are used preferentially to detect the boundaries of defective areas.
We interpret this problem as an active-learning (AL) of the level set estimation (LSE) problem.
The goal of AL-based LSE is to determine the level set of the physical property function defined on the surface with as small number of measurements as possible. 
Furthermore, to handle the situations in which materials with similar specifications are repeatedly produced, we introduce a transfer learning approach so that the information of previously produced materials can be effectively utilized. 
As a proof-of-concept, we applied the proposed methods to the red-zone estimation problem of silicon wafers and demonstrated that we could identify the defective areas with significantly lower measurement costs than those of conventional methods.
(179 words)

 \vspace{.1in}
 \begin{center}
 {\bf Keywords}
 \end{center}
Material Characterization, Spatial Mapping, Active Learning, Bayesian Modeling, Level Set Estimation, Transfer Learning
 \vspace{.1in}
\end{abstract}

\clearpage
\section*{Introduction}
\label{sec:introduction}

\noindent
Evaluating the spatial distribution of the relevant physical properties of a material surface is fundamental~\cite{balke2010nanoscale, newbury2013elemental, salzer2014infrared, shapiro2014chemical}.
In the manufacturing industry, identifying defective areas on a material surface where the physical properties do not satisfy the desired level is particularly important.
For example, in the manufacturing of semiconductor materials, the defective areas must be identified on a brick surface where the carrier lifetime is shorter than the desired level~\cite{zhong2014influencing, schroder2015semiconductor, vahanissi2017full}.
In conventional defective area identification problems, it is common to measure the physical properties point-by-point at predetermined mesh grid points evenly allocated to the material surface.
Unfortunately, when the cost of physical property measurements is expensive, this conventional approach is inefficient.
To improve the efficiency, It is reasonable to place fewer measurement points in areas that can be identified as clearly normal or defective, while placing more measurement points near the boundary between the normal and defective areas.
However, such an adaptive approach is impossible before the physical property distribution of the material surface is determined. 
Thus, it is desirable to develop a method that enables the adaptive selection of measurement points while estimating the physical property distribution.

In this study, we interpret the defective area identification problem as a \emph{Level Set Estimation (LSE)} problem~\cite{bryan2006active,gotovos2013active,zanette2018robust,shekhar2019multiscale,iwazaki2020bayesian,inatsu2020activeB}, in which we regard the physical property distribution on a material surface as a real-valued function defined on a two-dimensional input space and estimate the level set --- the  set of input spaces where the function takes on a given constant value. 
When defective areas are defined as points on the material surface at which the physical property is smaller than a certain threshold $\theta$, the level set at $\theta$ can be interpreted as the boundary between the normal and defective areas.
Here, we introduce \emph{active learning (AL)-based LSE} methods that are designed to estimate the level set of a function with as small number of function evaluations as possible by adaptively selecting the points to be evaluated. 
AL is a special case of supervised machine learning (ML) in which a learning algorithm can adaptively select the input points of the training set~\cite{settles2009active,settles2011theories,snoek2012practical,shahriari2015taking,inatsu2020activeA}, and it is also called the optimal experimental design in the statistics literature~\cite{chaloner1995bayesian,queen2002experimental,pukelsheim2006optimal,kirk2012experimental}.
The AL-based LSE methods can adaptively select the measurement points on the material surface while effectively searching for the boundaries of the normal and defective areas. 

When materials with the same specifications are repeatedly produced, the normal and defective areas of a newly produced material can be estimated by referring to the spatial distributions of the physical properties of the previously produced materials.
In ML, an approach for efficiently solving new problems using solutions to past similar problems in the past is called \emph{transfer learning (TL)}~\cite{pan2009survey}.
In this study, we propose a TL approach for the LSE problem called \emph{active transfer learning (ATL)-based LSE}.
By effectively transferring knowledge of previously produced materials, ATL-based LSE methods can identify defective areas with fewer measurements than those without knowledge transfer. 
In manufacturing, location and scale displacements can easily occur owing to environmental and operational variabilities.
In such cases, conventional TL methods are insufficient because they simply assume that the two functions of the source (old) problem and the target (new) problems are similar. 
To address this problem, we propose a variant of the ATL-based LSE method that can be applied even with location and/or scale shifts.
This variant leverages the similarity between the source and target problems after changes in the location and/or scale.

This study is motivated by the \emph{red-zone} estimation problem in solar cell ingots~\cite{schroder2015semiconductor, vahanissi2017full, kutsukake2015characterization,zhang2016growth,hu2017grain,li2017minority}.
One measure of the quality of silicon ingots, which are the base material for solar cells, is the value of the carrier lifetime.
The carrier lifetime is the time required for carriers stimulated by light to return to the ground state and is positively correlated with the conversion efficiency of the solar cell.
In actual ingot production, impurities enter the outer part of the material that touches the crucible during production, resulting in a low lifetime value.
The area of low lifetime values within a material surface is referred to as the red-zone.
Thus, It is important to identify the red-zone (defective area) on a silicon brick surface before sending the material to the next manufacturing stage. 
As a proof-of-concept, we demonstrate the effectiveness of the proposed AL-based and ATL-based LSE methods by applying them to red-zone estimation problems for solar cell ingots.
\clearpage
\section*{Results}
\label{sec:sec2}

\paragraph{Red-zone estimation in silicon ingots for solar cells}
This study is motivated by the red-zone estimation problem in multicrystalline silicon ingots for solar cells, which are commonly used material for solar cell substrates (Figure~1a).
The periphery of the ingots is contaminated by impurity diffusion from the crucible walls and cannot be used for solar cells.
Such unusable areas are called \emph{red-zone}, and their distribution and volume are important parameters of the material yield of the ingots~\cite{kutsukake2015characterization,zhang2016growth,hu2017grain,li2017minority}.
Conventionally, the red-zone is defined as the area where the carrier lifetime is lower than a threshold value and is identified by their spatial mapping measurements.
The carrier lifetime is a basic physical property of semiconductor materials and is positively correlated with solar cell performance.
In most cases, mapping measurements are performed point-by-point along the mesh grid coordinates with equal spaces (Figure~1b).
However, such mesh grid mapping is time consuming, and the cost is a bottle neck in material development.
In this study, AL-based LSE methods are used to reduce the number of measurement points in the carrier lifetime (Figure~1d). 
As with the cases of general industrial production, the spatial distributions of the red-zone are similar among different silicon ingots; however, the location and scale of the carrier lifetime values are often shifted in each product.
To treat these features, we apply ATL-based LSE methods to this problem (Figure~1e). 
We consider the problem of estimating red-zone of a surface in a the brick of a silicon ingot with 6,586 mesh grid points.
As a proof-of-concept, the lifetime values at all grid points are measured in advance, and the performances of the LSE methods are evaluated by checking whether the methods can correctly identify the ground-truth red-zone only with lifetime values measured at a small number of selected points. 

\paragraph{Problem Setup: Level Set Estimation (LSE)}
We interpret the problem of identifying defective area, such as red-zone in solar cell ingots, as an LSE problem. 
Let $f(x)$ represent the relevant physical property, such as  the carrier lifetime, at position $x$ in a two-dimensional material surface. 
If the physical property is desired to be no smaller than a certain threshold $\theta$, the defective area identification problem is formulated as a problem of identifying the super-level set $\cU_\theta = \{x \in \cX \mid f(x) \ge \theta\}$ and the sub-level set $\cL_\theta = \{x \in \cX \mid f(x) < \theta\}$, where $\cX$ is the entire input domain, i.e., the entire material surface. 
Assuming that the physical property changes smoothly in the two-dimensional material surface, the super- and sub-level sets at $\theta$ of the function $f$ can be estimated by measuring the physical property at sufficiently fine mesh grid points $x_1, \ldots, x_N \in \cX$, where $N$ is the number of grid points. 
However, if the measurement cost is expensive, the total cost to colect the observations at all $N$ grid points can be unacceptably high. 
Our basic idea involves reducing the total measurement cost by evaluating function $f$ only at a specific portion of the grid points at which the function evaluation is effective for estimating the super- and sub-level sets at $\theta$.
That is, measuring the physical properties at which they are clearly higher or lower than the threshold $\theta$ is inefficient, whereas effectively using measurement resources at which the physical properties are near $\theta$ is reasonable.
Based on this idea, we propose an active learning-based (AL-based) methods for the defective area identification problem, which aims to estimate the level set at $\theta$ with as few function evaluations as possible.
In the following sections, we present AL-based and active transfer learning-based (ATL-based) methods for effectively solving the LSE problem by adaptively selecting measurement points.
Figure~2 shows an illustratinve example of an LSE problem. 

\paragraph{Active learning (AL)-based LSE method}

AL is a type of supervised machine learning method in which the training input points can be selected so that the function can be estimated with as few function evaluations as possible. 
There are several AL-based methods for LSE problem~\cite{bryan2006active,gotovos2013active,zanette2018robust,shekhar2019multiscale,iwazaki2020bayesian,inatsu2020activeB}.
%
The basic idea behind AL-based LSE methods is to employ a Bayesian regression model to approximate the physical property function $f$. 
AL-based LSE methods iteratively update the Bayesian regression model by adding the selected measurement data into the training set. 
%
In this study, we employed the \emph{Gaussian Process (GP)}  model~\cite{williams2006gaussian} as a Bayesian regression model. 
Suppose we are now at step $t$ at which a small subset of grid points $x_i, i \in \cS_t \subset \{1, \ldots, N\}$ of the physical properties, have been measured, where $\cS_t \subset \{1, \ldots, N\}$ is a subset of grid point indices. 
Then the Bayesian regression model is fitted using the set of labeled instances $\{(x_i, f(x_i))\}_{i \in \cS_t}$. 
The advantage of the Bayesian regression model is that it provides not only the predictive mean but also the predictive variance at each grid point. The predictive variance indicates the uncertainty of the physical property at each grid point. 
Based on the mean and the variance of the function values at the grid points, a new point $x_{i(t)}, t \notin \cS_t$, which is likely to be effective in estimating the super- and sub-level sets at $\theta$, is selected, and the physical property $f(x_{i(t)})$ at the selected position $x_{i(t)}$ is measured. 
Then, in the next step, $t+1$, the subset is updated as $\cS_{t+1} \lA \cS_t \cup \{i(t)\}$, and the same process is conducted. 
These steps are repeated until the super- and sub-level sets at $\theta$ are estimated sufficiently well.
Figure~3 illustrates the AL-based LSE method for a one-dimensional function.

Figure~4{\bf a}  is a snapshot of a certain step of the AL-based LSE method.
The key idea behind the AL-based LSE methods is to utilize the \emph{credible interval} of the physical property at each grid point; if the lower end of the credible interval is greater than the threshold $\theta$, the physical property is highly likely to be greater than $\theta$, whereas if the upper end of the credible interval is smaller than the threshold $\theta$, the physical property is highly likely to be smaller than $\theta$. 
That is, we should effectively use our measurement resources for the grid points at which the credible intervals contain the threshold $\theta$ because this suggests that whether the physical property is greater or smaller than the threshold remains unclear.
Specifically, the next evaluation point is selected based on the \emph{violation} defined as $\alpha(x) = \min\{\max\{0, \theta - \ell(x)\}, \max\{0, u(x) - \theta\}\}$.
Figure~4{\bf b} illustrates the violations for several input candidates (see Methods and Supplementary Information 1-3s for details).

\paragraph{Evaluation Criteria}
Based on the credible interval,  each unmeasured point $x_i, i \notin \cS_t$ can be classified into the following three classes: 
\begin{itemize}
 \item $f(x_i) > \theta$ (if the lower end is greater than $\theta$);
 \item $f(x_i) < \theta$ (if the upper end is smaller than $\theta$);
 \item Undetermined (otherwise). 
\end{itemize}
This means that the goodness of the AL-based LSE method at each step is evaluated based on the counts of each cell in the following table :
\begin{center}
\begin{tabular}{c|c||c|c|c}
 \multicolumn{2}{c||}{} & \multicolumn{3}{c}{Prediction by the AL-based LSE method} \\ \cline{3-5}
 \multicolumn{2}{c||}{} & $\hat{f}(x_i) > \theta$ & Undetermined & $\hat{f}(x_i) < \theta$ \\ \hline \hline
 \lw{Ground-Truth} & $f(x_i) > \theta$ & TP & UP& FN\\ \cline{2-5}
 & $f(x_i) < \theta$ & FP & UN& TN\\ 
\end{tabular}
\end{center}
where TP, FN, FP, and TN respectively indicate true positive, false negative, false positive, and true negative, respectively, and $\hat{f}$ denotes the predictions based on the Gaussian process model.
In addition, we call the cases in which the output of the AL-based LSE method is ``undetermined'' as undetermined positive (UP) and undetermined negative (UN), respectively.
The treatment of these undetermined points depends on the problem setup. 
If removing the defective area is much more important than maintaining the normal area, undetermined points should be treated as (possibly) defective areas.
In this case, all undetermined points are treated as elements of the sub-level set $\mathcal{L}_{\theta}$. 
In particular, if $\x$ is an UP instance, $\x$ should be treated as an FN, and we use UP + FN as a \emph{false negatives} in the computation of sensitivity and specificity. 
Similarly, if $\x$ is an UN instance, $\x$ should be treated as a TN, and we use UN + TN as the \emph{true negatives} in the computation of sensitivity and specificity. 
This leads to the following ``risk-sensitive'' sensitivity and specificity: 
\begin{align*} 
\text{sensitivity}_{\text{risk-sensitive}} &= \frac{\rm TP}{\rm TP + UP + FN}, \\
\text{specificity}_{\text{risk-sensitive}} &= \frac{\rm TN + UN}{\rm TN + FP + UN}. 
\end{align*}
On the other hand, if maintaining the normal area is much more important than removing defective areas, undetermined points should be treated as (possibly) normal areas.
In this case, all undetermined points are treated as elements of the super-level set $\mathcal{U}_{\theta}$.  
In particular, if $\x$ is an UP instance, $\x$ should be treated as a TP, and we use UP + TP as the \emph{true positives} in the computation of sensitivity and specificity. 
If $\x$ is an UN instance, $\x$ should be treated as an FP, and we use UN + FP as the \emph{false positives} in the computation of sensitivity and specificity.  
This leads to ``cost-sensitive'' sensitivity and specificity measures as follows: 
\begin{align*} 
\text{sensitivity}_{\text{cost-sensitive}} &= \frac{\rm TP + UP}{\rm TP + UP + FN}, \\
\text{specificity}_{\text{cost-sensitive}} &= \frac{\rm TN}{\rm TN + FP + UN}. 
\end{align*}
Using these two types of metrics, we similarly define two types of area under the curves (AUCs) and F$_1$ scores, which are standard indices to evaluate the model performance~\cite{duda2012pattern}.

\paragraph{Results of AL-based LSE method}

We applied the AL-LSE method to red-zone estimation problems in solar cell ingots. 
Here, 74 $\times$ 89 grid points that were evenly allocated to an ingot surface of size 186 mm $\times$ 156 mm  were considered. 
We compared the proposed methods with two baseline methods. 
The first baseline method is RANDOM, in which the next measurement point is selected uniformly at random.
The second baseline method is NON-ADAPTIVE, which is a non-adaptive method and is illustrated in Supplementary Figure 1 (see Methods for the details of the experiment setups). 
Figures~5{\bf a}, {\bf b}, and {\bf c} show the behaviors of the RANDOM, NON-ADAPTIVE, and AL-based LSE methods, respectively (Supplementary Movies 1-3 show how each method selects measurement points at each step). 
In these plots, the blue, red, and gray regions indicate the estimated super-level set, sub-level set, and undetermined region, respectively. 
Figure~6 shows the risk-sensitive AUCs, risk-sensitive F$_1$ scores, cost-sensitive AUCs, and cost-sensitive F$_1$ scores for the RANDOM (red), NON-ADAPTIVE (purple), and AL-based LSE (blue) methods for solar cell ingots data along with methods using transfer learning (TL), which will is described below.

By comparing the plots in {\bf a}, {\bf b}, and {\bf c}, in the early stages (see left column), no significant differences are evident between the three methods; however, as the steps progress (see middle column), the AL-based LSE method has more measurement points at both the left and right ends, which is the true defective area.
Furthermore, in the final stage (see right column), the AL-based LSE method places many measurement points near the boundary between the normal  (blue) and defective (red) areas.
The AL-based LSE method determines almost all areas to be normal (blue) or defective (red), whereas the RANDOM and NON-ADAPTIVE still have some gray areas even in the final stage.
Figure~6 shows the clear advantages of the AL approach over the two baselines.
Note that the performance measures in the risk-and the cost-sensitive scenarios significantly vary in the two baseline methods because undetermined areas remain; areas that cannot be determined as either normal or defective.

\paragraph{Active Transfer Learning (ATL)}
To identify the red-zone of a newly produced material more efficiently, it is beneficial to utilize the measurement data of previously produced materials. 
We implemented this idea using a transfer learning (TL) framework, where the old and new problems are referred to as \emph{source} and \emph{target} domains, respectively.
TL covers a wide class of methods that utilize additional information in various ML tasks, including \emph{feature transfer} and \emph{model transfer} approaches.
In this study, we employ the idea of \emph{instance transfer}~\cite{pan2009survey} approach, in which data from the previously conducted experiments (source domain) are effectively utilized in the current experiment (target domain). 
Various methods, such as importance weighting~\cite{sugiyama2012density} and optimal transport~\cite{7586038}, have been proposed for instance transfer, and we employed the \emph{Diff-GP} approach~\cite{shilton2017regret} because it allows the direct use of the GP models we discussed above.
In Diff-GP, the difference between the target and source functions is modeled using a GP regression model. 
Since the source and the target problems are assumed to be similar, the difference function can be estimated with high accuracy using few training instances. 
Let $\{(x_j^\prime, y_j^\prime)\}_{j=1}^m$ be the training instances for the source problem.
Then, $\{(x_j^\prime, y_j^\prime + \hat{f}_{\rm diff}(x_j^\prime))\}_{j=1}^m$ can be used as an additional training instances for the target problem, where $\hat{f}_{\rm diff}$ is the estimated mean difference function.
In the proposed ATL-based LSE method, measurement points are selected by incorporating the uncertainty of the difference function. 
Figure~7{\bf a-c} illustrates the basic concept of Diff-GP. 

Although the spatial distributions of physical properties are typically similar among different products, their location and scale are often shifted in each product owing to the fluctuations in the measurement conditions and material qualities.
To address this problem, we also developed a variant of the ATL-based LSE method that can be applied even with shifted locations and/or scales.
This variant leverages the similarity between the source and target domains after the location and/or scale changes.
When the source domain is similar to the target domain after a location-scale shift, $\{(x^\prime_j, \gamma (y^\prime_j + f^\prime(x_j)) + \eta)\}_{j=1}^m$ can be used as additional training instances for the target domain, where $\eta$ and $\gamma$ are the location and scale shift parameters, respectively. 
In this variant of the ATL-based LSE method, the measurement points (for the target domain) are selected so that the difference function and the location-scale shift parameters can be estimated simultaneously. 
Figures~7{\bf d-f} illustrates the basic concept of location-scale shifts. 

Figures~5 and 6 show the results of the ATL-based LSE, and location-scale shift (LSS)-ATL-based LSE methods. 
In Figure~5, compared with the AL-based LSE method, the ATL-based and LSS-ATL-based LSE methods can identify many areas as normal (blue) or defective (red) at the early stage (see the left column).
This is because of the effective use of previously produced material information through TL.
On the other hand, the performances of the ATL-based and LSS-ATL-based LSE methods were insignificantly different in the present experiment.
This is because the present data are obtained from a well-controlled experiment without significant environmental or operational variants between the source and the target silicon ingots.
In addition, Figure~6 suggests a clear advantage of using the transferred knowledge in all four performance measures. 
These results suggest that it is effective to incorporate the knowledge of previously produced materials into the framework of transfer learning. 
\clearpage
\section*{Discussions}
In this study, we show that adaptively selecting measurement points is effective for identifying defective areas on material surfaces with a much lower measurement cost compared with the conventional approaches with predetermined measurement points.
As illustrated in Figure~5, this problem can be interpreted as a problem of dividing the material surface into normal and defective areas.
This means that the measurement points should be selected such that an undetermined region (the gray area in Figure~5), which cannot yet be determined as normal or defective, can be reduced.
The main contribution of this paper is to show that this problem can be formulated as an LSE problem and that the AL approach allows for an adaptive selection of effective measurement points for the LSE problem.
Indeed, in the red-zone identification problem in silicon ingots, we could identify the defective area at a significantly lower cost compared with conventional methods.
Furthermore, we demonstrate that a framework that utilizes information from previously produced materials with similar specifications can be viewed as a TL problem and introduce the ATL- and the LSS-ATL-based LSE methods.
In the manufacturing industry, using \emph{sampling inspection} is common when the measurement cost for material characterization is high. 
Using the proposed active transfer learning approaches, performing a complete inspection may be possible by reducing the examination cost of each material.

Here, as a proof-of-concept study on silicon ingots, AL- and ATL-based LSE methods significantly reduced the number of measurement points of the carrier lifetime mapping on the sample surface. 
This reduction is because of the features of the carrier lifetime of the silicon bricks, which are continuously distributed and similar among products.
These are the common features of the physical properties of industrially produced materials.
Thus, AL- and ATL-based LSE methods will be useful for evaluating the spatial distribution of various physical properties of various materials.
In contrast, this advantage suggests that these methods are difficult to apply to materials with discontinuous and specific distribution of physical properties.
For multicrystalline silicon ingots, the identification of dislocation clusters is such an issue.
In addition to the red-zone, low-lifetime regions caused by dislocation clusters are defective areas in multicrystalline silicon ingots, as shown in Figure~1{\bf c}.
Dislocation clusters are generated by the stress inside the ingots during the solidification process, and their locations vary for each ingot.
Therefore, if the information on the dislocation cluster positions of the previously produced ingots is transferred to the measurement of newly produced ingots, the performance of the ATL-based LSE might may decrease.

To cope with the physical properties that discontinuously change in material surfaces, it is effective to employ appropriate kernel functions such as Matern's kernel in the GP model. Matern's kernel has been used in the context of spatial statistics, geostatistics, and image analysis since it can capture discontinuous changes~\cite{genton2001classes,williams2006gaussian}. 
Generally, it is important to choose an appropriate kernel and hyperparameters based on he nature of the physical properties. 
Although we selected  hyperparameters by cross-validation when information on previously produced materials is available, it is also effective to transfer information on hyperparameters~\cite{jalem2018bayesian}.
In TL, a case in which transferring the source domain information may become an inappropriate bias to the target domain, which is referred to as \emph{negative transfer}~\cite{pan2009survey}.
When using information from previously produced materials, we need to carefully determine whether the negative transfer can occur.  
To the best of our knowledge, no method has been established for predicting negative transfers in AL settings. 
However, in practice, we can introduce a heuristic mechanism to make adaptive decisions, such as not using TL when the function values in the target domain significantly differ from those predicted by the transferred information. 
Nevertheless, many engineering problems involve materials with similar specifications being repeatedly produced. 
In such cases, ATL-based approaches are effective, as demonstrated in this study. 

Another issue in the physical property mapping of the material surface is moving the distance of the measurement probe.
As illustrated in Figure~1{\bf b}, in the conventional mapping on the mesh grid, physical property measurements are conducted with the minimum moving distance between the points. 
In contrast, in the proposed AL- and ATL-based methods, the next measurement point is selected from the entire surface area.
Consequently, the total moving distance of the probe becomes longer than that of conventional methods.
In some cases, the total measurement time increases because of the probe movement time, although the number of measurement points decreases.
This is a serious problem in the mapping equipment, which requires a certain amount of time for the probe movement.
To incorporate the cost of the probe movement into the adaptive selection of measurement locations, we need to introduce a new AL mechanism.
Many AL methods, including those we discussed in this paper, are called \emph{myopic approach} in the sense that the choice of the next training instances in each step does not incorporate the effect on subsequent steps.
To overcome this challenge, we will need to introduce the idea of non-myopic AL~\cite{jiang2017efficient} or \emph{reinforcement learning}~\cite{kaelbling1996reinforcement,sutton1998introduction}, which has been actively studied for optimal control under uncertain environments. 
\clearpage
\section*{Methods}
\label{sec:sec4}

\paragraph{Carrier lifetime data}
The carrier lifetime data used in this study were obtained from a size G2 semi-industrial multicrystalline silicon ingot (400 $\times$ 400 $\times$ 200 mm$^3$ and 70 kg).
The side surface area of each brick was 156 $\times$ 186 mm$^2$, and the mesh grid points arranged at intervals of 2 mm were considered, resulting in 6,586 grid points.
The grid points with lifetime values smaller than the threshold $\theta=2.0$ were defined as red-zone.
In order to check whether the LSE methods can correctly identify red-zones, the lifetime values of all grid points were measured in advance.
The microwave photoconductivity decay measurements were performed on the as-sliced surfaces of silicon bricks without passivation.
The location-scale shift of the carrier lifetime between ingots originates from the condition of the surface and material quality.
In the experiments of the LSE methods, the inspection points were selected from these grid points by hypothetically assuming that their lifetime values were unknown, and the performances were measured by how many of these grid points were correctly classified by the LSE methods as red-zone or normal.

\paragraph{Problem Formulation}
Suppose a black-box function $f: \mathcal{X} \rightarrow \Rbb$ and candidate points $x_1, ..., x_N \in \mathcal{X}$ are given.
Here, $\mathcal{X}$ denotes a set of input positions on the material surface, and the output of the function $f$ is a real-valued relevant physical property.
Furthermore, observation $y$ of the function value $f(x)$ for input $x$ is expressed as $y = f(x) + s$ with a Gaussian error of mean $0$ and variance $\sigma^2$, $s \sim \mathcal{N}(0, \sigma^2)$.
The LSE~\cite{gotovos2013active} problem is formulated as a problem of classifying each input point $x_i$ into the following two areas
\begin{align}
    \label{eq:LSE}
    \mathcal{U}_{\theta} = \{x \in \mathcal{X} \mid f(x) \ge \theta \}~~~\mbox{or}~~~
    \mathcal{L}_{\theta} =\{x \in \mathcal{X} \mid f(x) < \theta \}
\end{align}
for a certain threshold $\theta$. 
Then the defective area identification problem can be considered an LSE problem with a real-valued black-box function $f$ with two-dimensional inputs.
%
%
In ordinary regression analysis, executing (1) is possible for a new input by estimating $f$ using multiple pairs of input and output $\mathcal{D}_N = \{(x_i, y_i)\}_{i=1}^N$ we have already obtained as training data .
However, in many scientific fields, such as material science, observing the output $y$ for input $x$ can be quite expensive. 
Hence, it is often required to execute (1) with few observations of $y$.
To this end, we employed the AL approach with the GP modeling~\cite{williams2006gaussian} for $f$. 

\paragraph{Modeling and Algorithm}
The zero-mean GP model $\mathcal{GP}(0, k(x, x^{\prime}))$ of $f$ is characterized by a kernel function $k(x, x^{\prime})$. 
The kernel function incorporates prior knowledge of the smoothness of $f$ into the model and evaluates the similarity between inputs. 
In this study, we used the radial basis function (RBF) kernel, defined as follows:
\begin{align}
    \label{eq:rbf_kernel}
    k(x, x^{\prime}) = 
    v\exp \left(- \frac{\|x - x^{\prime}\|^2}{2 \ell^2} \right),
\end{align}
where $v$ and $\ell$ are hyperparameters that we have to appropriately tune. 
Suppose we are at step $t$ and the data $\mathcal{D}_{t} = \{(x_i, y_i)\}_{i=1}^{N_t}$ of the input and output pairs have been observed.
Then the predictive distribution of the objective value $f(x)$ at the new input $x$ under the observation of data $\mathcal{D}_N$ is a normal distribution $\mathcal{N}(\mu_t(x), \sigma^2_t(x))$ with predictive mean $\mu_t(x)$ and predictive variance $\sigma^2_t(x)$, defined as follows:
\begin{align}
    \label{eq:pred_dist}
    \mu_t(x) &= \k(x)^{\top}(\K + \sigma^2\I)^{-1}\y, \\ 
    \sigma^2_t(x) &= k(x, x) + \k(x)^{\top}(\K + \sigma^2\I)^{-1}\k(x),
\end{align}
where $\k(x) = (k(x, x_1), ..., k(x, x_{N_t}))$, $\K$ is a matrix of size $N_t \times N_t$ whose $(i, j)$-th element is defined by $k(x_i, x_j)$, $i, j = 1, ..., N_t$, and $\I$ is the identity matrix.
Based on this modeling, our AL algorithm for LSE is performed using the following procedure: 
(i) Define the confidence region $Q_t(x) = [\mu_t(x) \pm 1.96 \sigma_t(x)]$. 
(ii) Classify candidate point $x \in \mathcal{X}$ as $x \in \mathcal{U}_{\theta}$ if $\min Q_t(x) + \varepsilon = \mu_t(\x) - 1.96 \sigma_t(\x) + \varepsilon \ge \theta$ and $x \in \mathcal{L}_{\theta}$ if 
$\max Q_t(x) - \varepsilon =  \mu_t(\x) + 1.96 \sigma_t(\x) - \varepsilon < \theta$. 
(iii) Select the next point to be evaluated by maximizing the acquisition function (called the straddle)~\cite{bryan2006active} defined as follows:
\begin{align}
    \label{eq:uncertainty_sampling}
    x_{N_t + 1} = \argmax_{x \in \mathcal{X}} [\min\{\max Q_t(x) - \theta, \theta - \min Q_t(x))\}]
    = \argmax_{x \in \mathcal{X}} \underbrace{1.96\sigma_t(x) - |\mu_t(x) - \theta|}_{\mbox{straddle}}.
\end{align}
(iv) Update the GP model using the data $\mathcal{D}_{t+1} = \mathcal{D}_t \cup \{(x_{N_t+1}, y_{N_t+1})\}$. 
The details of the algorithm is shown in Supplementary Information 1. 
Figure~3 shows a demonstration of the AL for LSE with synthetic data. 

\paragraph{Active Learning-based LSE with Diff-GP Model}
Let us consider a situation in which, in addition to the data from the current experiment, the dataset $\mathcal{D}^{\prime} = \{(x_j^{\prime}, y_j^{\prime})\}_{j=1}^M$ is obtained from a similar experiment $f^{\prime}$.
The Diff-GP model~\cite{shilton2017regret} is a method that transforms $\mathcal{D}^{\prime}$ so that it can be regarded as the data from the current experiment.
The transformed $D^{\prime}$ is aggregated with the data from the current experiment and used for modeling.
Briefly, in the Diff-GP model, the difference $\Delta f = f - f^{\prime}$ between the current experiment $f$ and the similar experiment $f^{\prime}$ is modeled by the GP with mean function $\mu_t(x)$; then, the transformed dataset 
$\hat{\mathcal{D}} = \{{x}_j^{\prime}, \hat{y}_j = y_j^{\prime} + \mu_t(x_j^{\prime})\}_{j=1}^{M}$ is regarded as the data from the current experiment. 
Then, the GP model is constructed using $\mathcal{D}_t \cup \hat{\mathcal{D}}$ as a dataset, and the LSE is adaptively executed using AL, as described in the previous section.
The details of the algorithm is shown in Supplementary Information 2. 

\paragraph{Active Learning-based LSE with Diff-GP Model under the Location-Scale Shifts}
The Diff-GP model is a method that implicitly assumes that $\mathcal{D}_t$ and $\mathcal{D}^{\prime}$ are similar and models the the difference between $f$ and $f^{\prime}$ using the GP.
%
Here, if the experimental environment is improperly controlled, the output location and scale can fluctuate significantly (i.e., the above assumption is violated), and the Diff-GP model is not be directly applicable.
However, even if this assumption does not hold, cases remain in which $\mathcal{D}^{\prime}$ can be transformed to be similar to $\mathcal{D}_t$ so that the Diff-GP model can be applied.
In this study, we consider a situation in which a {\it location-scale shift}~\cite{takeuchi2009nonparametric} version of $f^{\prime}$ can be represented as
\begin{align}
    \label{eq:LSS}
    f^{\prime\prime}(x) = \gamma f^{\prime}(x) + \eta, 
\end{align}
and we can apply the Diff-GP model to $f$ and $f^{\prime\prime}$. 
Here, $\gamma$ and $\eta$ are scale and location shift parameters respectively which are estimated from the observed data. 
If $\gamma$ and $\eta$ are estimated, the dataset $\tilde{\mathcal{D}}$ obtained by transforming $\mathcal{D}^{\prime}$ according to (\ref{eq:LSS}) is integrated with $\mathcal{D}_t$, and the Diff-GP model described in the previous section is applied. 
The details of the algorithm is shown in Supplementary Information 3. 

\paragraph{Non-adaptive baseline method}
When a set of predetermined measurement points is provided, it is common to perform the measurement sequentially from the edge of the material surface  as illustrated in Figure~1{\bf b}.
However, with this approach, some of the regions remain unexamined until all measurements are completed, and the performance is poor fot a baseline method. 
Thus, we considered a baseline method based on the recursive partitioning of two-dimensional material surfaces. 
For the baseline method, as shown in Supplementary Figure~1, the centers of the recursively generated rectangles, which are defined by the already measured points, are selected as new measurement points.

\paragraph{Reporting Summary}
Further information on the experimental design is available in the Nature Research Reporting Summary linked to this article. 

\paragraph{Data Availability}
Data supporting the findings of this study are available from the corresponding author upon reasonable request. 

\clearpage

\bibliographystyle{unsrt}
\bibliography{ref}

\clearpage

\noindent
{\bf Acknowledgements}\\
This work was supported by Grants-in-Aid from the Japan Society for the Promotion of Science (JSPS) for Scientific Research 
(KAKENHI grant nos. 
20K19871 and 19H04071 to K.M. and, 
20H00601 to I.T.), 
the Japan Synchrotron Radiation Research Institute (JASRI) (Proposal nos. 2018A3585, 2018B3587, 2019A3587, 2019B3586, and 2020A3586) to K.K., 
the Japan Science and Technology Agency (JST) CREST (grant no. JPMJCR21D3), 
JST Moonshot R$\&$D (grant no. JPMJMS2033-05), 
JST AIP Acceleration Research (grant no. JPMJCR21U2), and
the New Energy and Industrial Technology Development Organization (NEDO, grant nos. JPNP18002 and JPNP20006) to I.T.; 
K.K, K.M., T.U., and I.T. received support from the RIKEN Center for Advanced Intelligence Project.

\vspace*{12pt}

\noindent
{\bf Author contributions}\\
K.K., K.M., T.U., and I.T. contributed to the study design. K.K. collected, managed, and analyzed the solar cell ingots data. H.S., K.M., K.K., and I.T. constructed the machine learning methods (AL-based LSE and ATL-based LSE methods). H.S. and S.K. implemented the methods and conducted numerical experiments. H.S., K.K, K.M., and I.T. wrote the manuscript. All authors discussed and commented on the manuscript. 

\vspace*{12pt}

\noindent
{\bf Competing Iinterests}\\
The authors declare that they have no conflicts of interests. 

\clearpage

\begin{figure*}
\includegraphics[bb = 0 0 496 482, clip, width=0.90\textwidth]{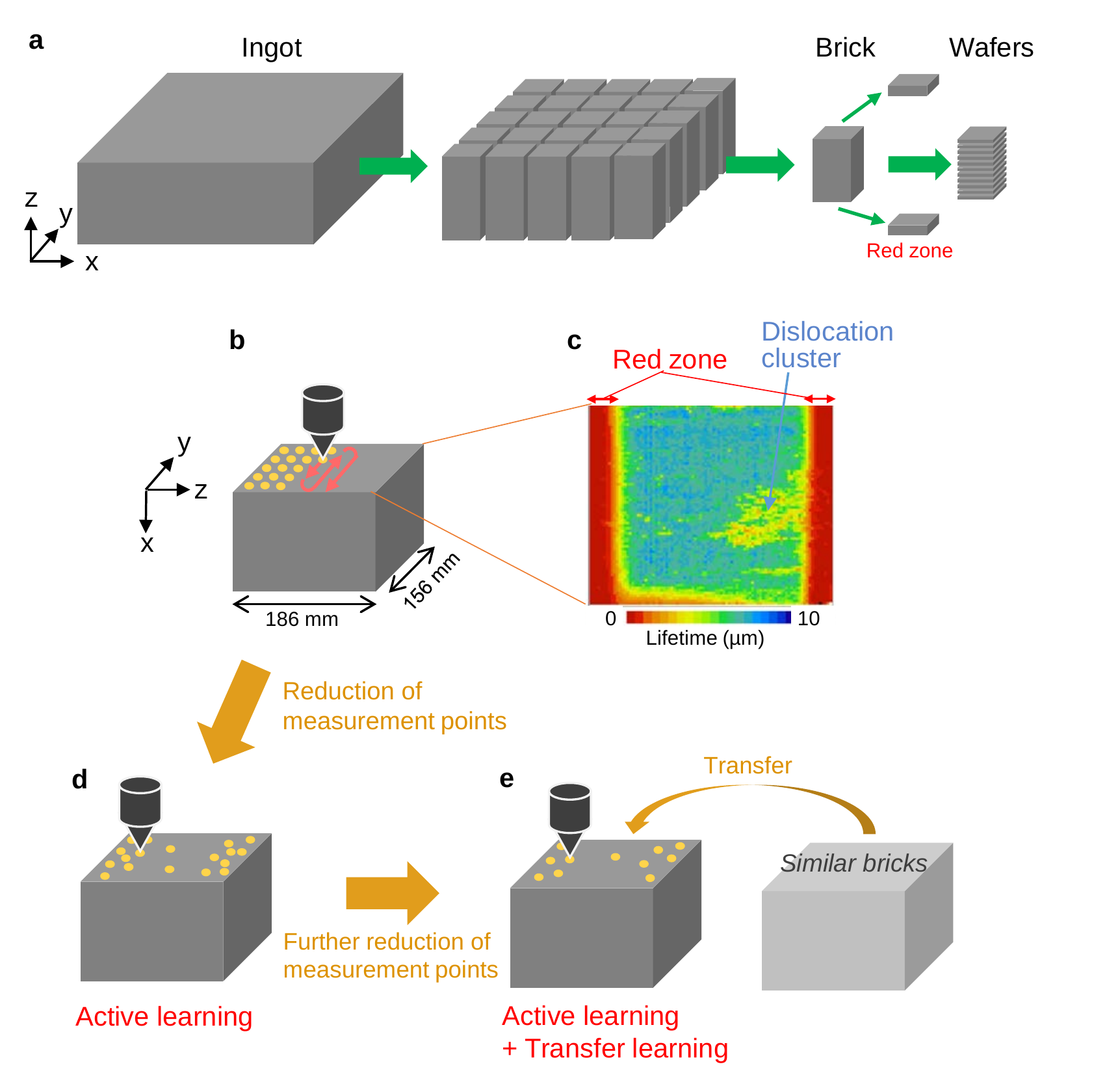}
\end{figure*}
\noindent
{\bf Figure 1}:
An illustration of the red-zone identification problem for silicon ingots for solar cells.
{\bf a}
Fabrication process of multicrystalline silicon wafers from an ingot.
Red-zones are removed from the bricks before the wafering process.
{\bf b}
Conventional spatial mapping of the carrier lifetime on the mesh grid points of the y-z brick side surface, i.e., ingot cross-section point-by-point.
{\bf c}
Typical spatial distribution of carrier lifetime. 
{\bf d}
Reduction of the measurement points by active learning (AL)-based adaptive spatial mapping.
{\bf e}
Further reduction of the measurement points by transfer learning (TL) using information from previously produced similar bricks.
\clearpage 

\begin{center}
\begin{tabular}{ll}
{\bf a} &
  {\bf b} \\
\includegraphics[bb=0 0 360 416, clip, width=0.36\textwidth]{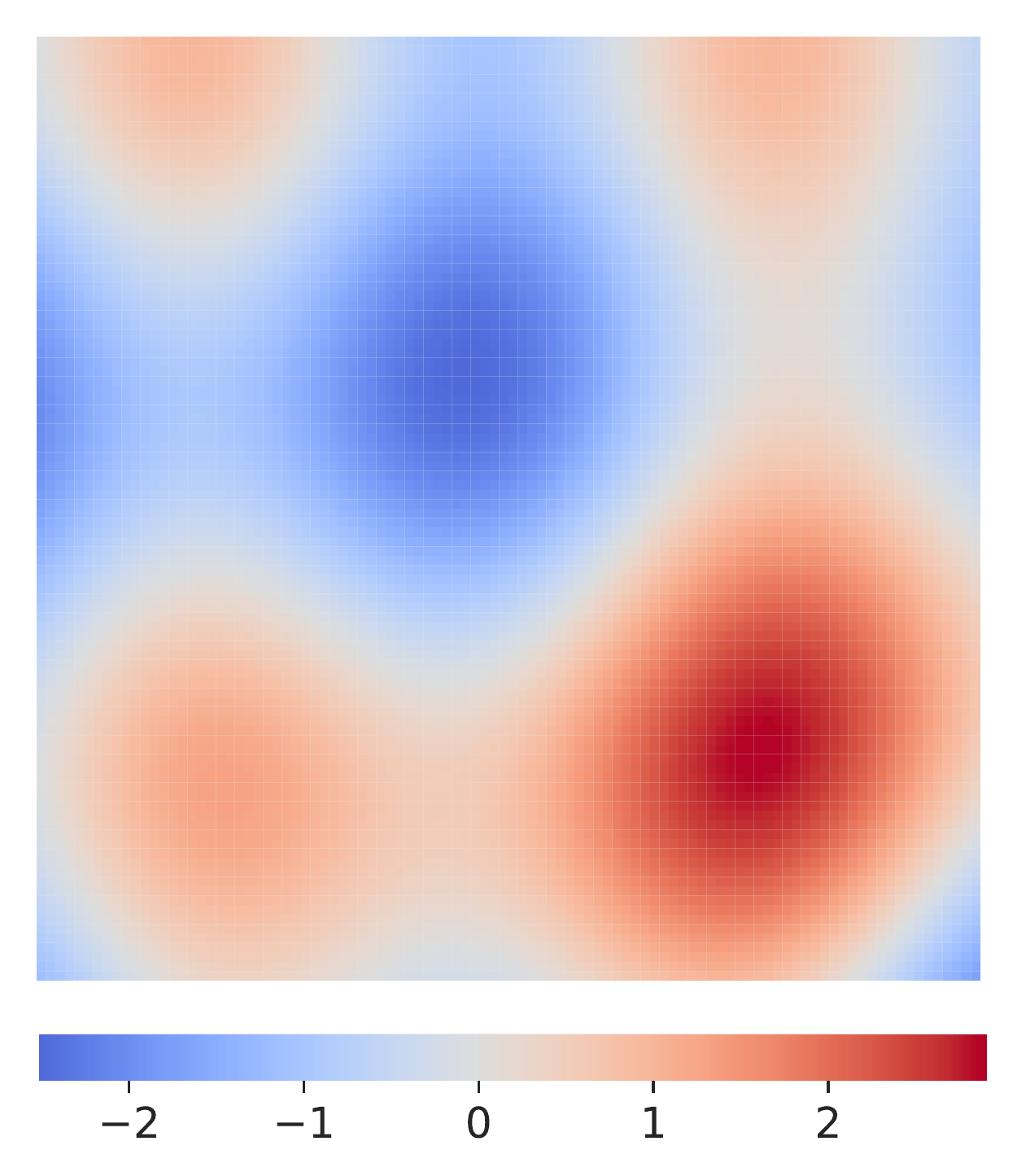} &
\includegraphics[bb=0 0 360 360, clip, width=0.42\textwidth]{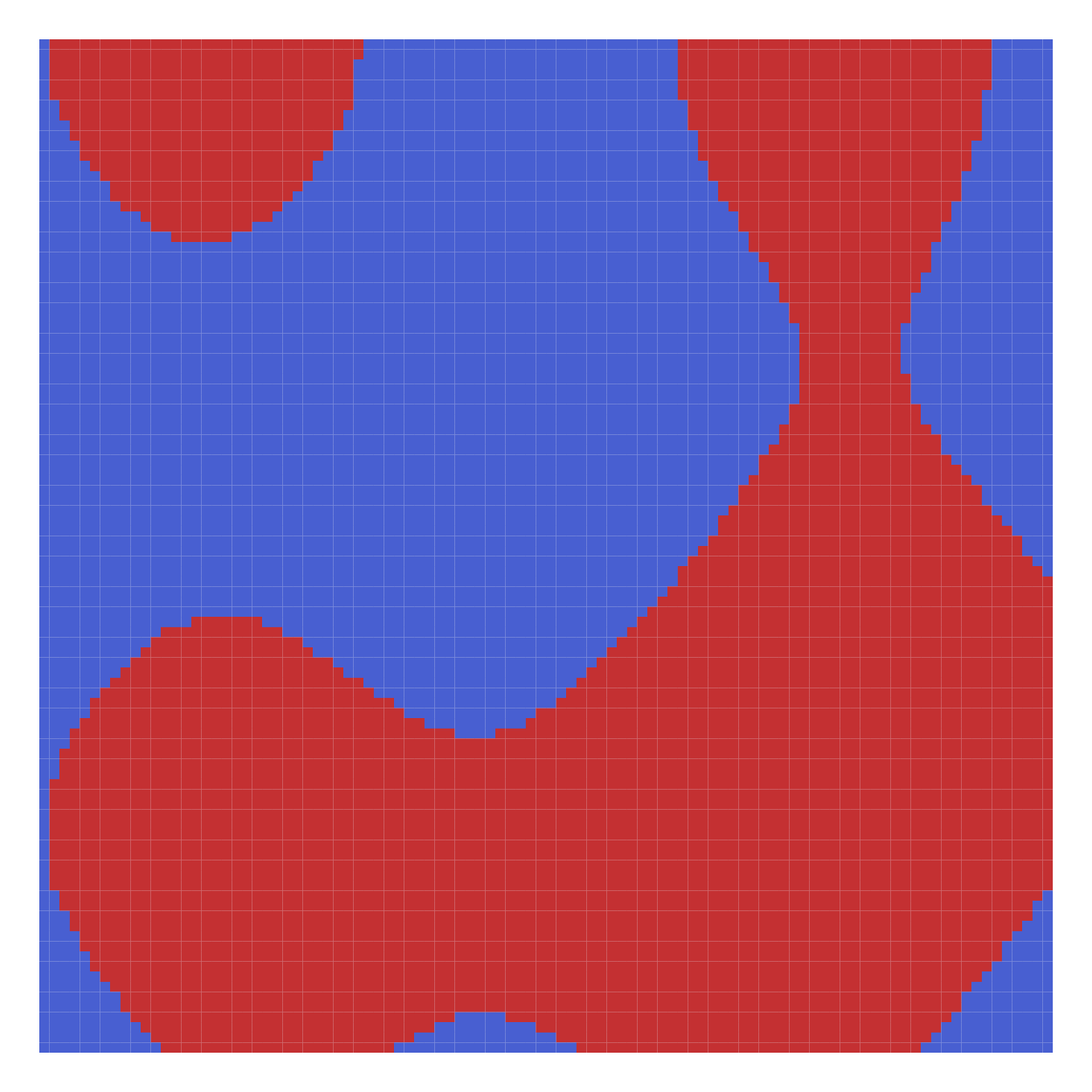} \\
{\bf c} &
  {\bf d} \\
\includegraphics[bb=0 0 360 360, clip, width=0.42\textwidth]{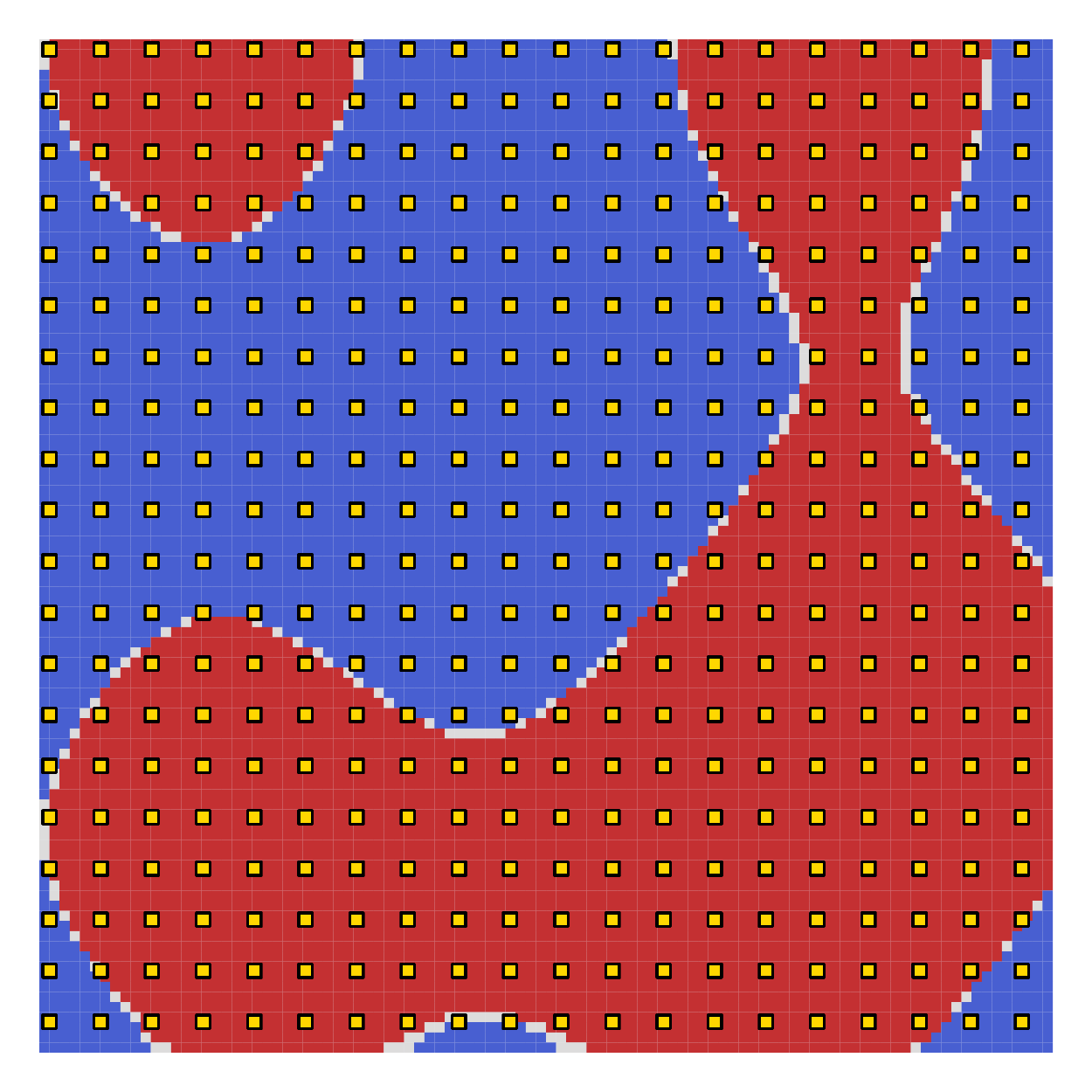} & 
\includegraphics[bb=0 0 360 360, clip, width=0.42\textwidth]{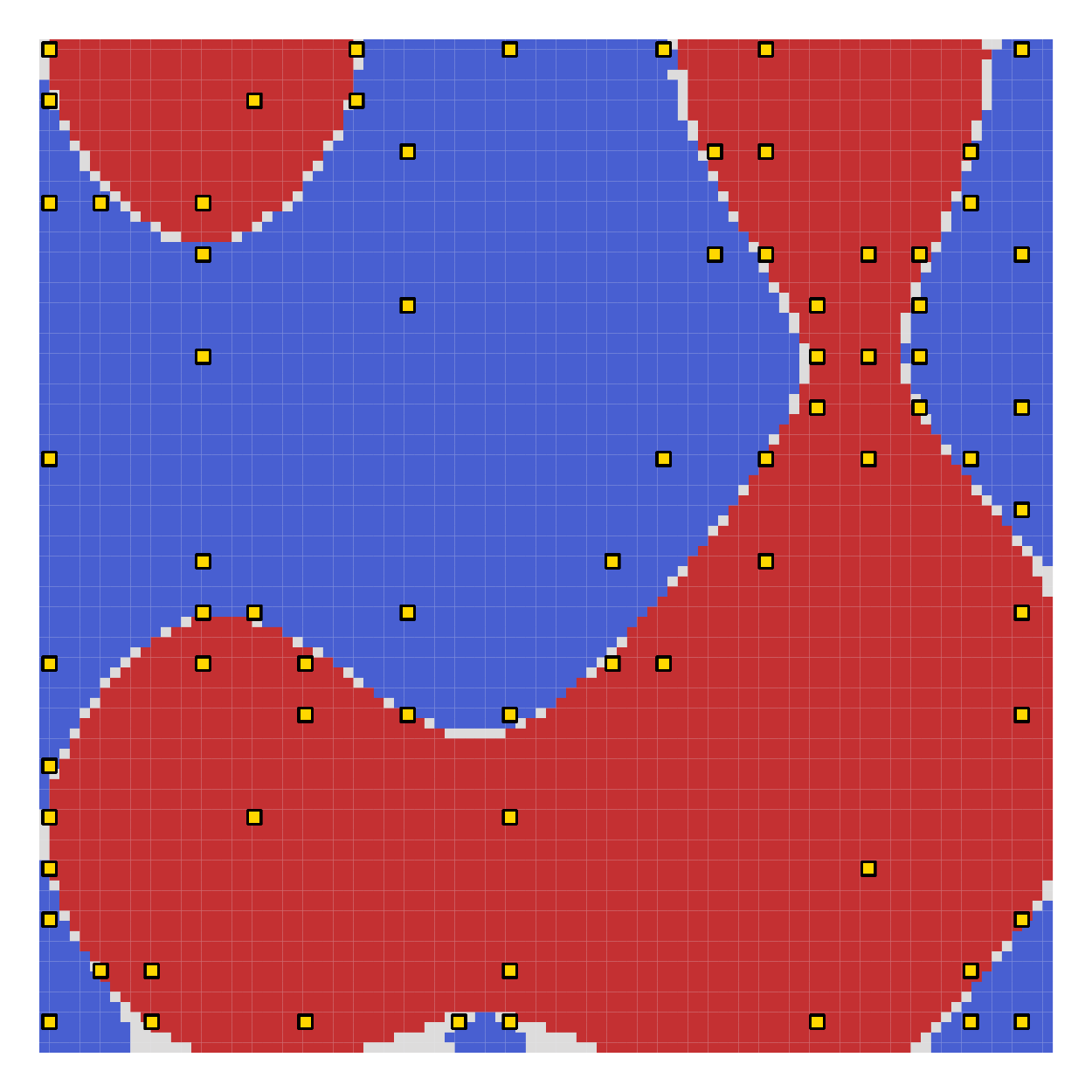}  
\end{tabular}
\end{center}
\noindent
{\bf Figure 2}:
An illustration of level set estimation (LSE) problems for a two-dimensional function $f$; the colors indicate the values of the function defined over a two-dimensional input space $\cX$. 
{\bf a}
Example of true spatial distribution of physical properties of a material surface.
{\bf b}
The super- and sub-level sets for a threshold $\theta$ are defined as $\cU_\theta = \{\bm x \in \cX \mid f(\bm x) > \theta\}$ and $\cL_\theta = \{\bm x \in \cX \mid f(\bm x) \le \theta\}$, respectively. 
For the problem in Figure~1, the red-zone is interpreted as the sub-level set for a given threshold of the career lifetime. 
{\bf c}
Conventional LSE using \emph{exhaustive} measurements at all predetermined mesh grid points.
The super- and sub-level sets are accurately estimated once the function values at all predetermined mesh grid points are evaluated; however, this approach is often inefficient. 
{\bf d}
Active learning (AL)-based LSE using \emph{adaptive} measurements at selected grid points. 
The super- and sub-level sets, $\cU_\theta$ and $\cL_\theta$, are accurately estimated, even if the functions are evaluated only at selected grid points near the boundary of the defective area. 
\clearpage

\begin{center}
\label{Fig:03}
 \begin{tabular}{ccc}
  \includegraphics[bb=0 0 432 288, clip, width=0.3\textwidth]{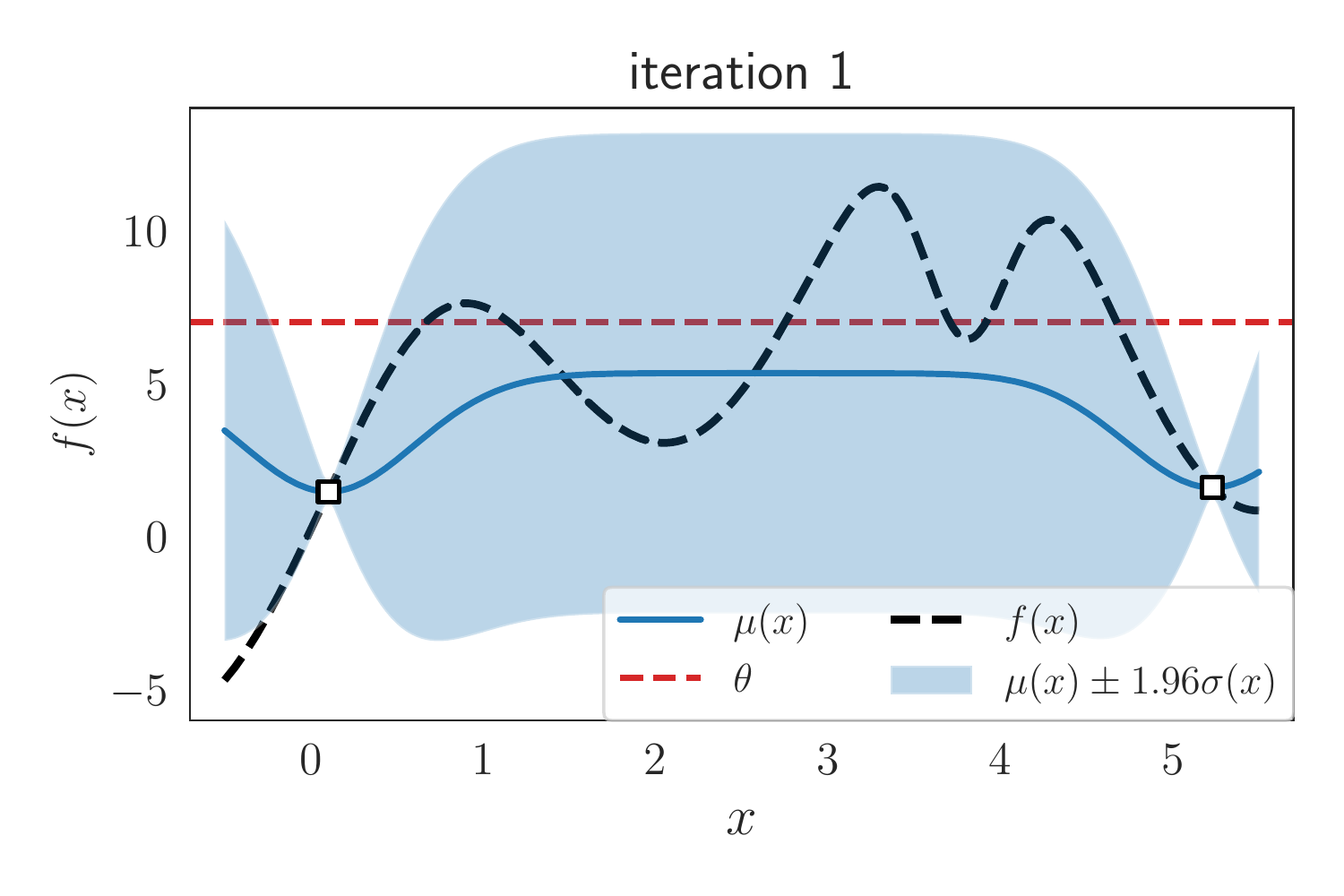} &
  \includegraphics[bb=0 0 432 288, clip, width=0.3\textwidth]{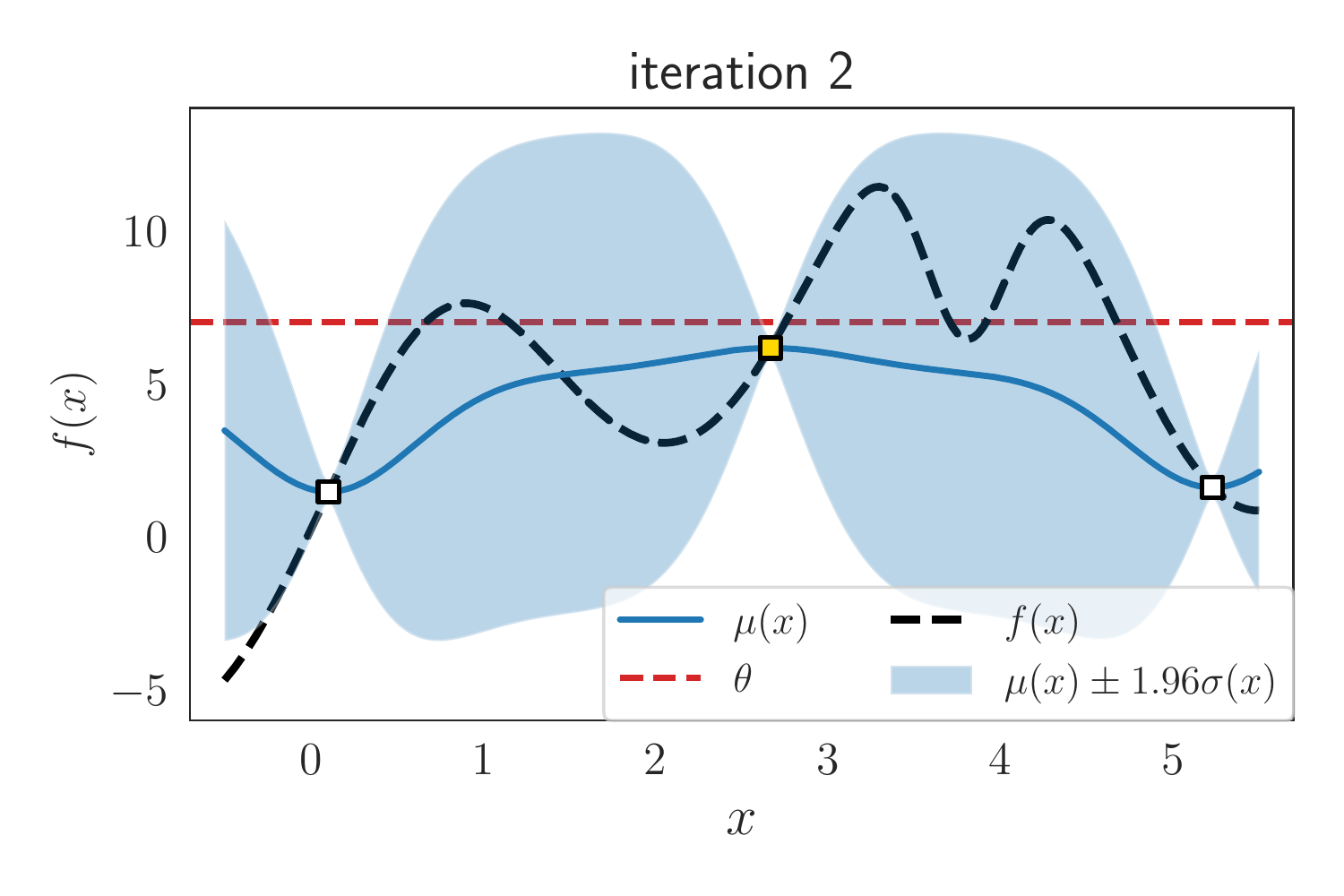} &
  \includegraphics[bb=0 0 432 288, clip, width=0.3\textwidth]{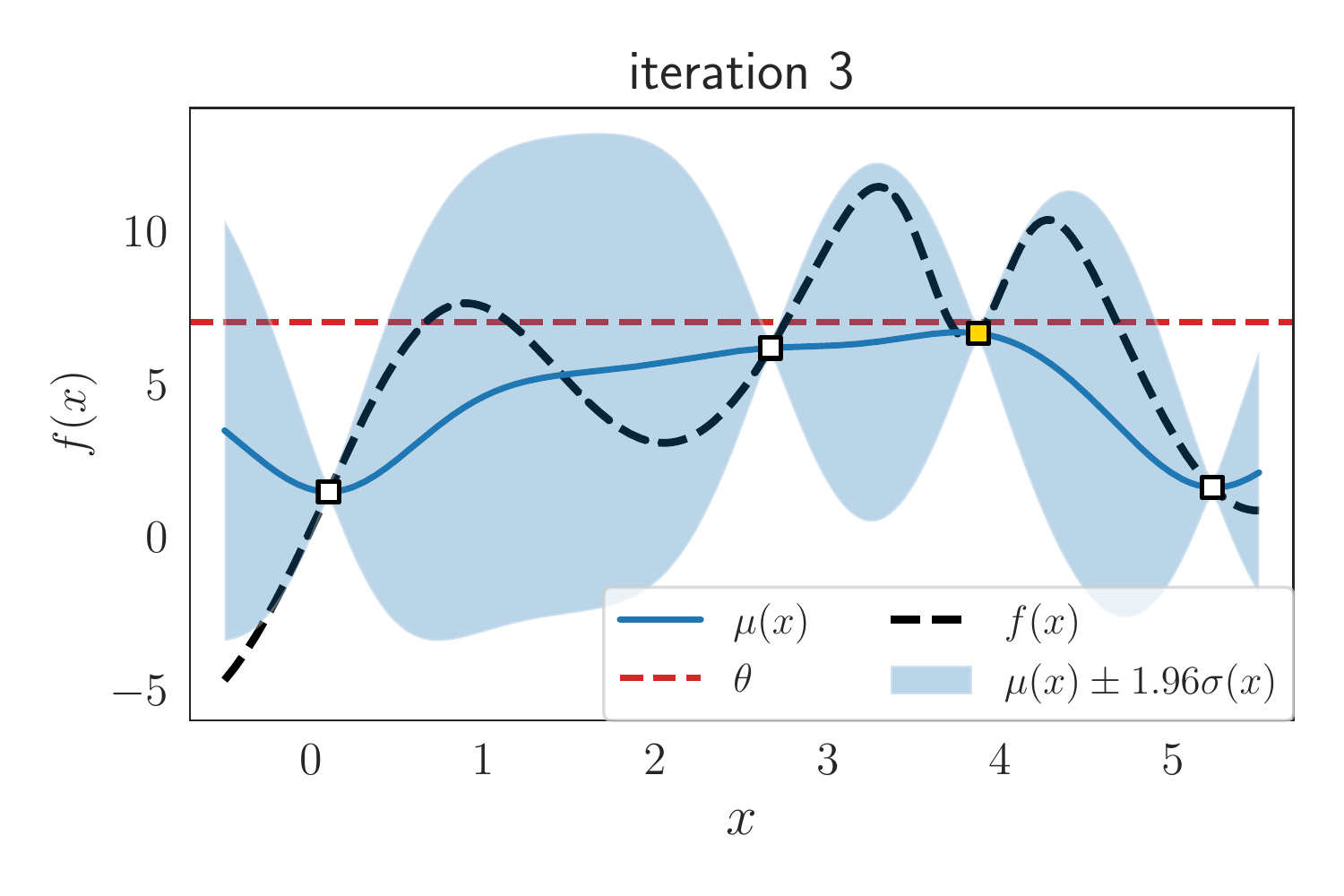} \\
  \includegraphics[bb=0 0 432 288, clip, width=0.3\textwidth]{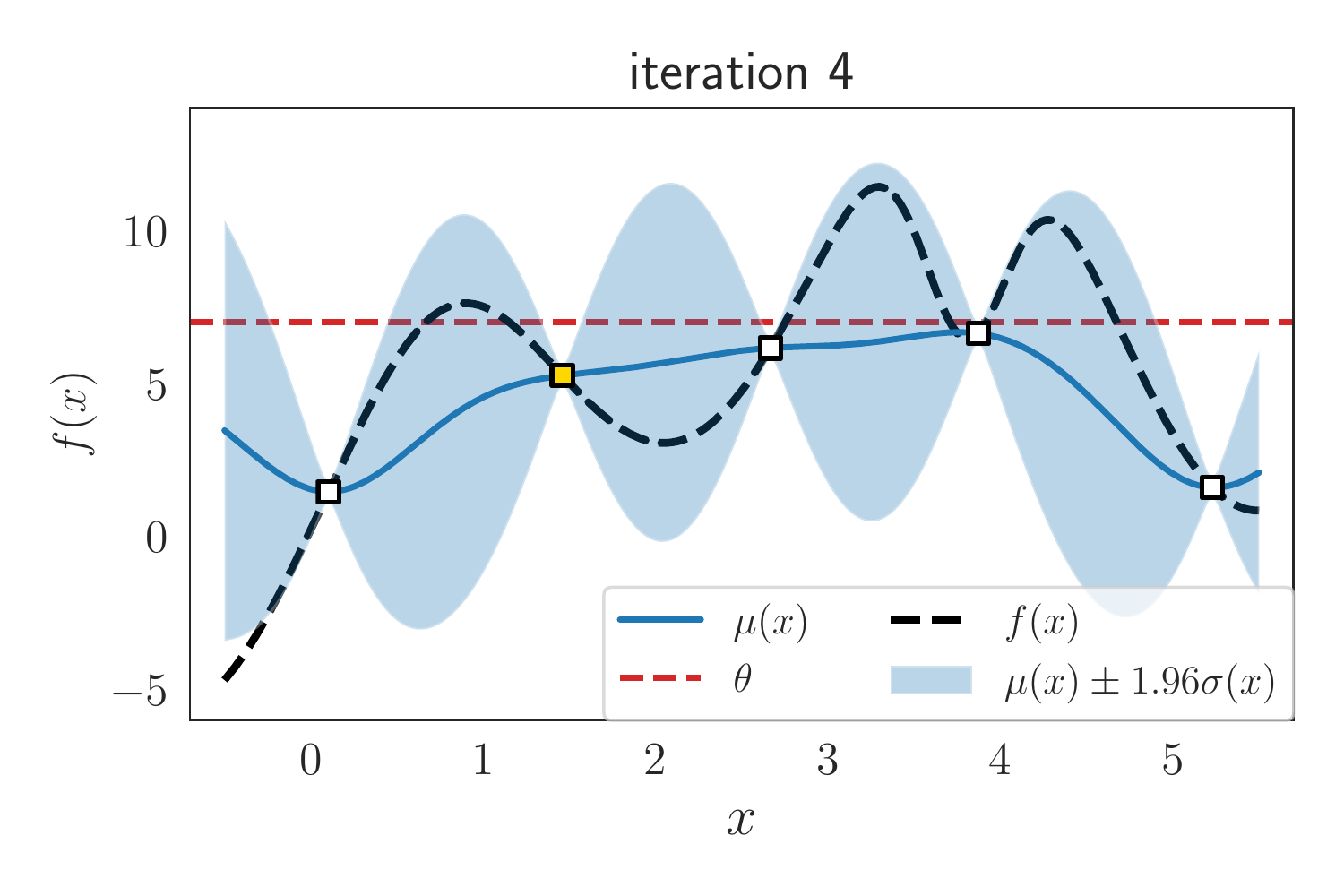} &
  \includegraphics[bb=0 0 432 288, clip, width=0.3\textwidth]{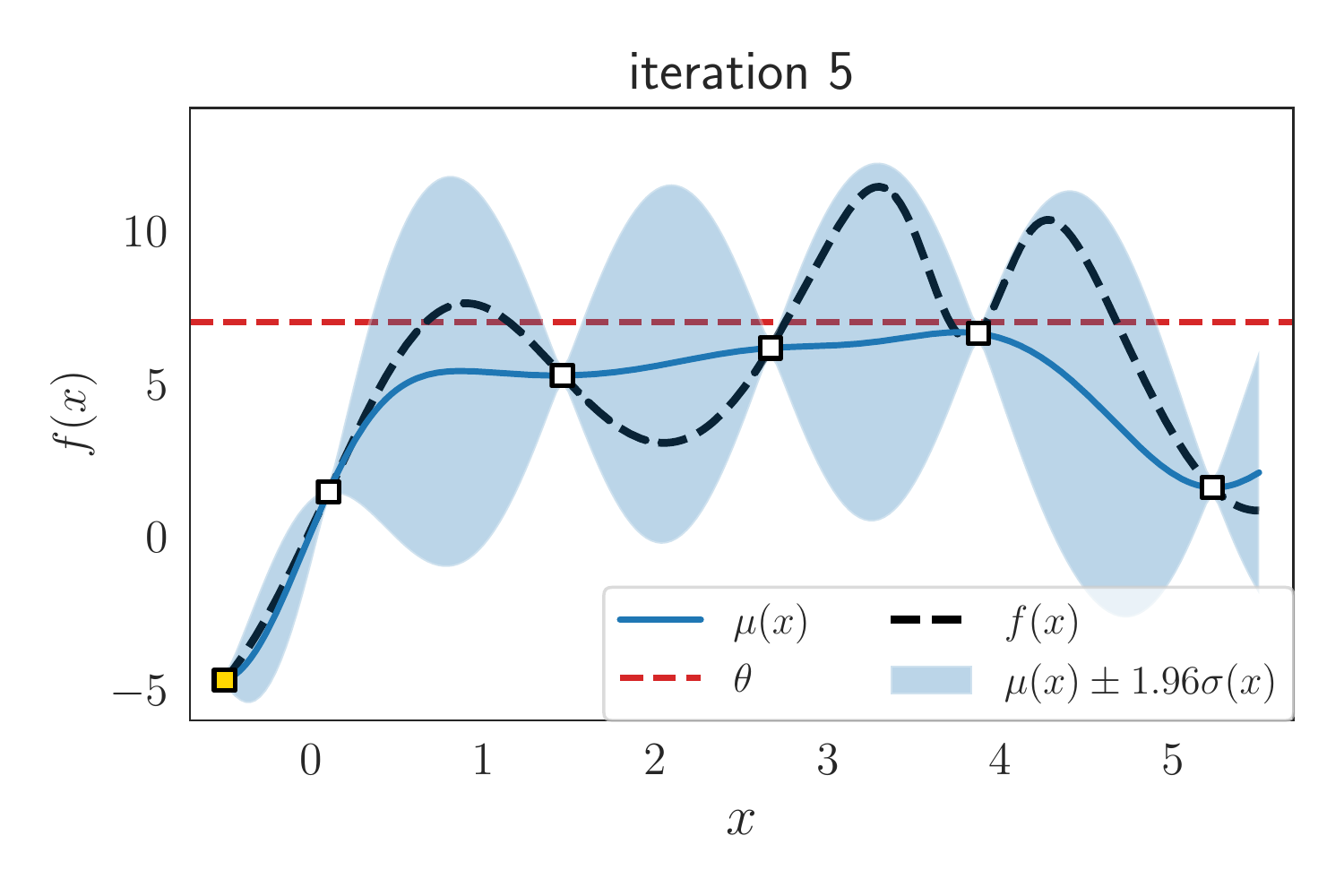} &
  \includegraphics[bb=0 0 432 288, clip, width=0.3\textwidth]{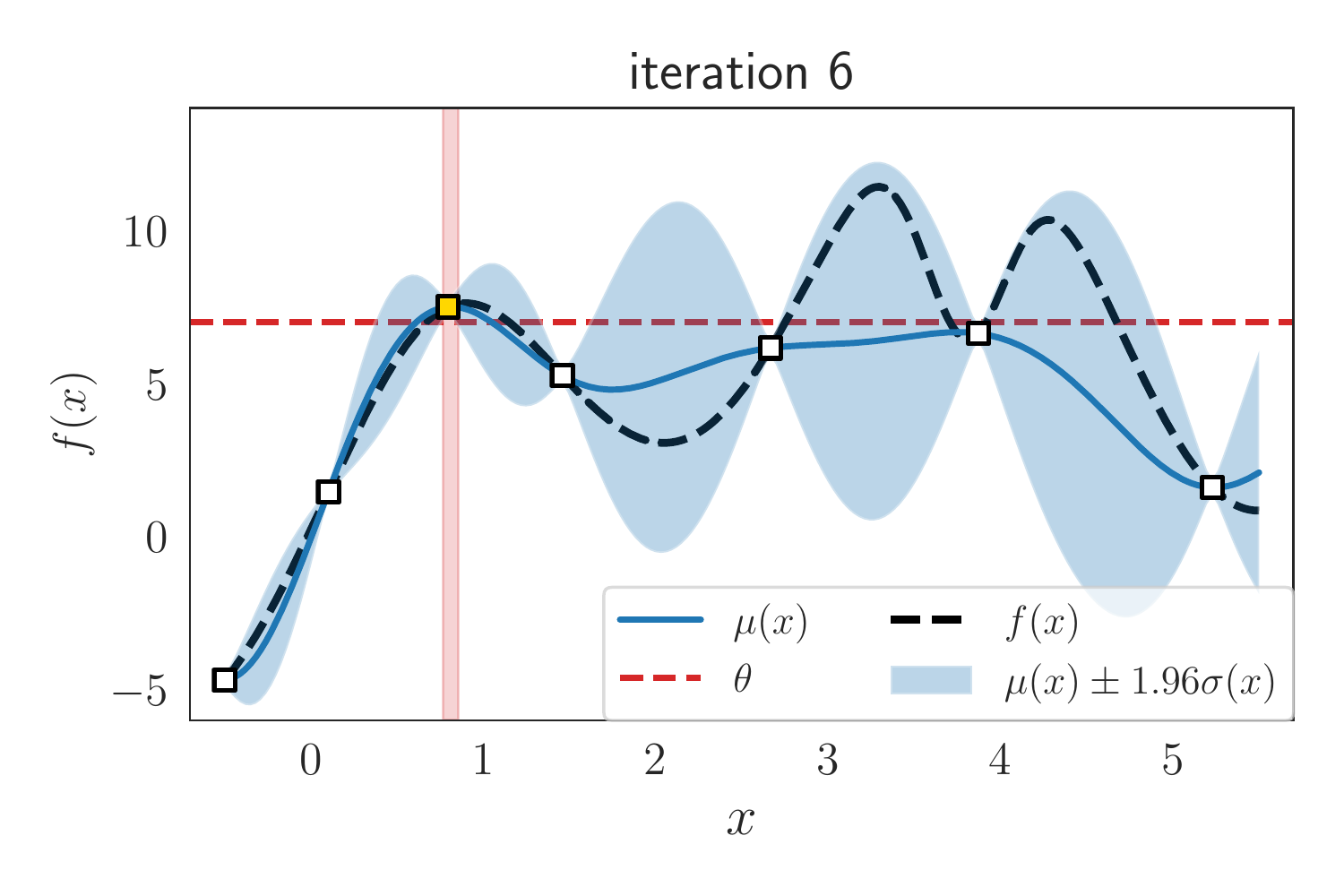} \\
  \includegraphics[bb=0 0 432 288, clip, width=0.3\textwidth]{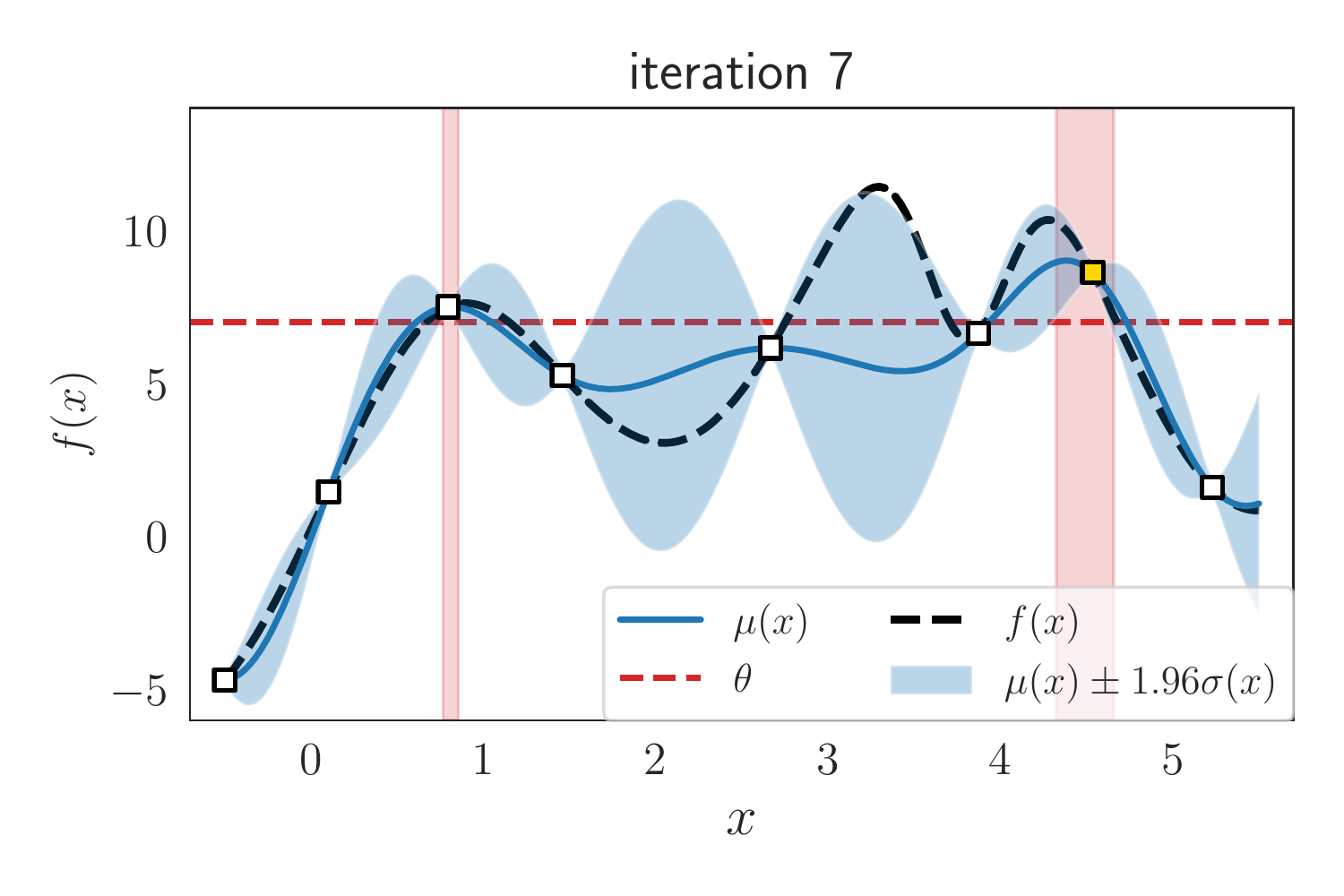} &
  \includegraphics[bb=0 0 432 288, clip, width=0.3\textwidth]{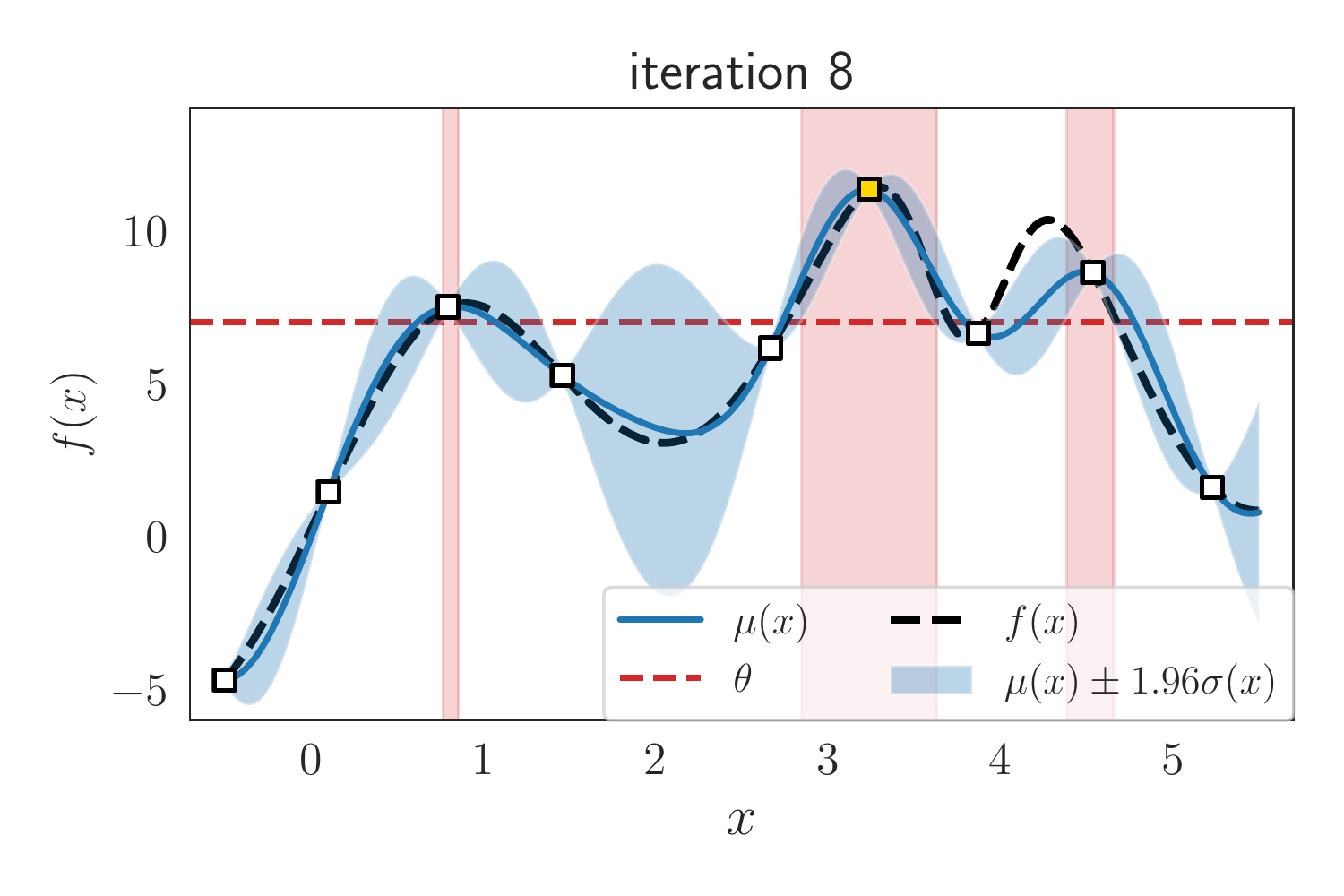} &
  \includegraphics[bb=0 0 432 288, clip, width=0.3\textwidth]{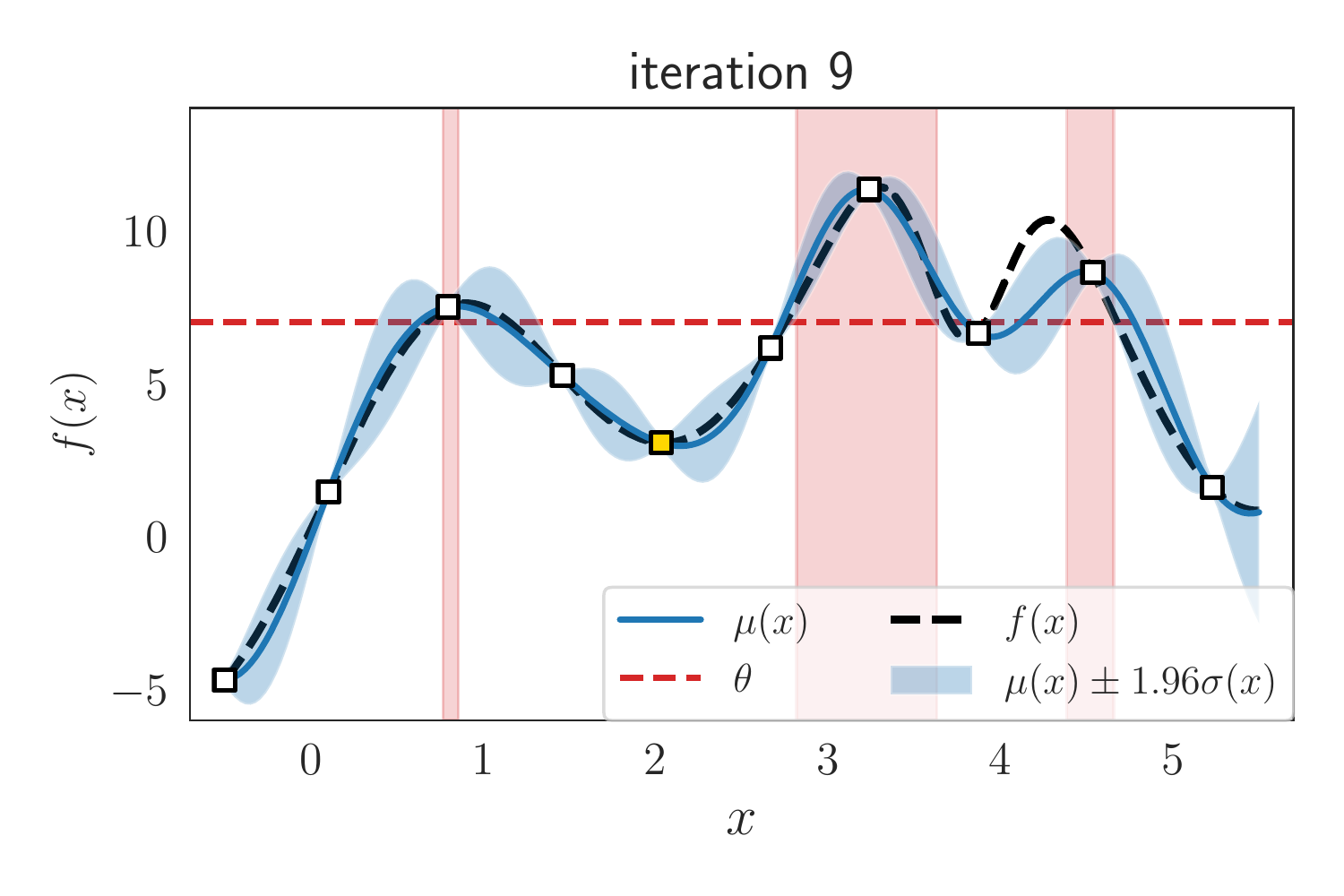} \\
  \includegraphics[bb=0 0 432 288, clip, width=0.3\textwidth]{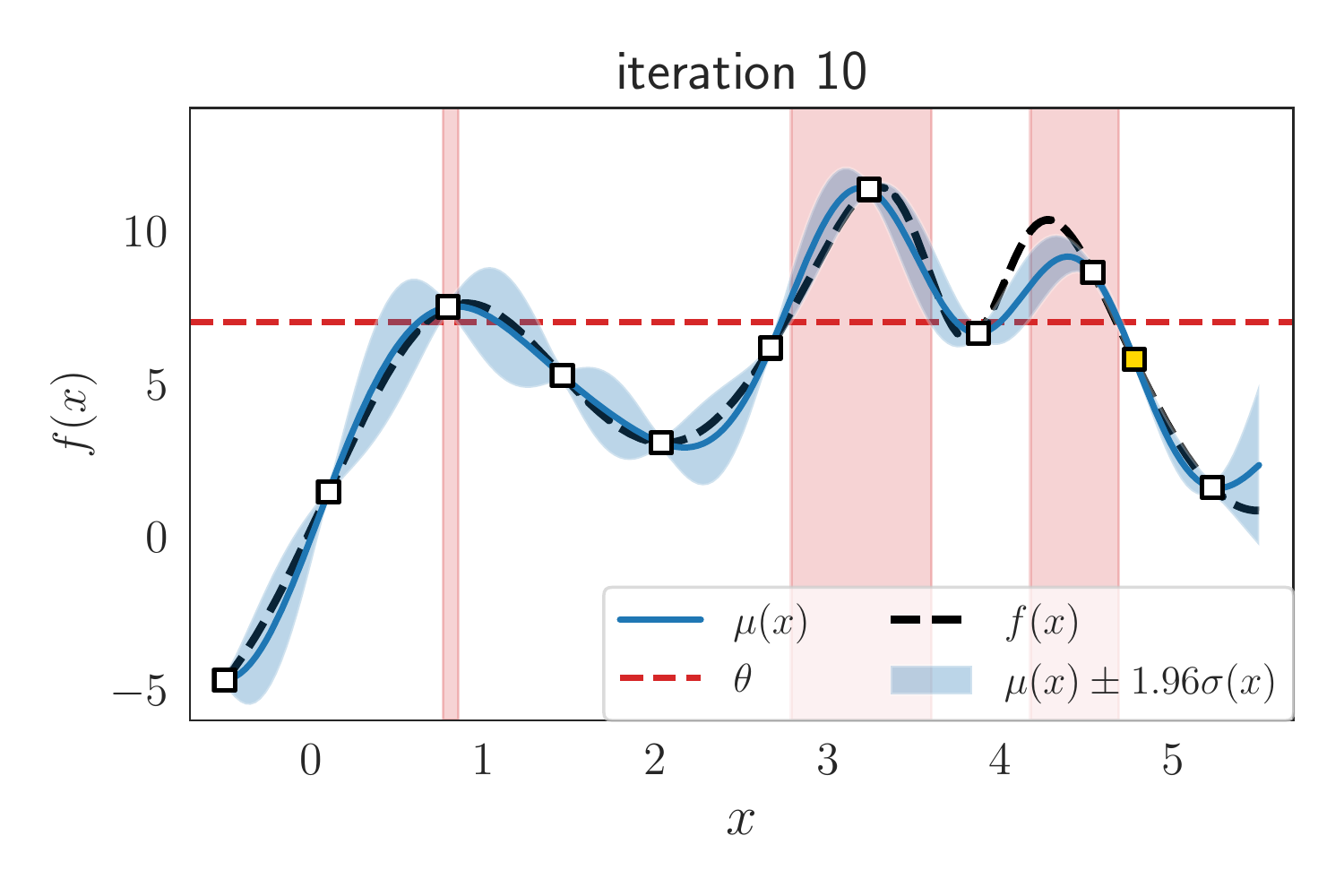} &
  \includegraphics[bb=0 0 432 288, clip, width=0.3\textwidth]{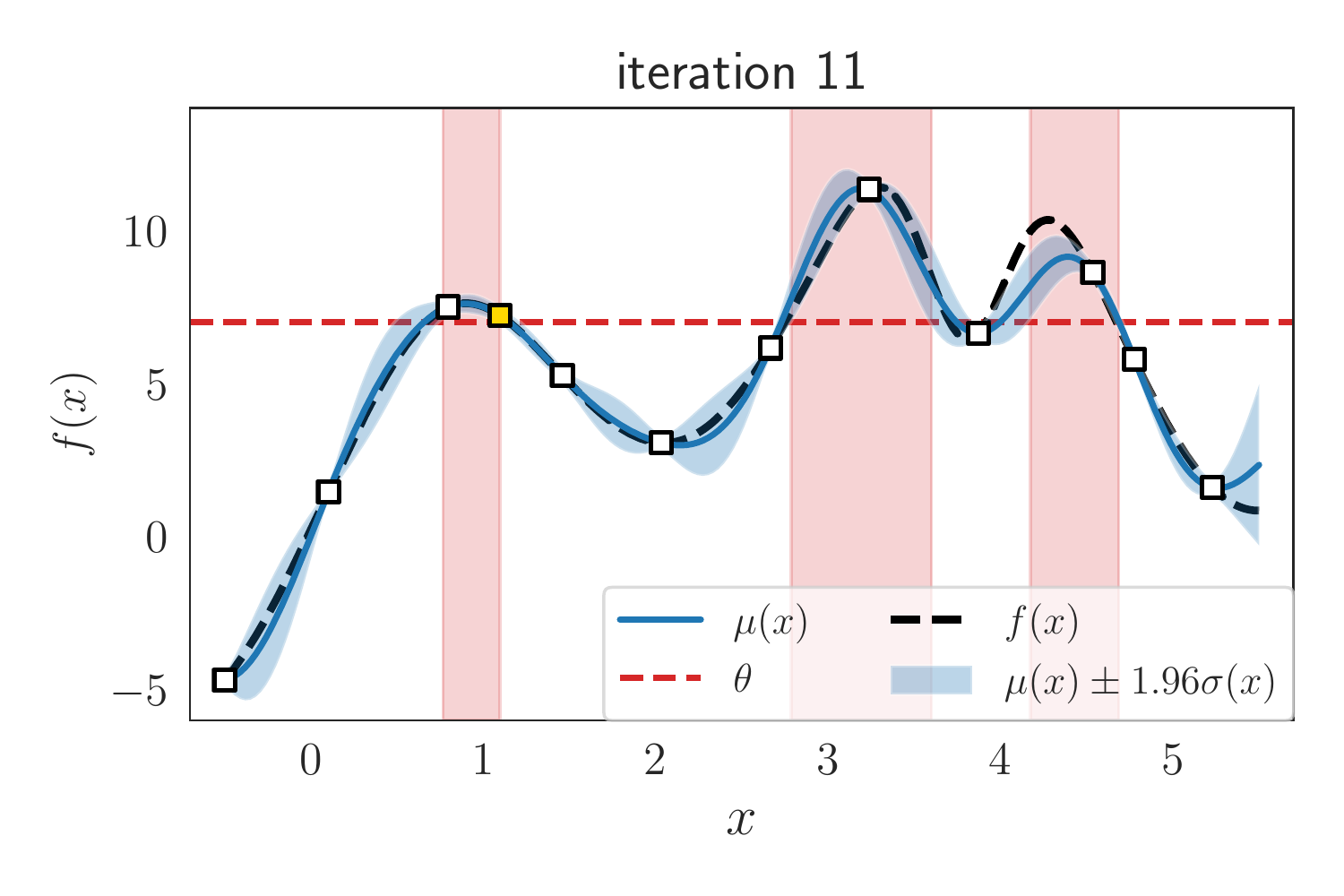} &
  \includegraphics[bb=0 0 432 288, clip, width=0.3\textwidth]{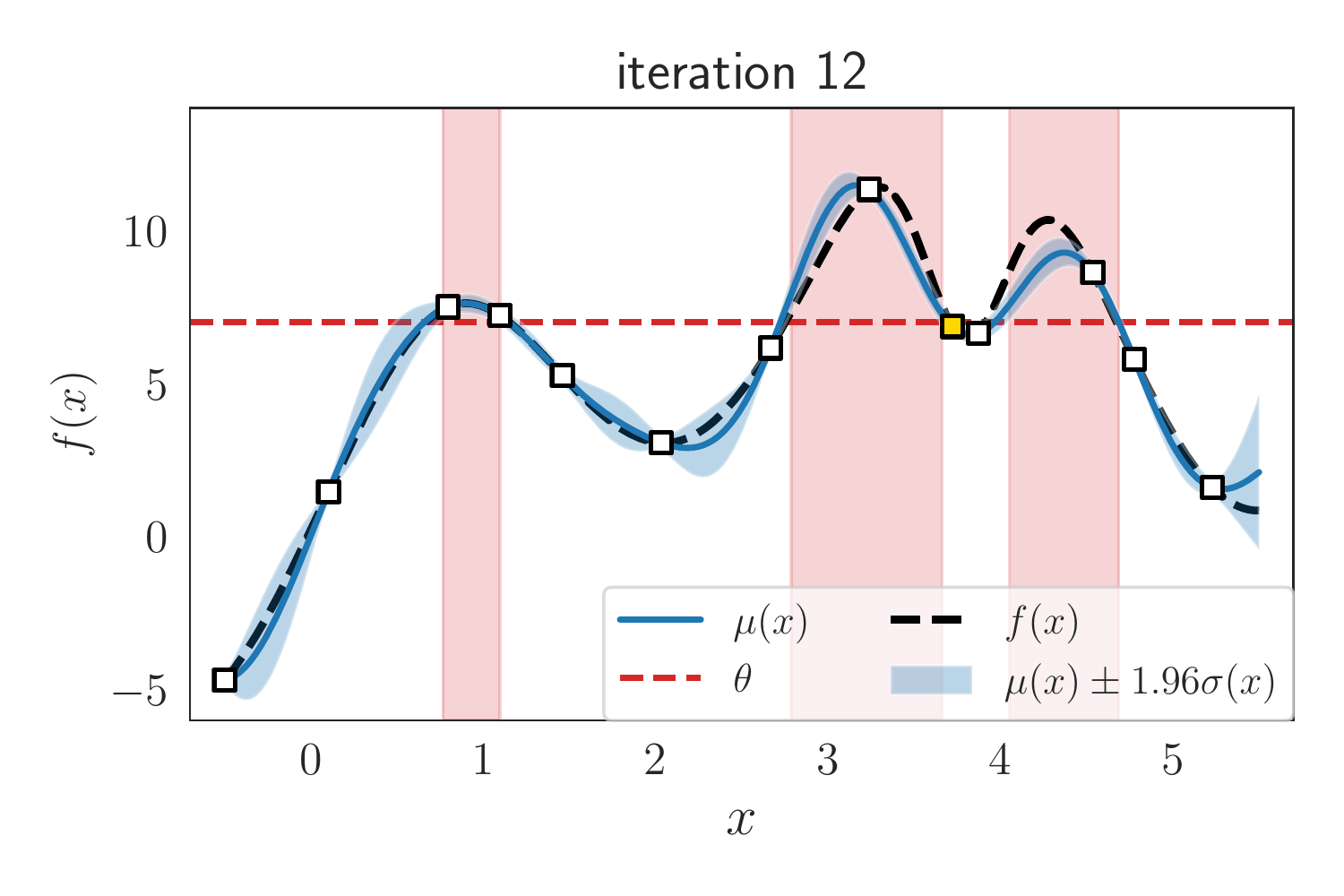} \\
  \includegraphics[bb=0 0 432 288, clip, width=0.3\textwidth]{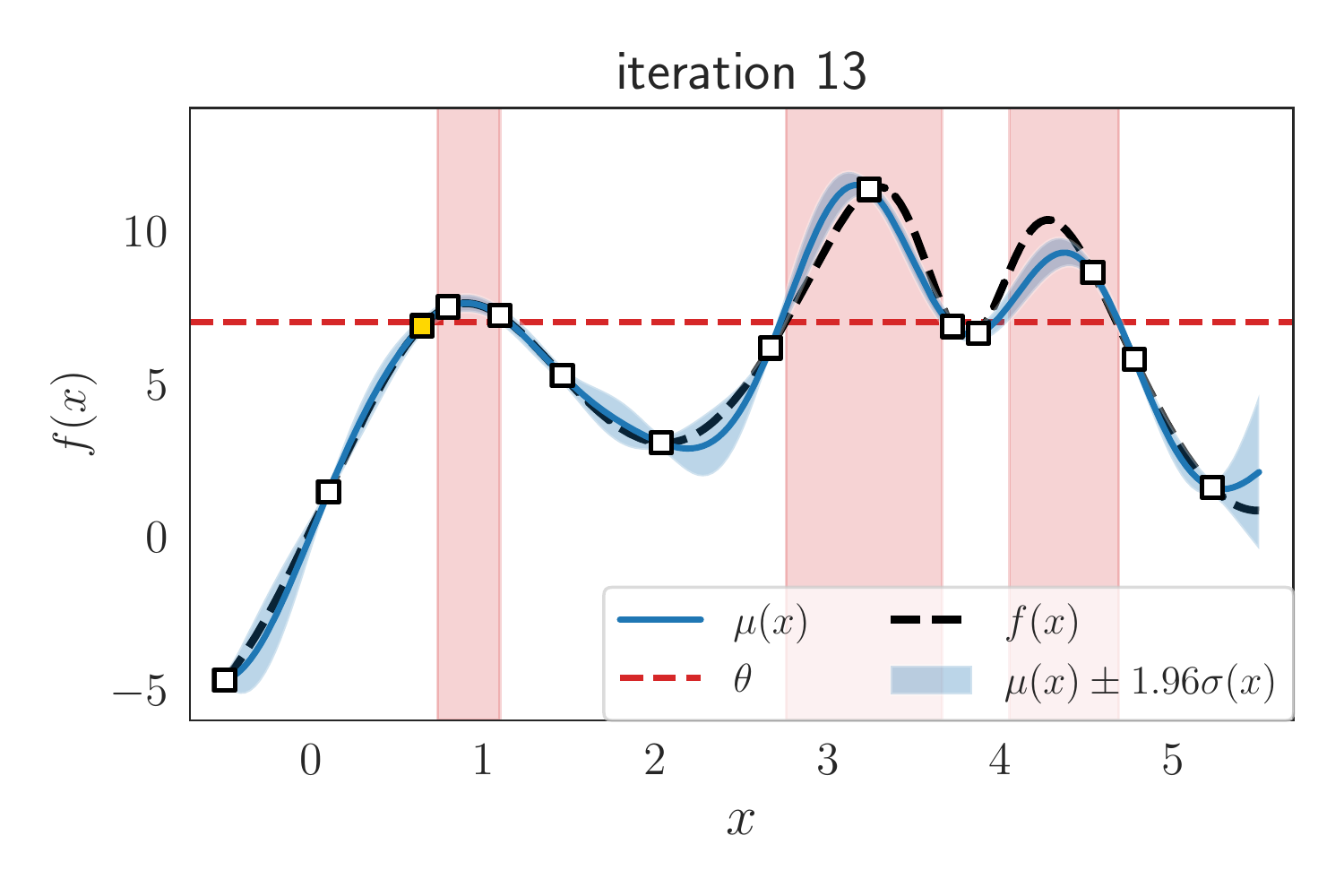} &
  \includegraphics[bb=0 0 432 288, clip, width=0.3\textwidth]{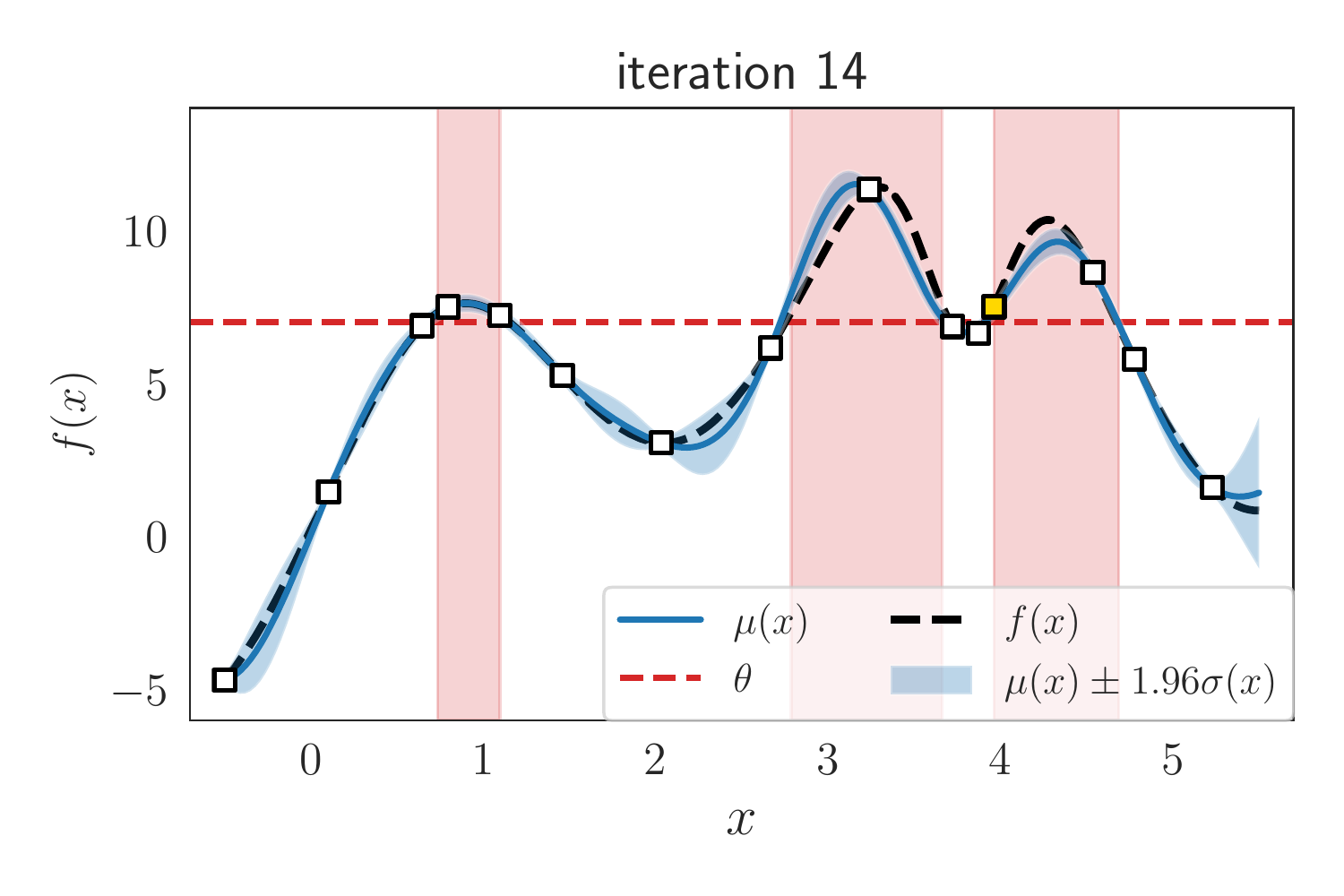} &
  \includegraphics[bb=0 0 432 288, clip, width=0.3\textwidth]{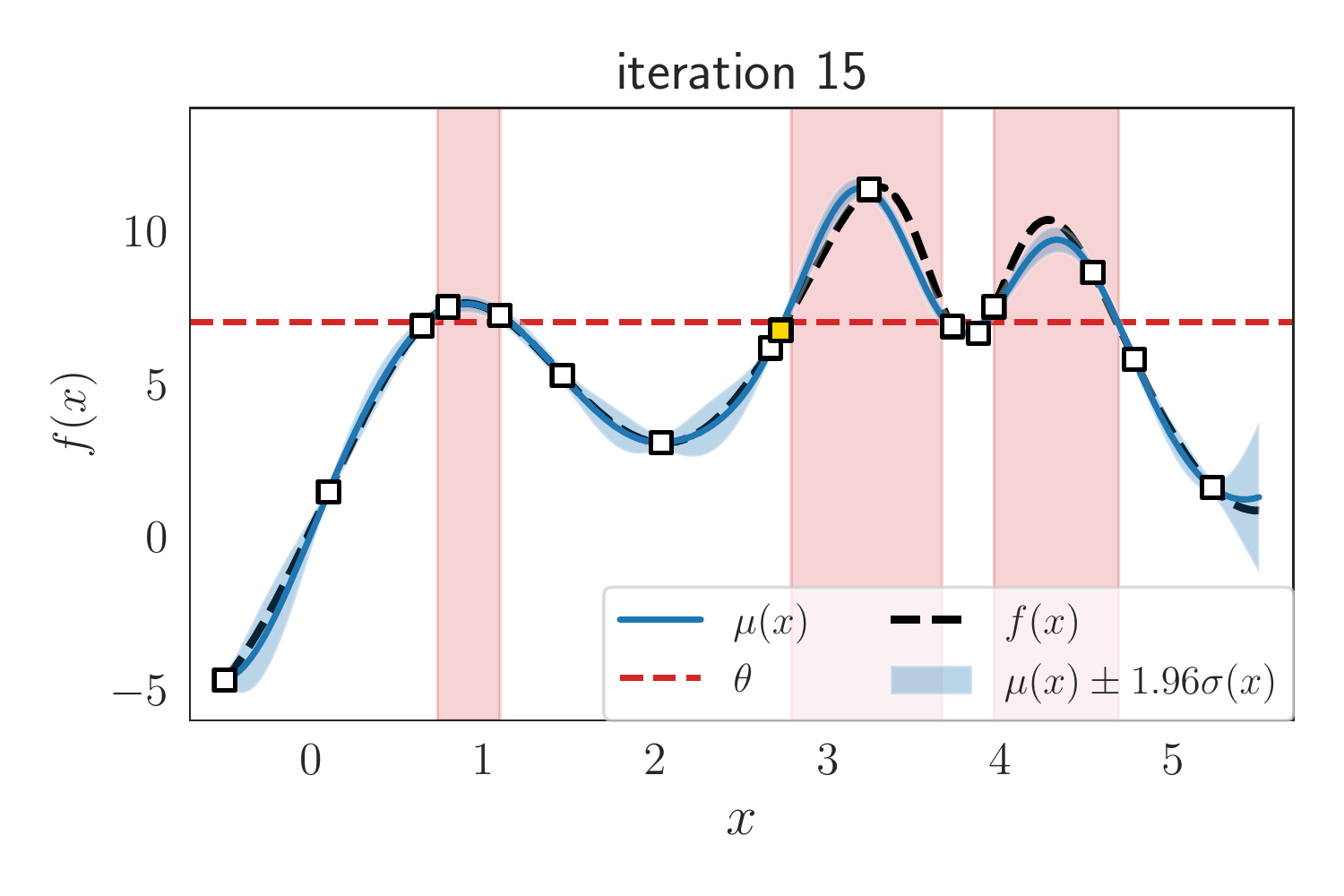} 
 \end{tabular}
\end{center}
\noindent
{\bf Figure 3}:
An illustration of defective area identifications using the AL-based LSE method for an illustrative one-dimensional function $f$. 
Each plot from top left to bottom right shows the training data points (squares) obtained so far and the Bayesian regression model at each iteration.
Fitting a Bayesian regression (Gaussian process) model yields the mean $\mu(x)$ (dark blue curve) and confidence interval $\mu(x) \pm 1.96 \sigma(x)$ (light blue region) at each input $x \in \cX$.
The goal of this LSE problem is to identify the subset of input positions at which the function value is greater than $\theta$ (red dashed line); the red shaded region in each plot indicates the estimated defective areas.
This example demonstrates that the input points are more likely to be selected at positions where the function values are close to the threshold.
\clearpage

\begin{center}
 \begin{tabular}{ll}
  {\bf a} &
  {\bf b} \\
  \includegraphics[bb=0 0 243 175, clip, width=0.5\textwidth]{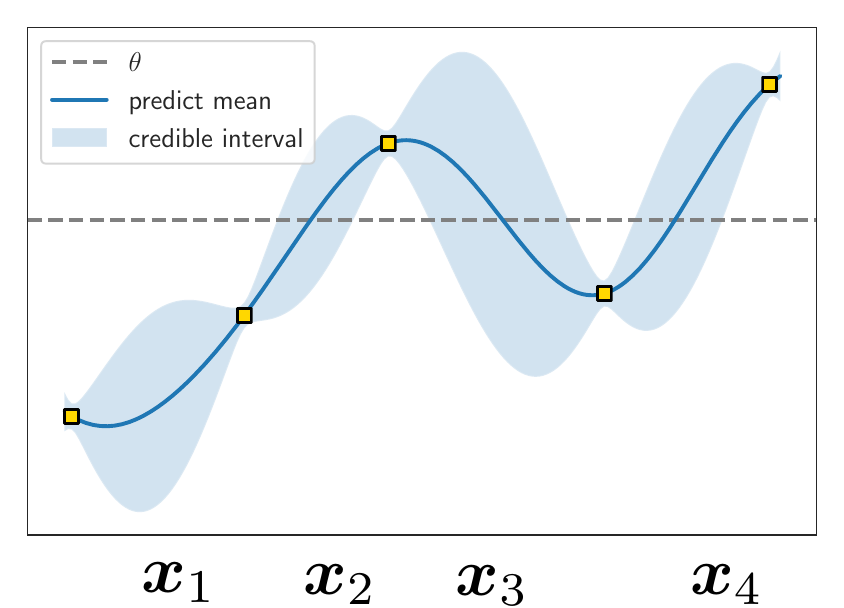} &
  \includegraphics[bb=0 0 243 175, clip, width=0.5\textwidth]{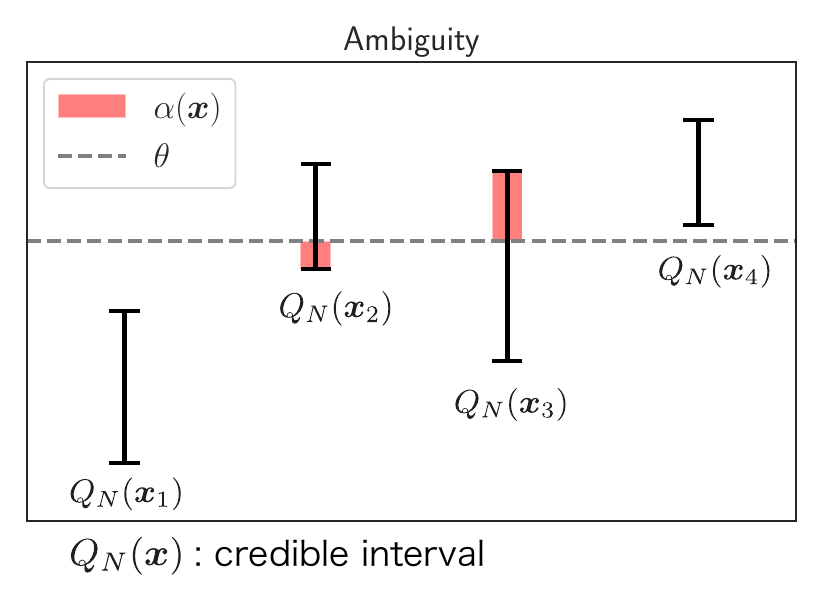}  
 \end{tabular}
\end{center}
\noindent
{\bf Figure 4}:
An illustration of adaptive selection criteria for the AL-based LSE method for illustrative one-dimensional function $f$. 
{\bf a} Bayesian regression model at a certain iteration.
{\bf b} With four input candidates $x_1, x_2, x_3, x_4$, the method selects $x_3$ because the violation $\alpha(x_3)$ (light-red region) is the largest among the four candidates.
The violation at each input candidate is defined as $\alpha(x) = \min\{\max\{0, \theta - \ell(x)\}, \max\{0, u(x) - \theta\}\}$ where $\ell(x)$ and $u(x)$ are the lower and upper ends of the credible interval $Q_N(x)$.
\clearpage


\noindent
{\bf a}
\begin{center}
 \begin{tabular}{cccc}
  \includegraphics[bb=0 0 461 346, clip, width=0.3\textwidth]{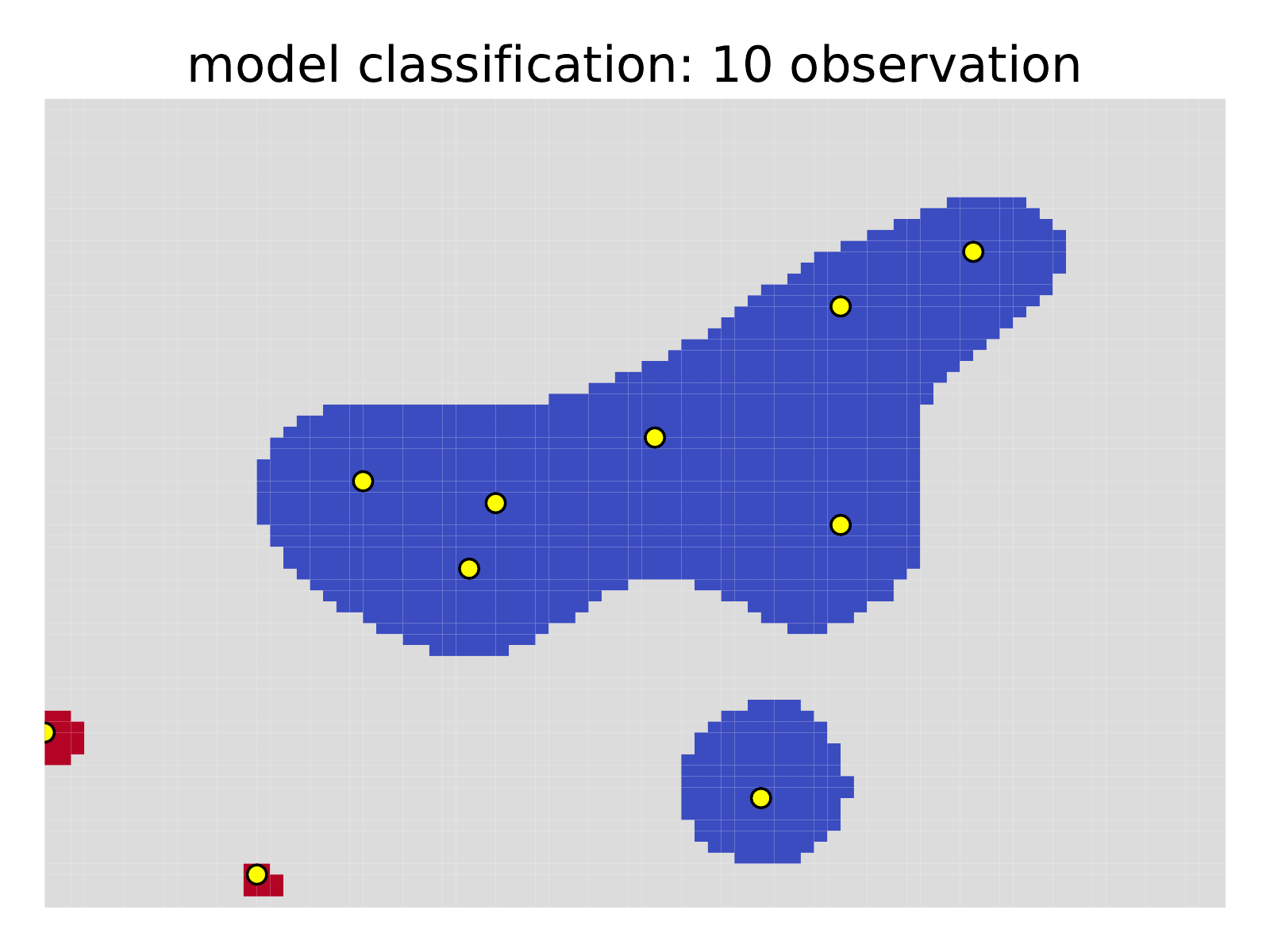} &
  \includegraphics[bb=0 0 461 346, clip, width=0.3\textwidth]{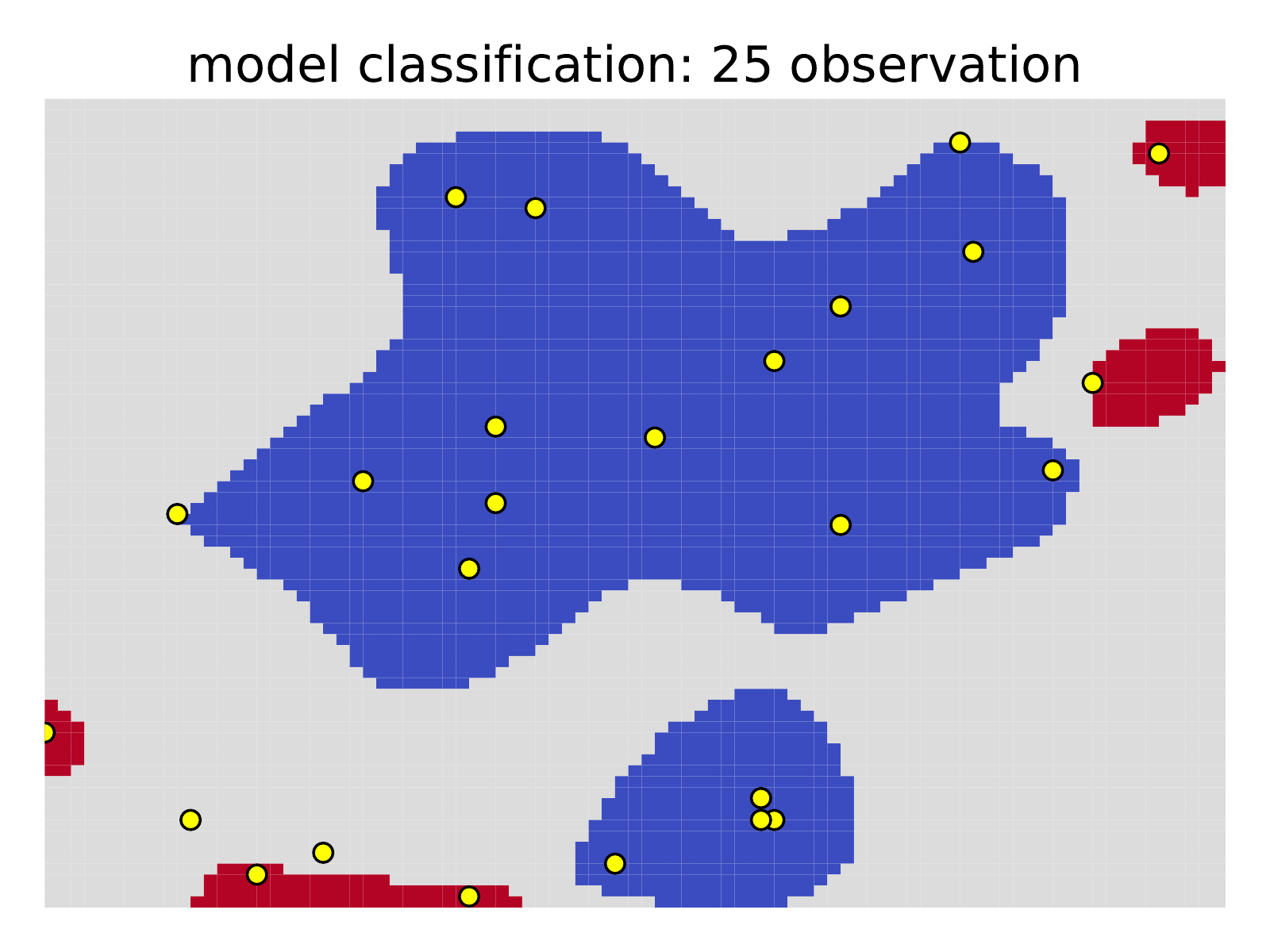} &
  \includegraphics[bb=0 0 461 346, clip, width=0.3\textwidth]{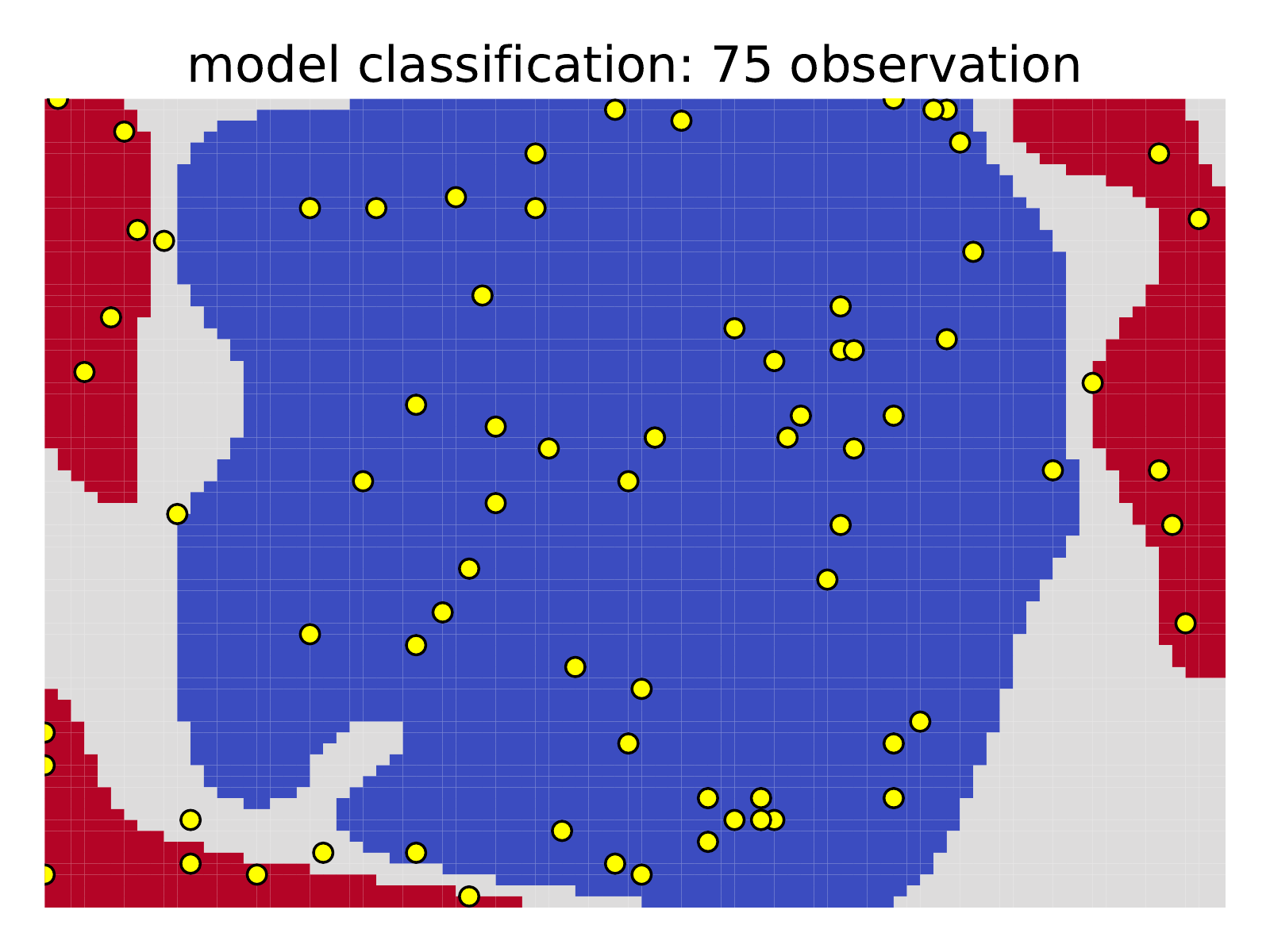} \\
 \end{tabular}
 {\tt random} method
\end{center}

\noindent
{\bf b}
\begin{center}
 \begin{tabular}{cccc}
  \includegraphics[bb=0 0 461 346, clip, width=0.3\textwidth]{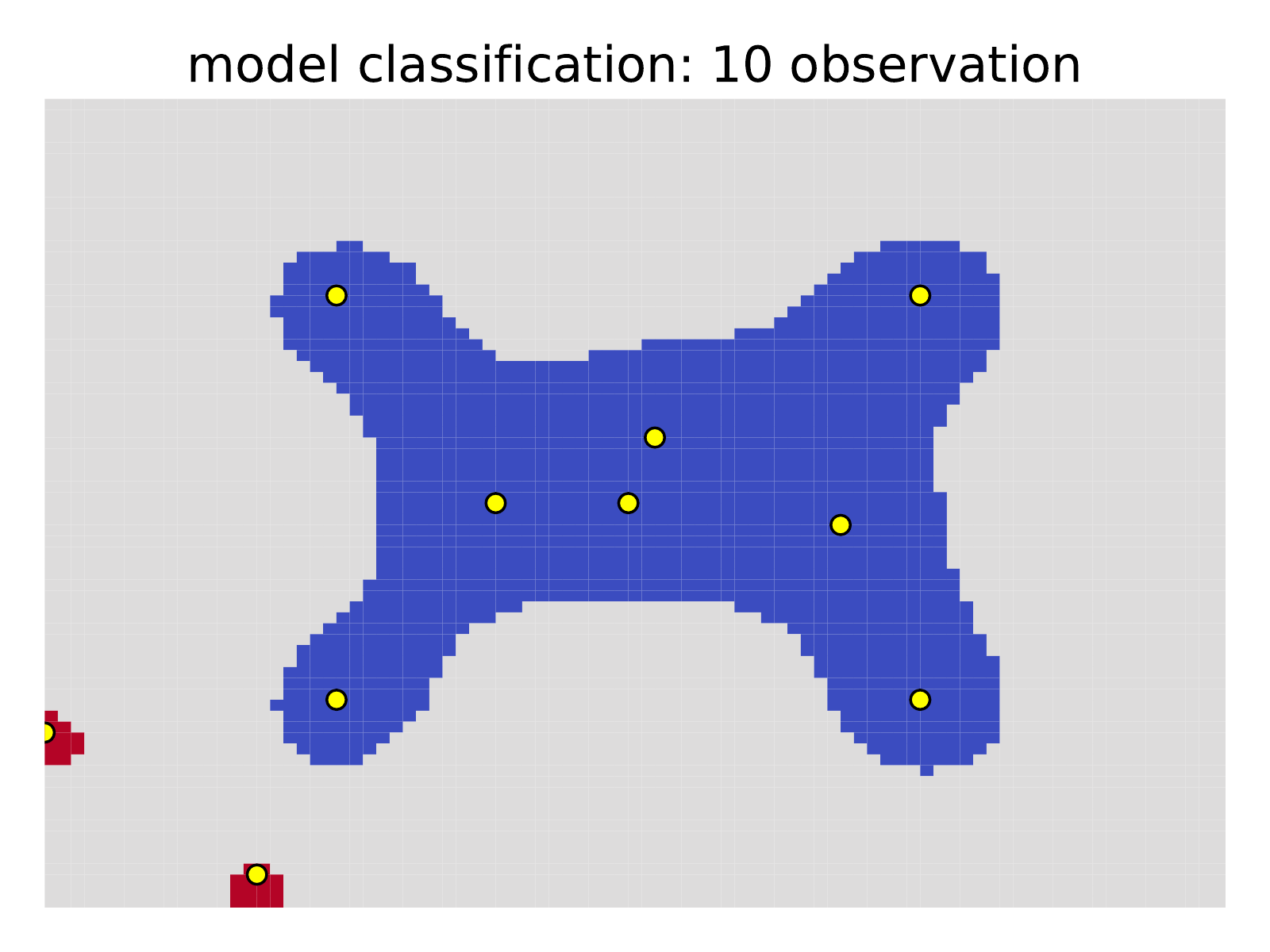} &
  \includegraphics[bb=0 0 461 346, clip, width=0.3\textwidth]{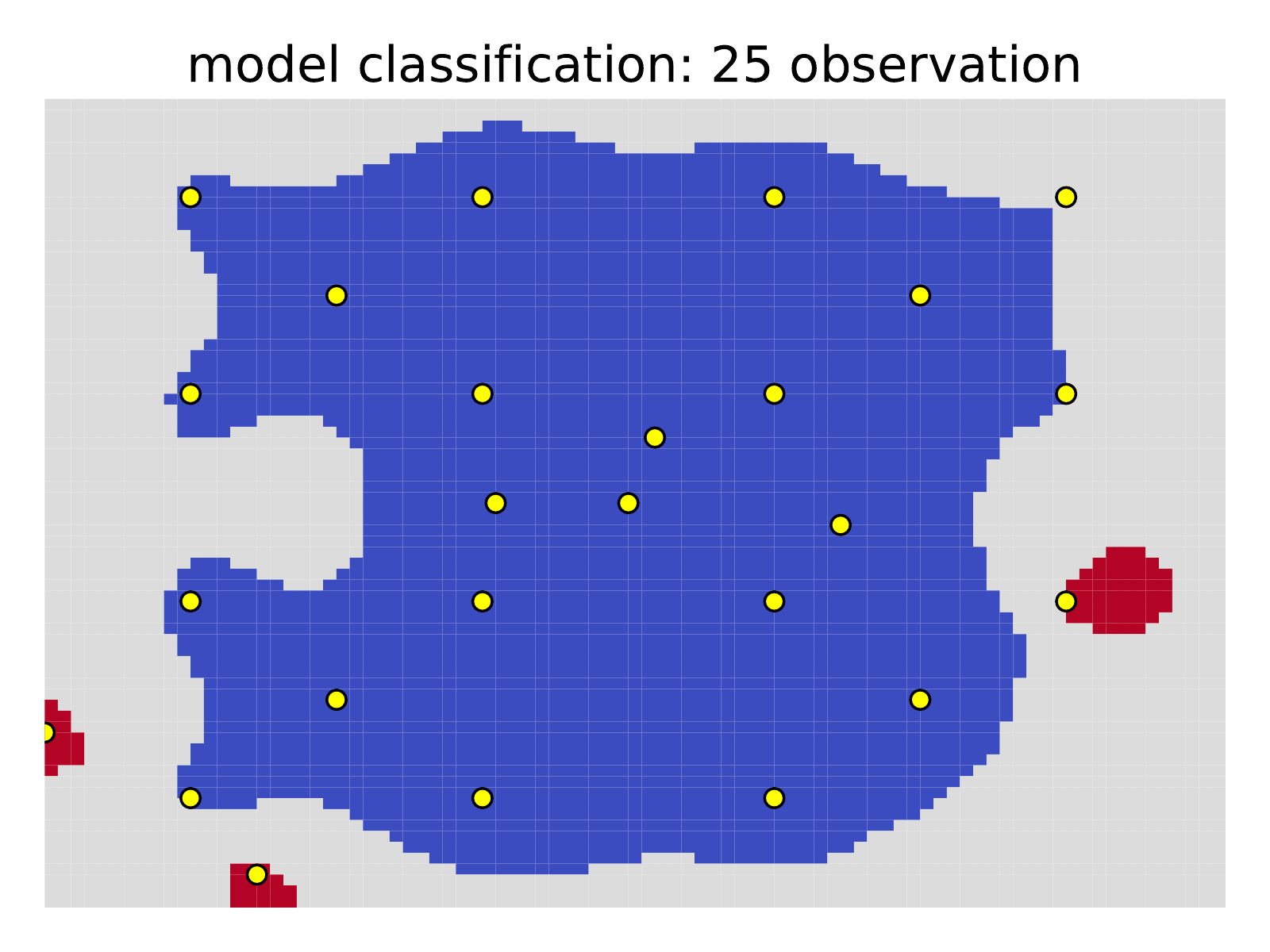} &
  \includegraphics[bb=0 0 461 346, clip, width=0.3\textwidth]{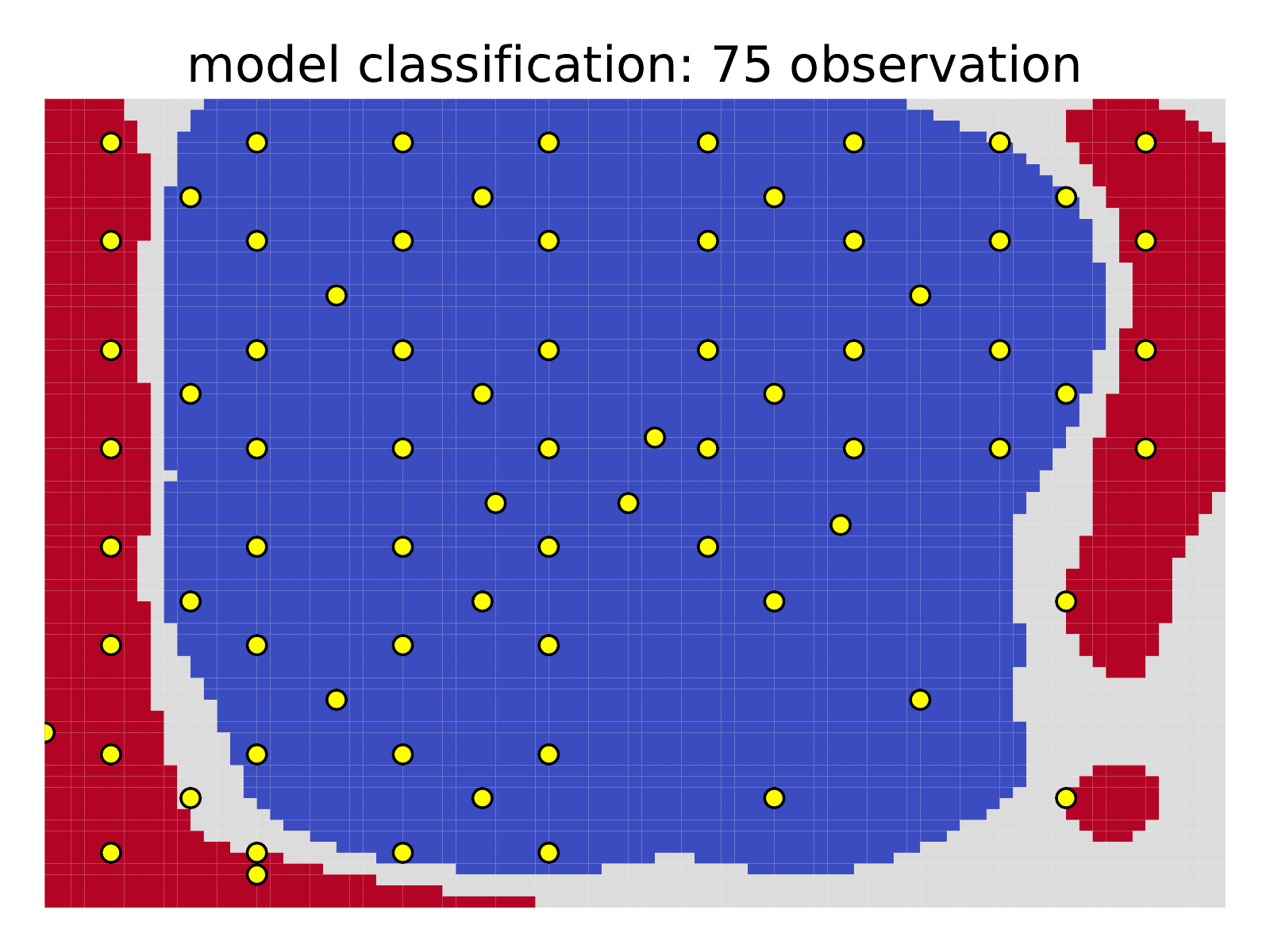} \\
 \end{tabular}
 {\tt non-adaptive} method
\end{center}

\noindent
{\bf c}
\begin{center}
 \begin{tabular}{cccc}
  \includegraphics[bb=0 0 461 346, clip, width=0.3\textwidth]{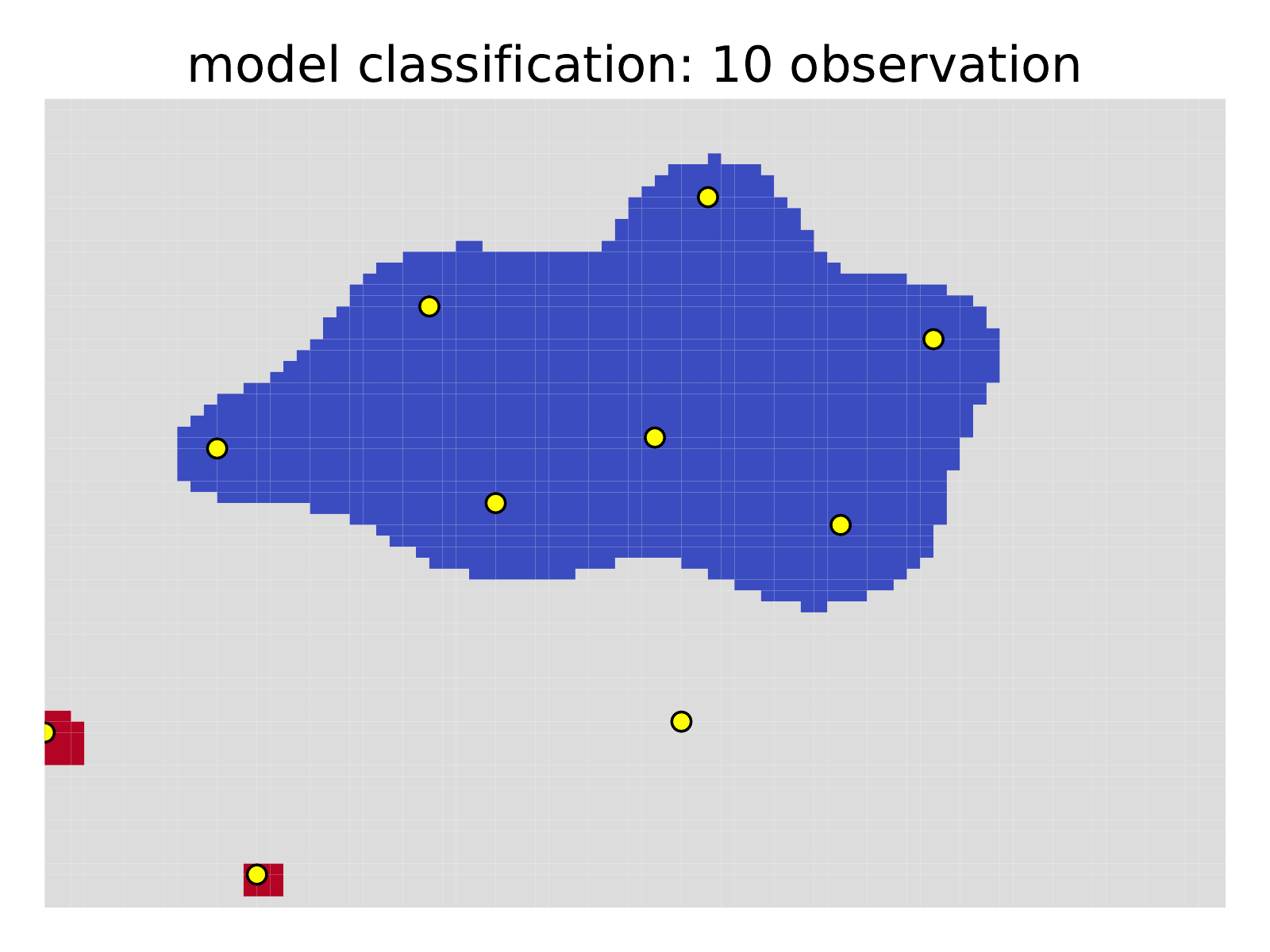} &
  \includegraphics[bb=0 0 461 346, clip, width=0.3\textwidth]{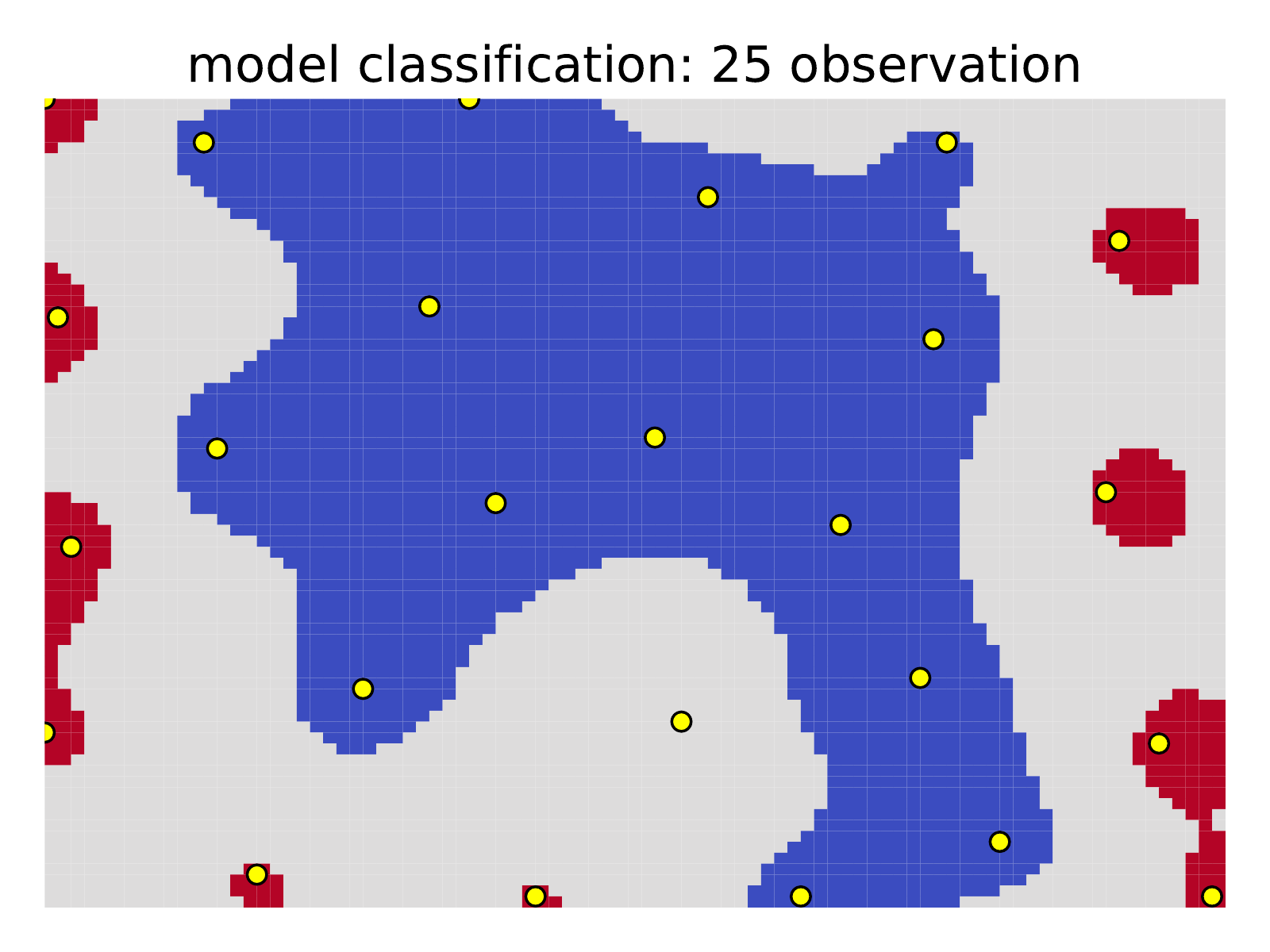} &
  \includegraphics[bb=0 0 461 346, clip, width=0.3\textwidth]{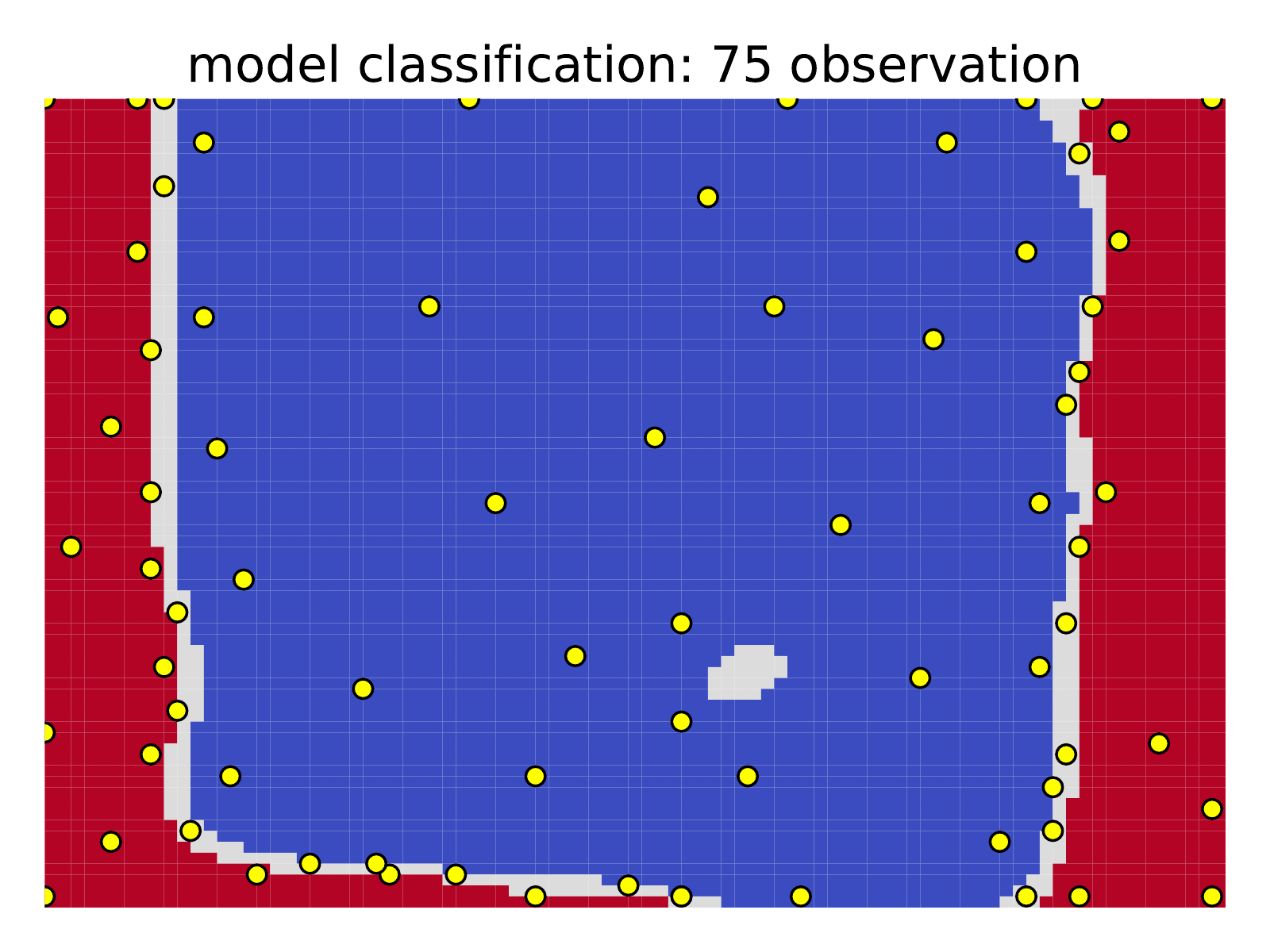} \\
 \end{tabular}
 AL-based LSE method
\end{center}

\noindent
{\bf d}
\begin{center}
 \begin{tabular}{cccc}
  \includegraphics[bb=0 0 461 346, clip, width=0.3\textwidth]{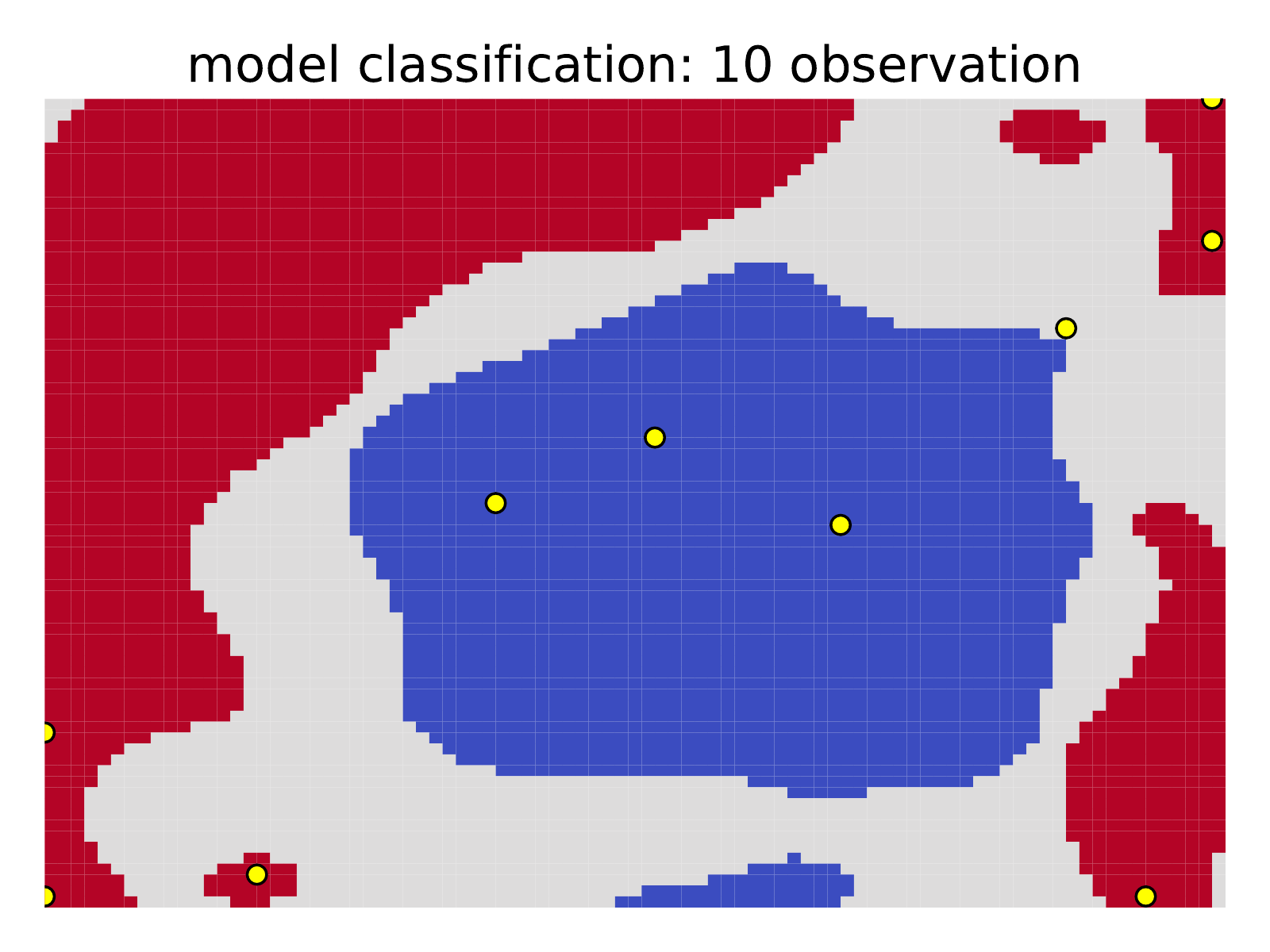} &
  \includegraphics[bb=0 0 461 346, clip, width=0.3\textwidth]{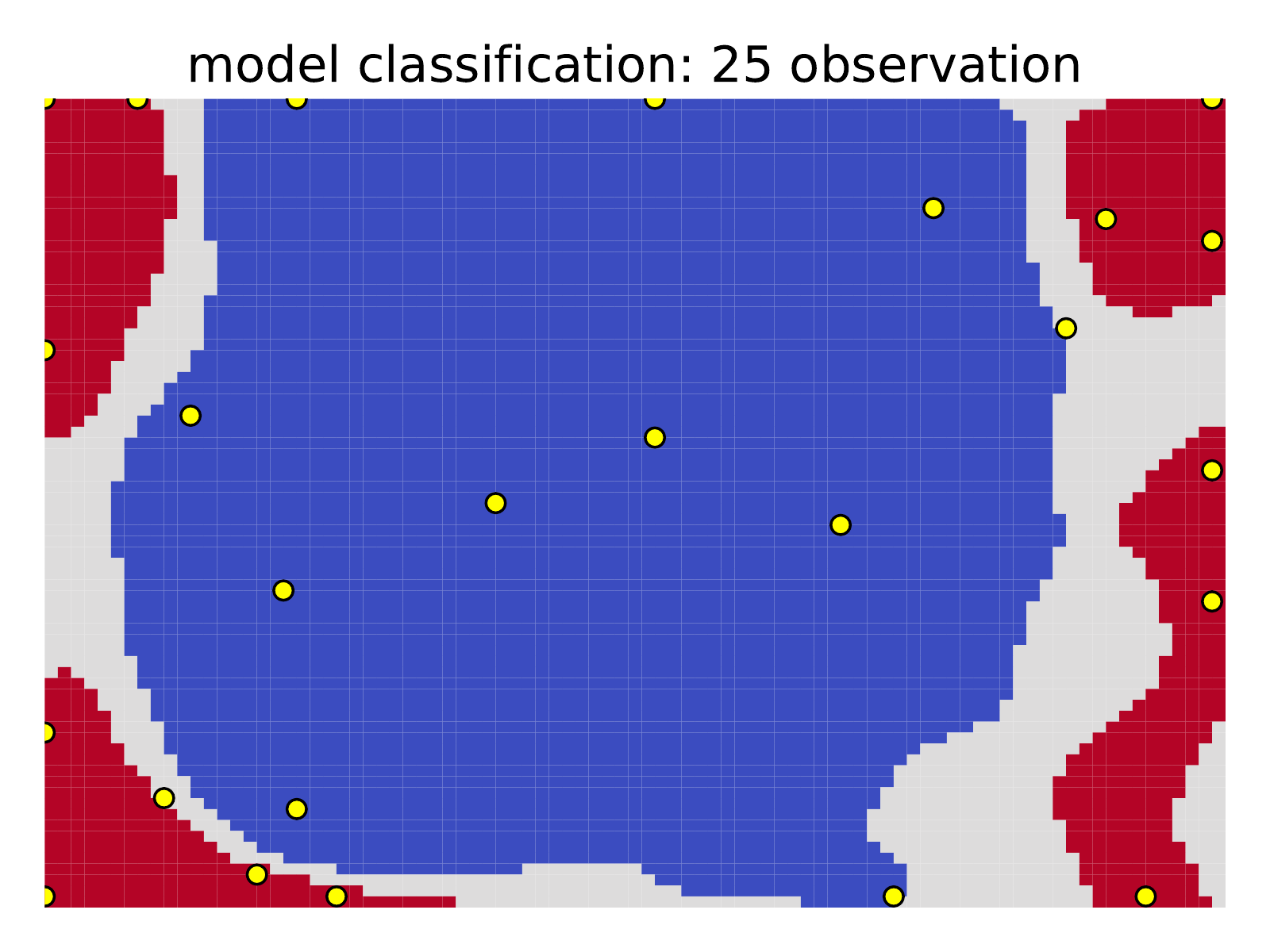} &
  \includegraphics[bb=0 0 461 346, clip, width=0.3\textwidth]{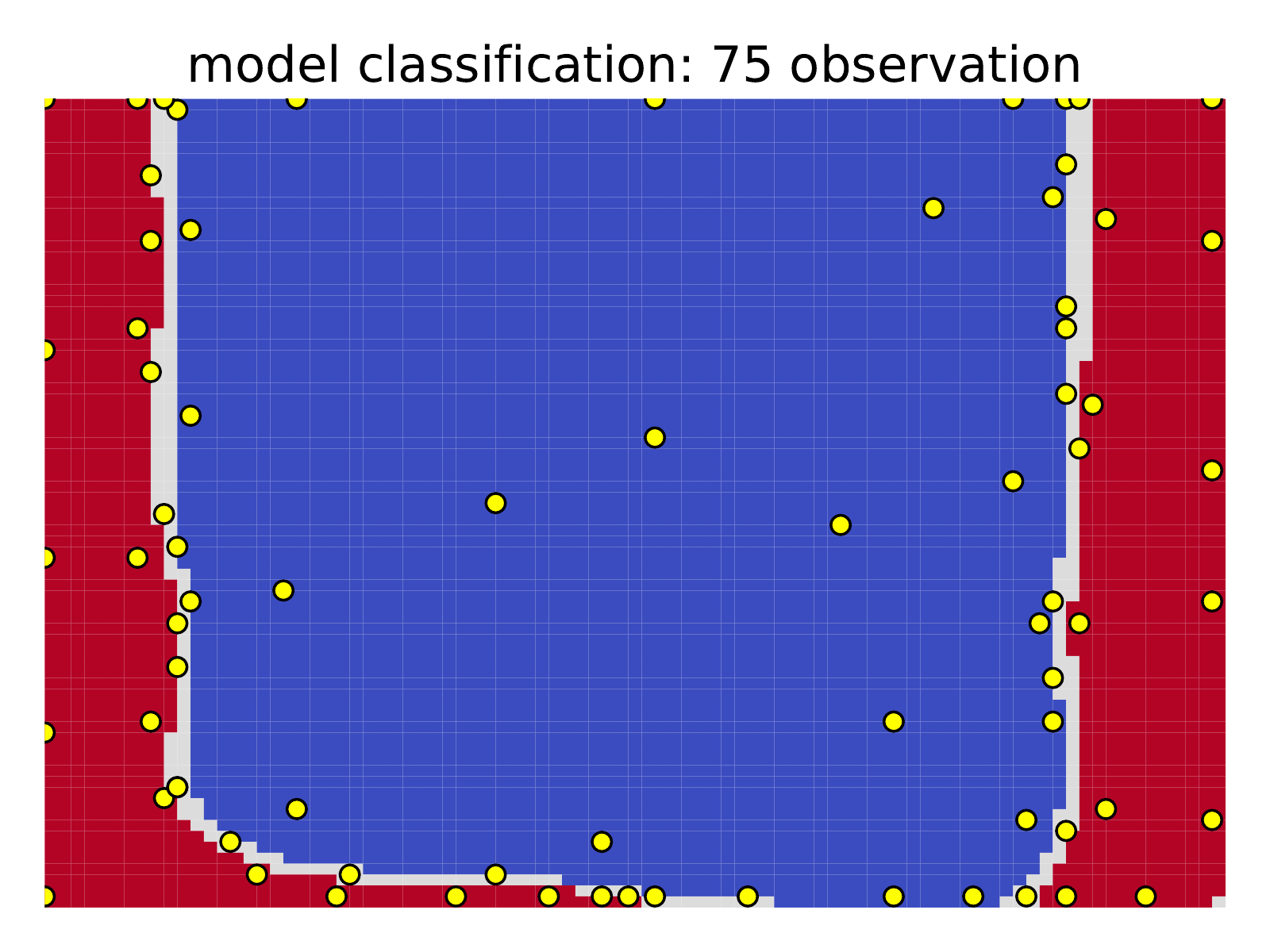} \\
 \end{tabular}
 ATL-based LSE method
\end{center}

\noindent
{\bf e}
\begin{center}
 \begin{tabular}{cccc}
  \includegraphics[bb=0 0 461 346, clip, width=0.3\textwidth]{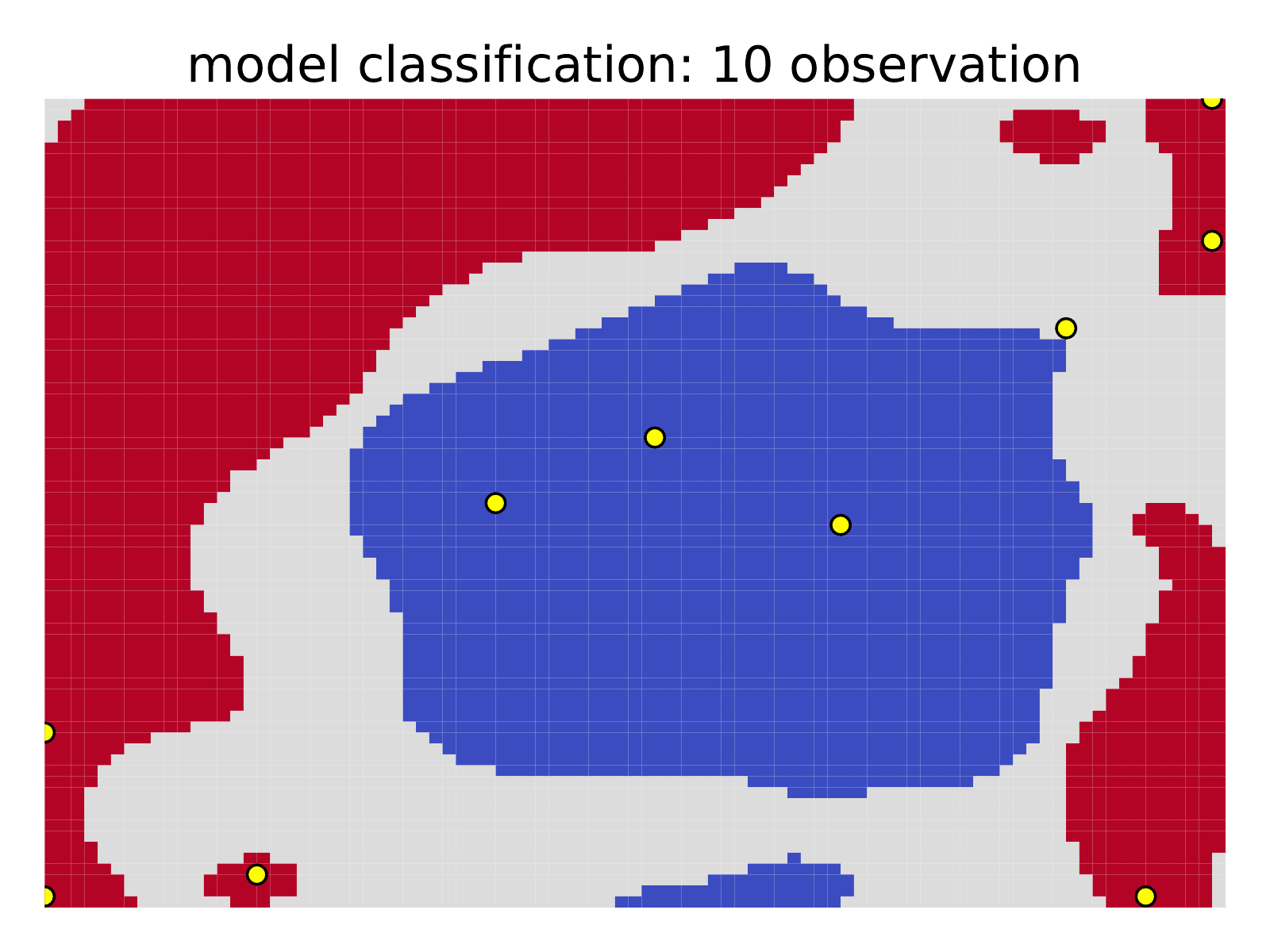} &
  \includegraphics[bb=0 0 461 346, clip, width=0.3\textwidth]{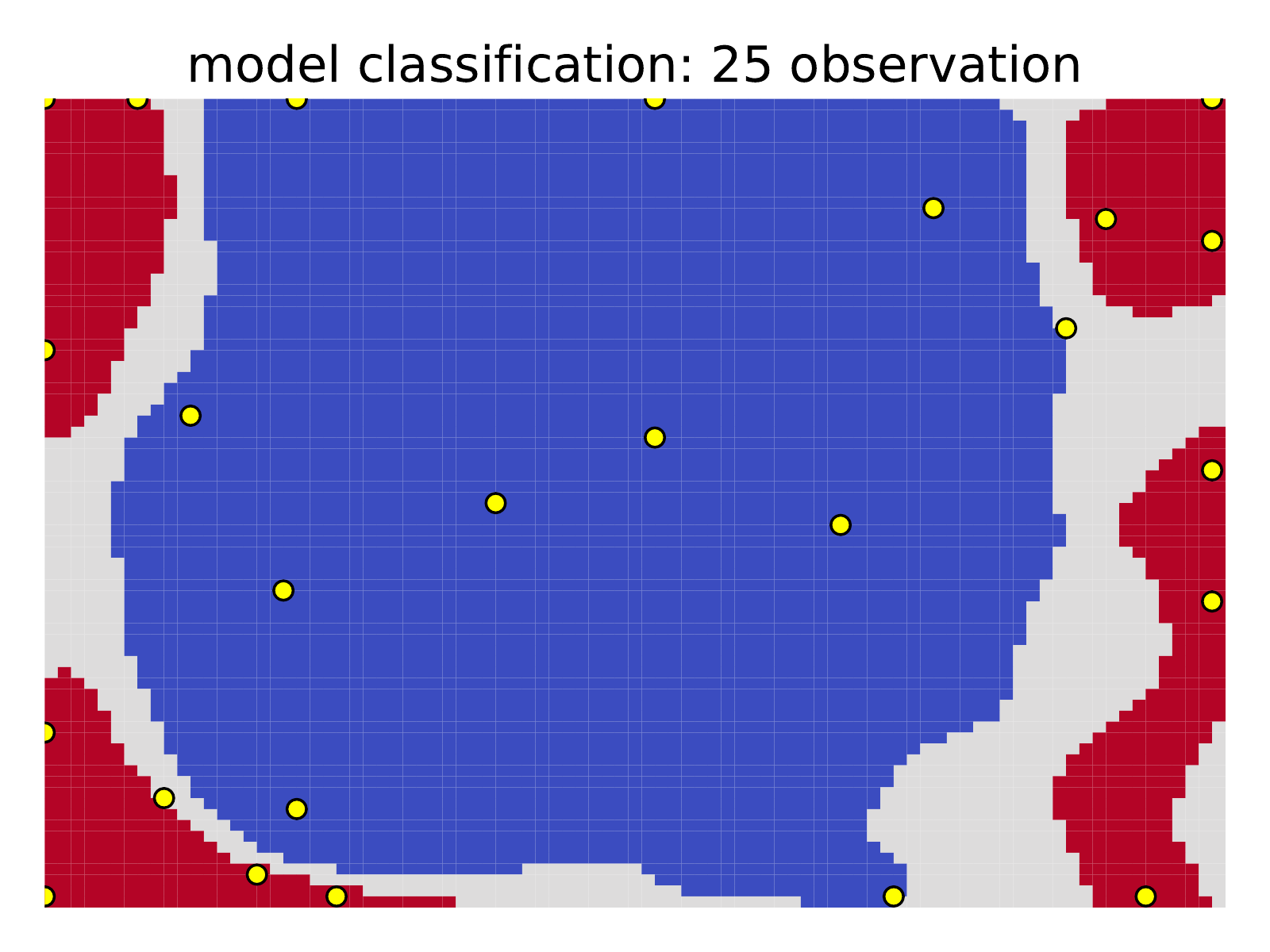} &
  \includegraphics[bb=0 0 461 346, clip, width=0.3\textwidth]{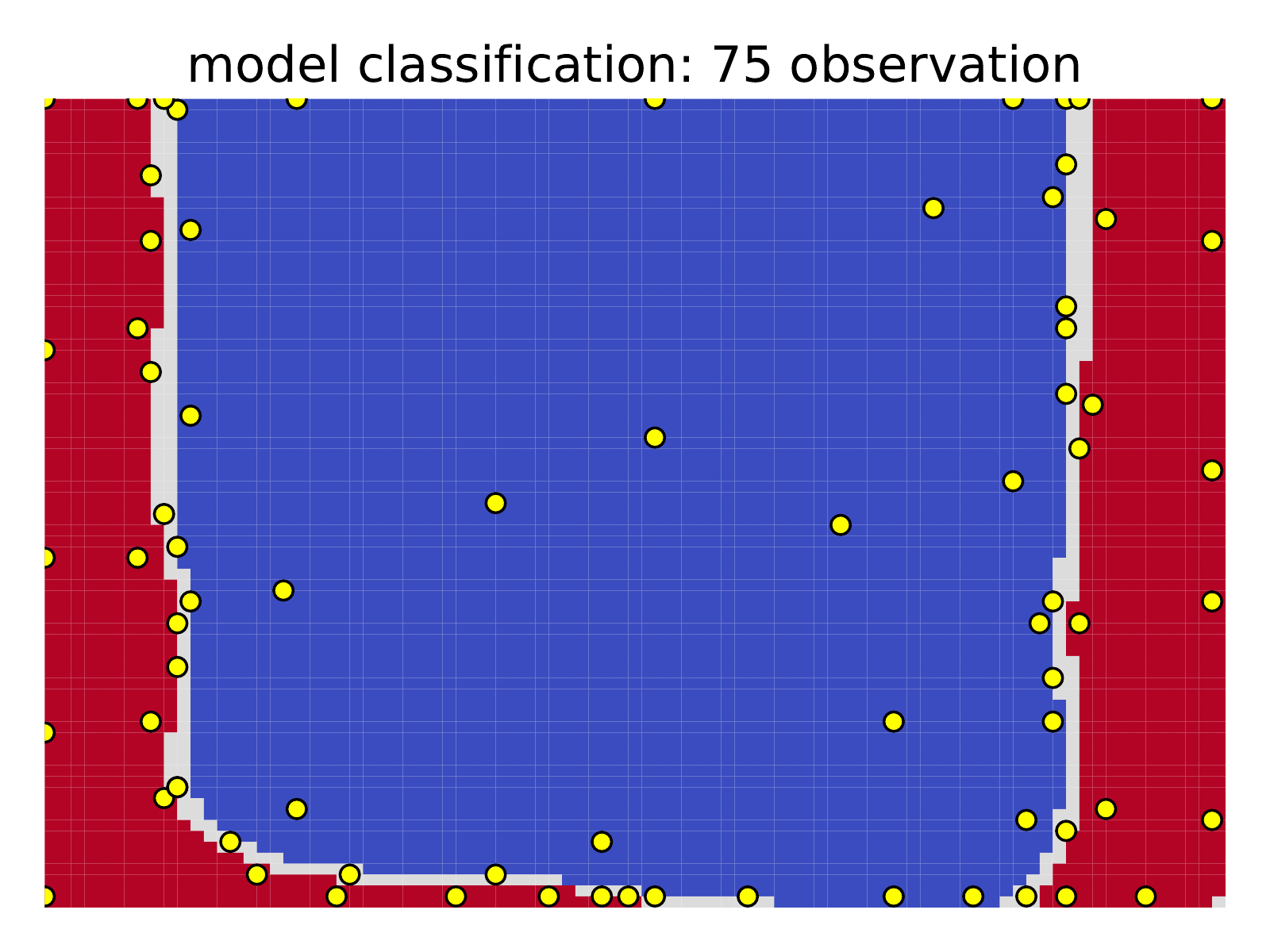} \\
 \end{tabular}
 LSS-ATL-based LSE method
\end{center}

\noindent
{\bf Figure 5}:
Comparison of two baseline methods (random and non-adaptive) and three proposed methods (AL-, ATL-, and LSS-ATL-based LSE) on the red-zone identification problem in a solar cell ingot. 
The three plots in each method indicate the measurement points and estimated super-level set (blue), sub-level set (red), and uncertain region (gray) after 10, 25, and 75 measurements were conducted.
%
\clearpage

\begin{center}
 \begin{tabular}{ll}
  {\bf a} &
  {\bf b} \\
  \includegraphics[bb=0 0 461 346, clip, width=0.5\textwidth]{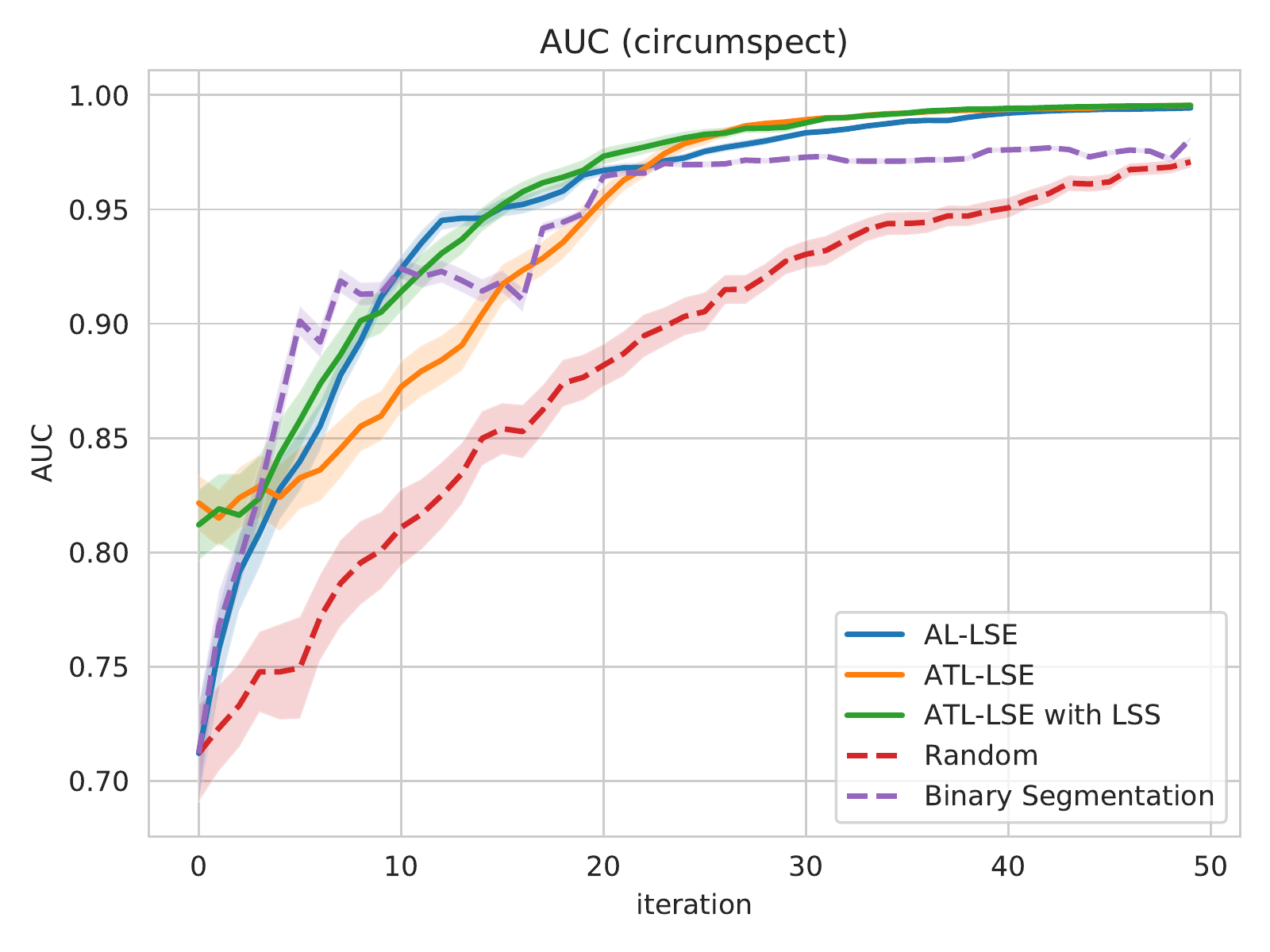} &
  \includegraphics[bb=0 0 461 346, clip, width=0.5\textwidth]{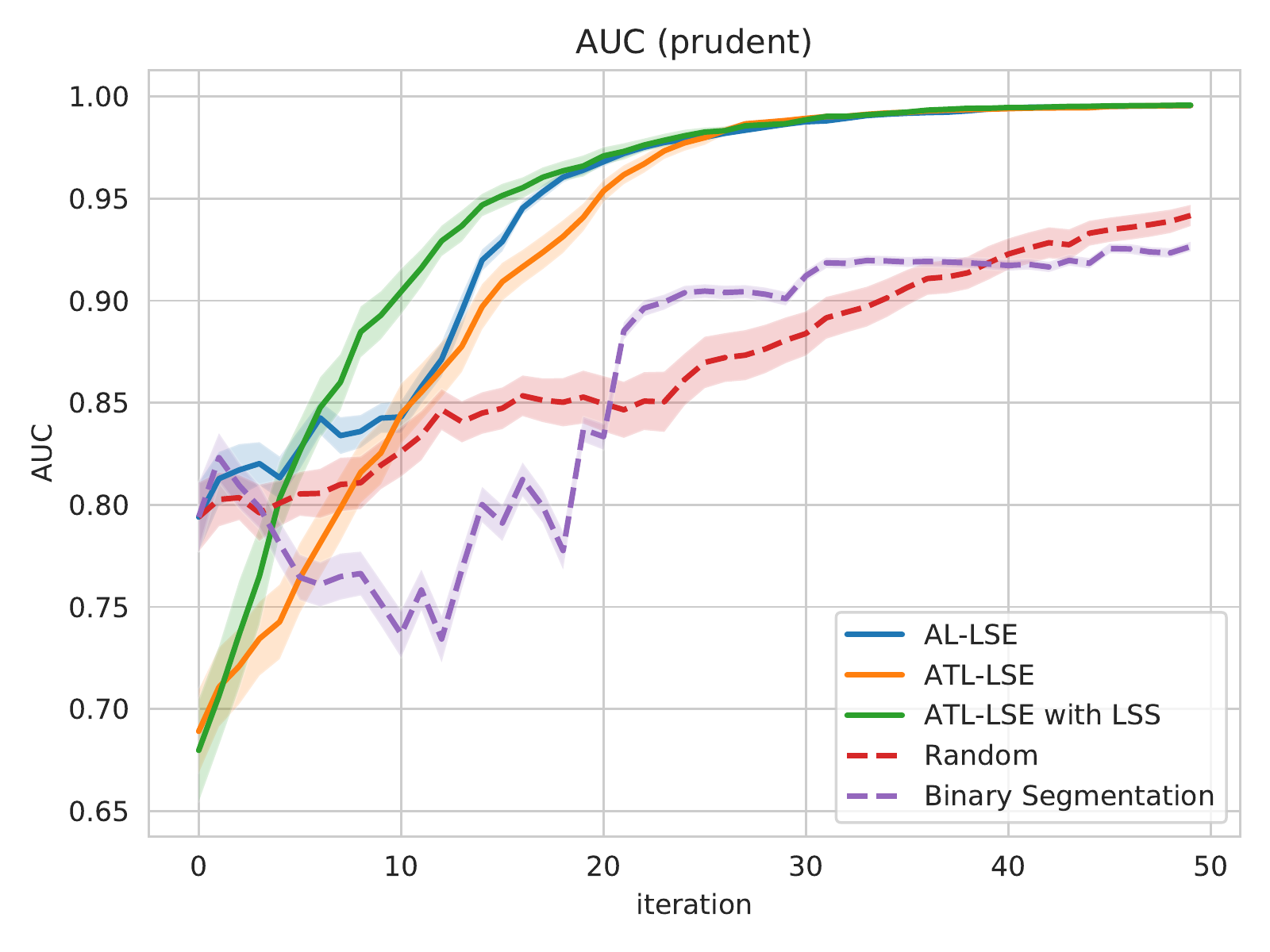} \\
  \begin{minipage}[c]{0.5\textwidth}
   \begin{center}
    risk-sensitive AUC 
   \end{center}
  \end{minipage}
  &
  \begin{minipage}[c]{0.5\textwidth}
   \begin{center}
    cost-sensitive AUC
   \end{center}
  \end{minipage}
  \\
  {\bf c} &
  {\bf d} \\
  \includegraphics[bb=0 0 461 346, clip, width=0.5\textwidth]{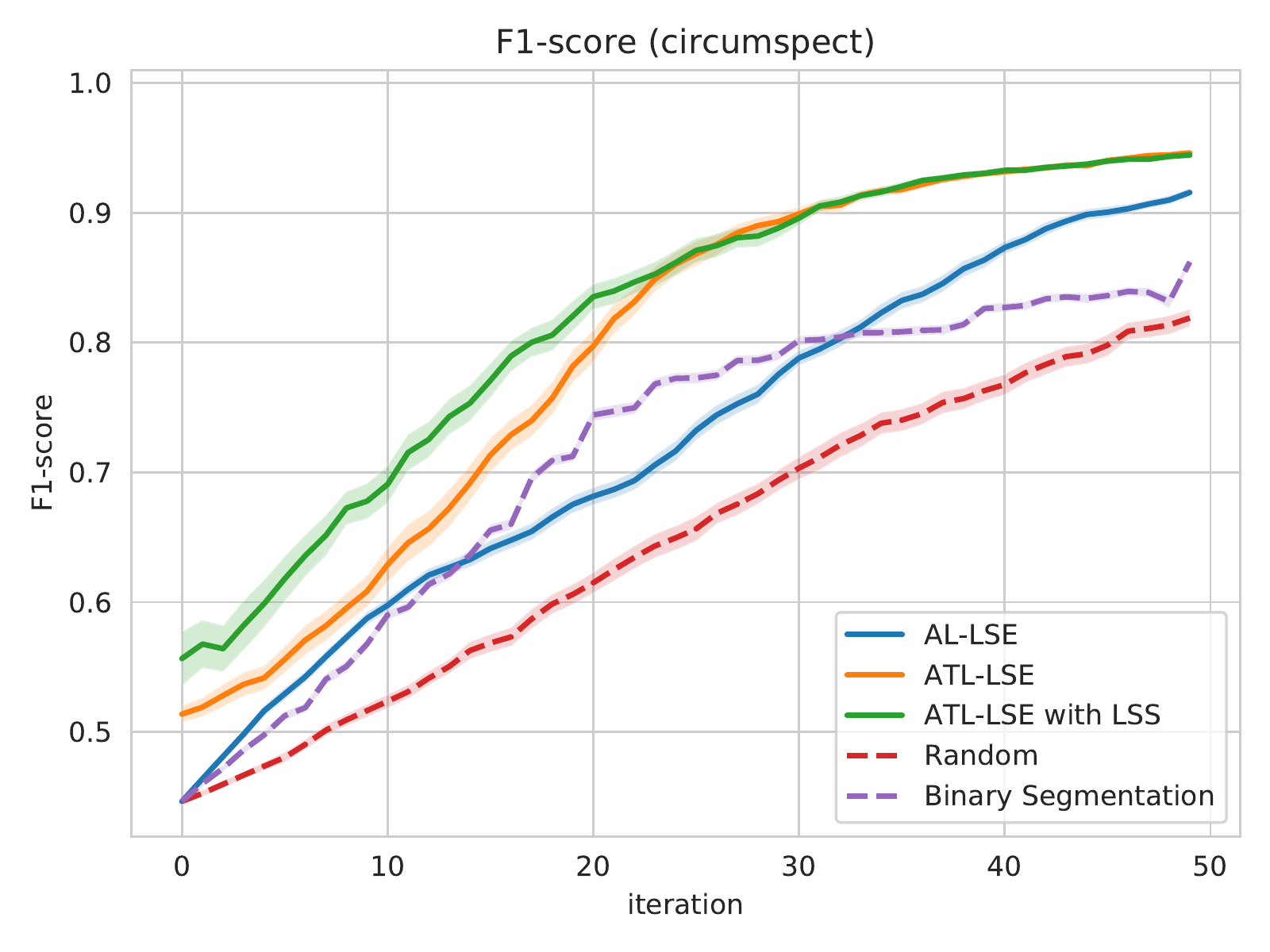} &
  \includegraphics[bb=0 0 461 346, clip, width=0.5\textwidth]{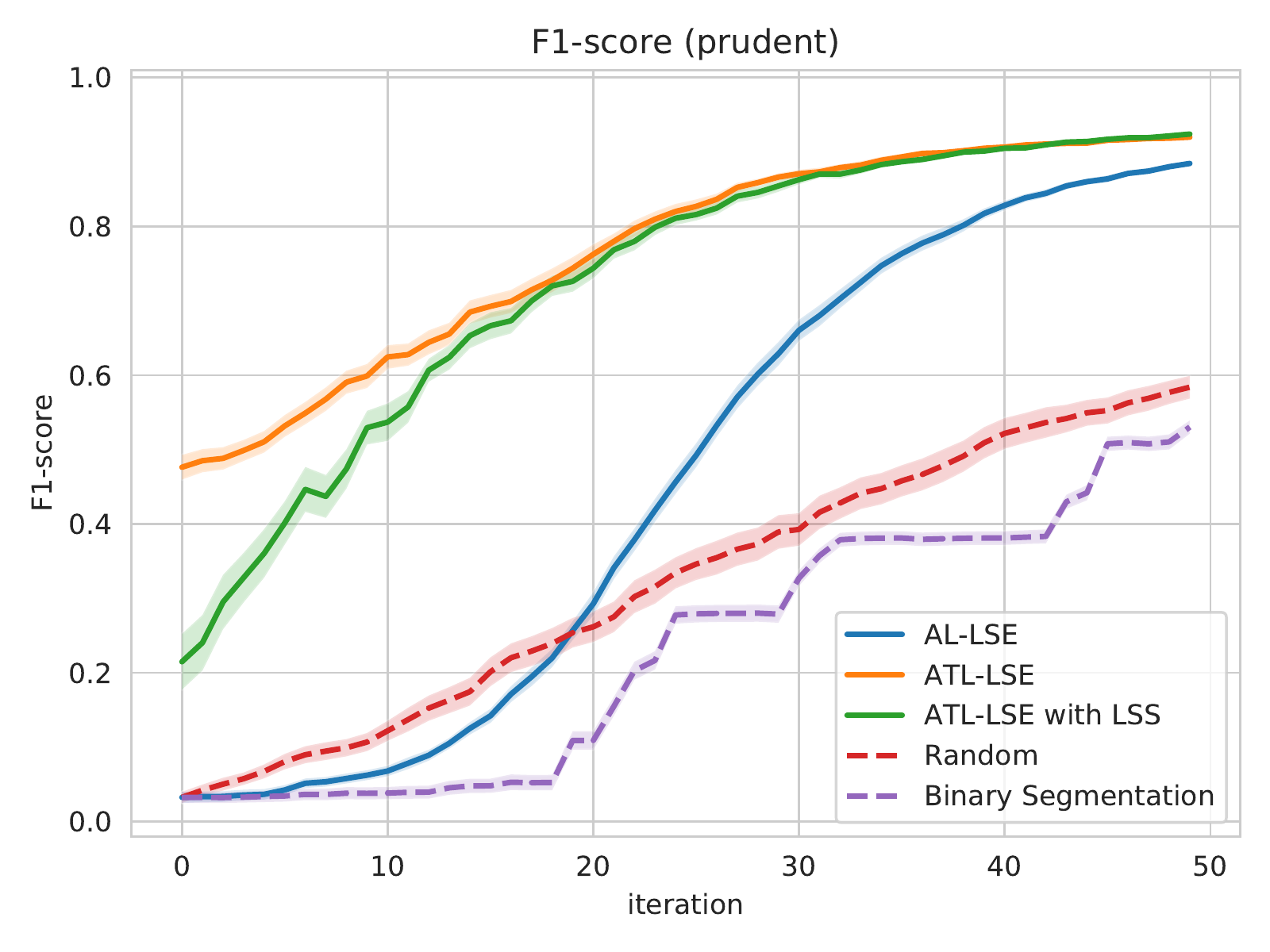} \\
  \begin{minipage}[c]{0.5\textwidth}
   \begin{center}
    risk-sensitive F1-score
   \end{center}
  \end{minipage}
  &
  \begin{minipage}[c]{0.5\textwidth}
   \begin{center}
    cost-sensitive F1-score
   \end{center}
  \end{minipage}
 \end{tabular}
\end{center}
\noindent
{\bf Figure 6}:
Comparisons of area under the curve (AUCs) and F1-scores of two baseline methods (random  and non-adaptive binary segmentation) and three proposed methods (AL-, ATL-, and LSS-ATL-based LSE) for risk- and cost-sensitive scenarios.
\clearpage

\begin{center}
 \begin{tabular}{lll}
  {\bf a} &
  {\bf b} &
  {\bf c} \\
  \includegraphics[bb=0 0 432 288, clip, width=0.3\textwidth]{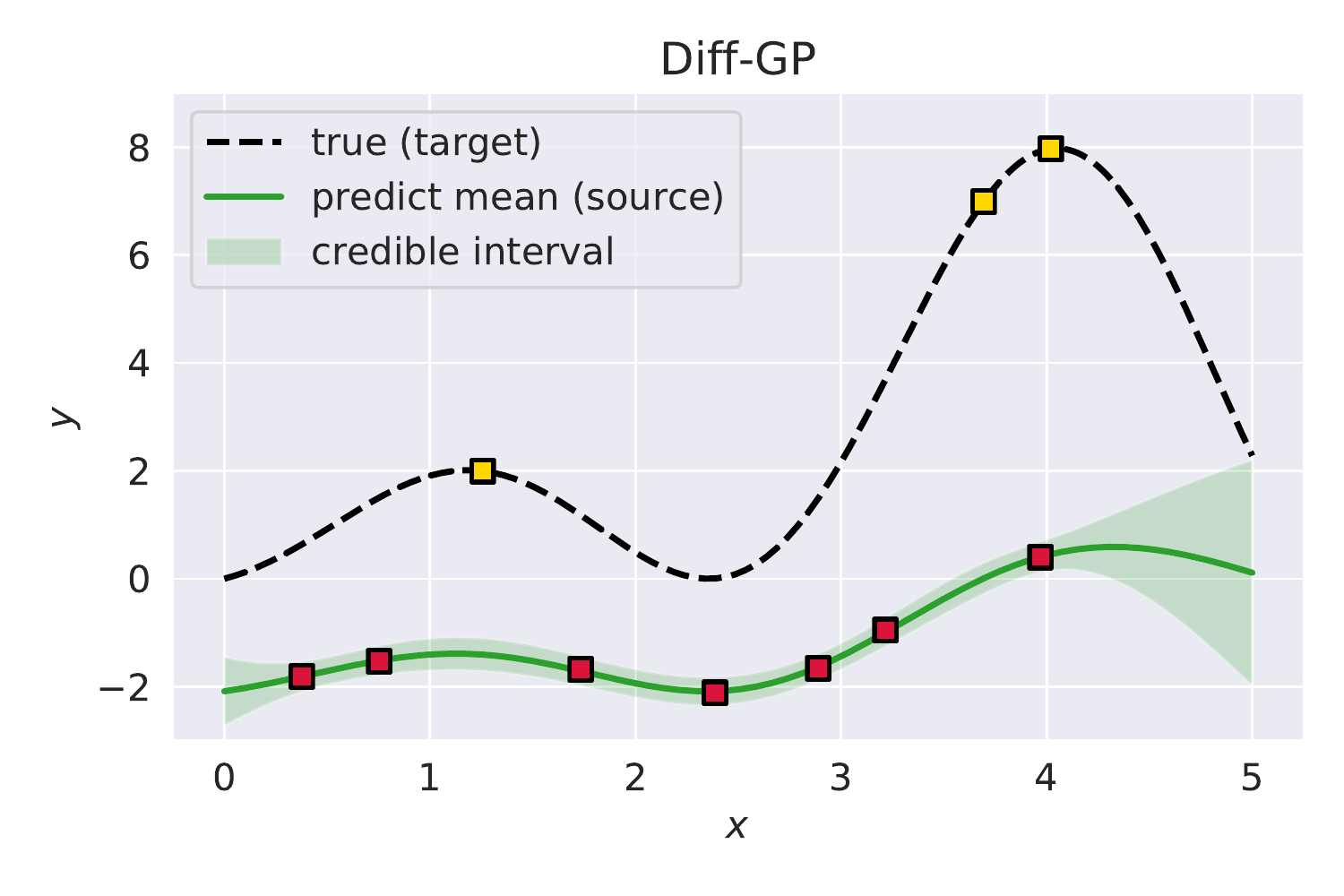} &
  \includegraphics[bb=0 0 432 288, clip, width=0.3\textwidth]{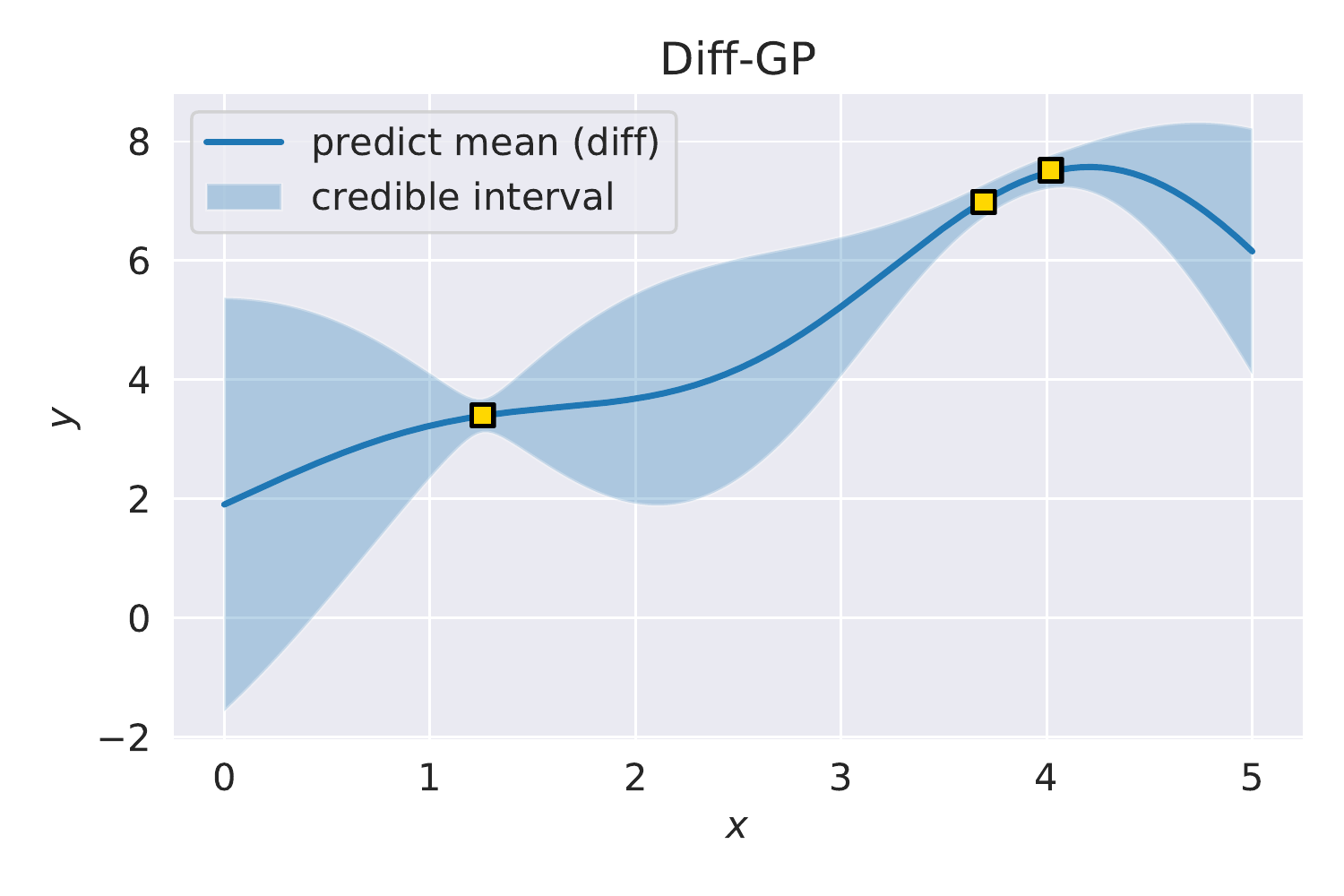} &
  \includegraphics[bb=0 0 432 288, clip, width=0.3\textwidth]{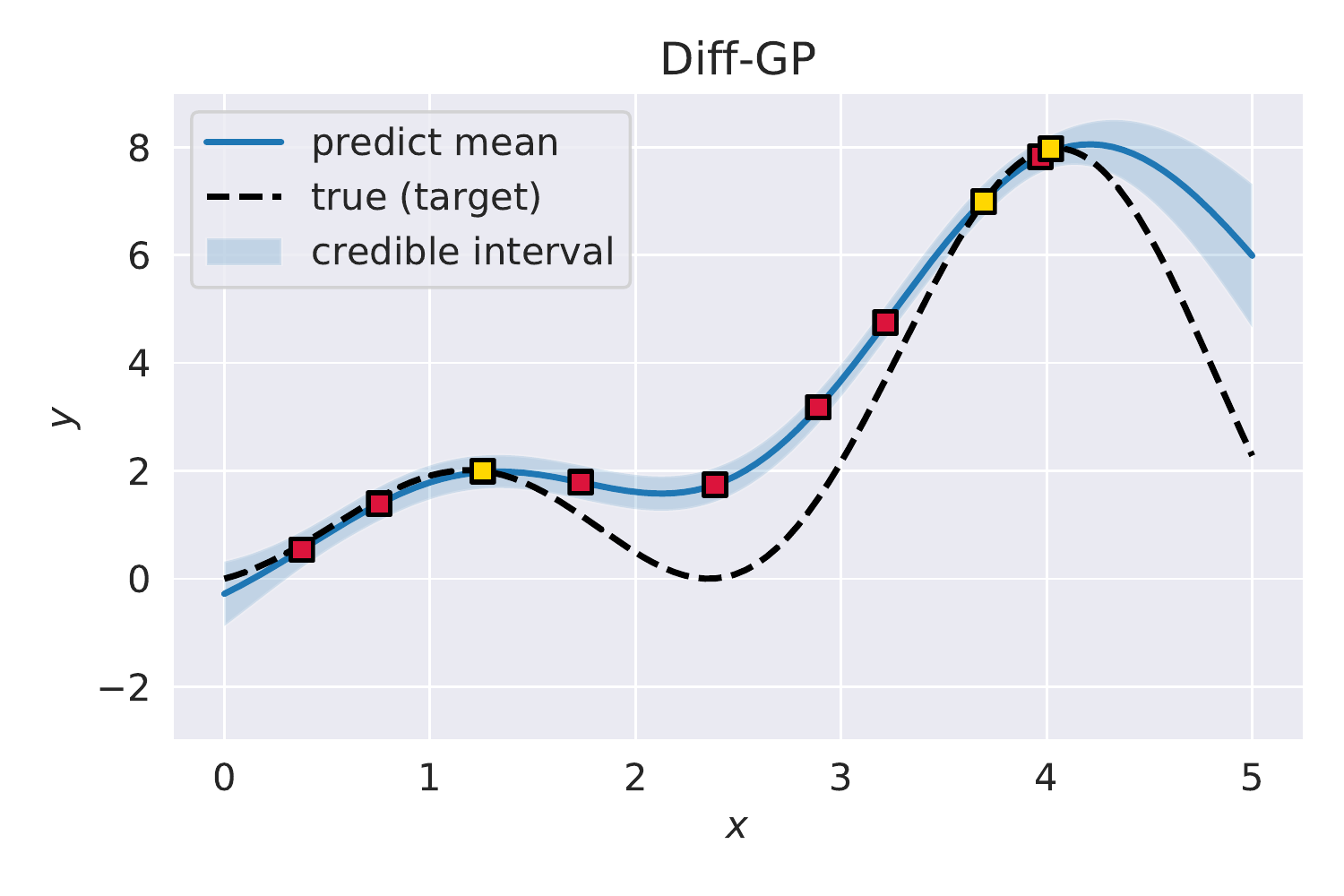} 
 \end{tabular}
\end{center}

\begin{center}
 \begin{tabular}{lll}  
  {\bf d} &
  {\bf e} &
  {\bf f} \\
  \includegraphics[bb=0 0 432 288, clip, width=0.3\textwidth]{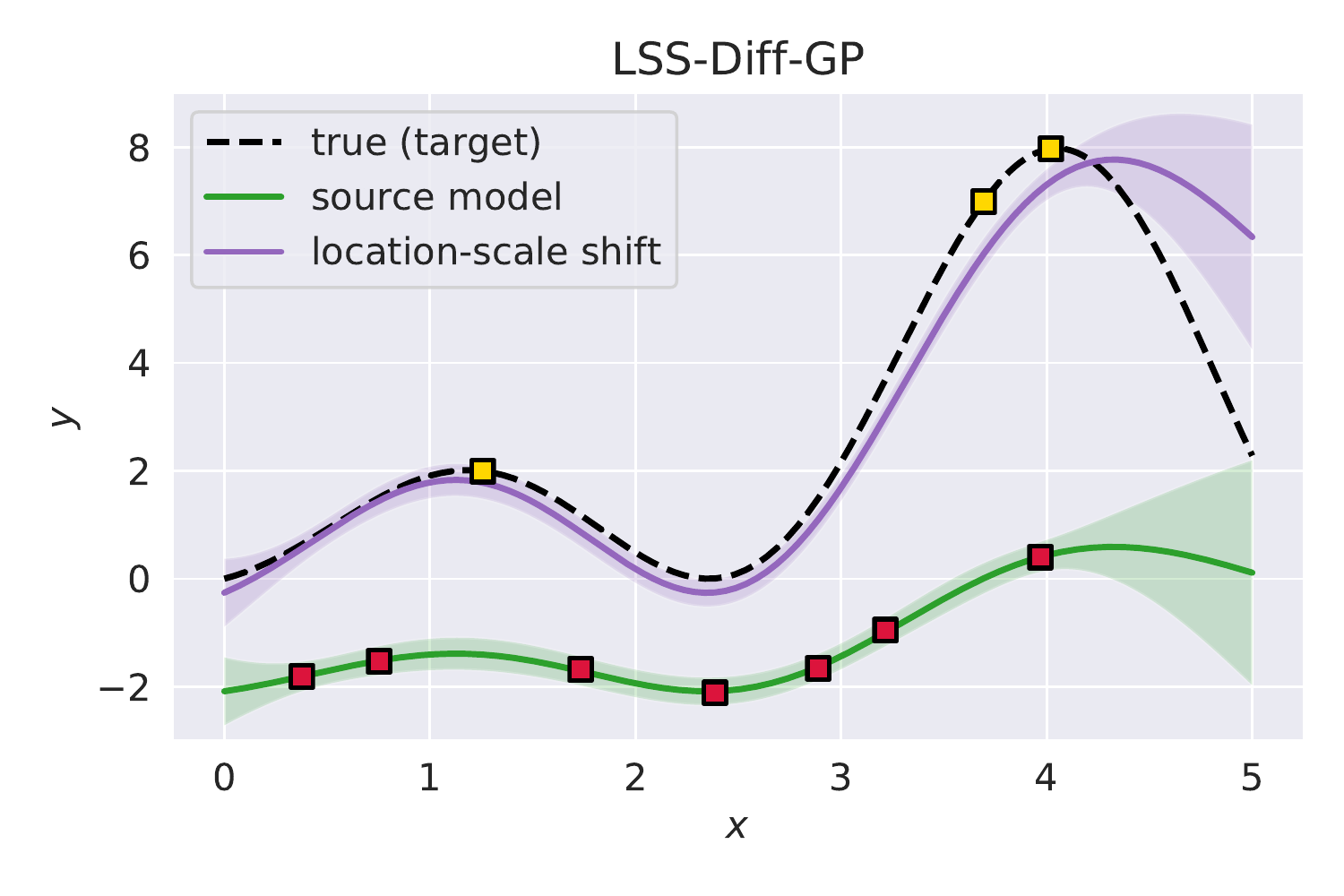} &
  \includegraphics[bb=0 0 432 288, clip, width=0.3\textwidth]{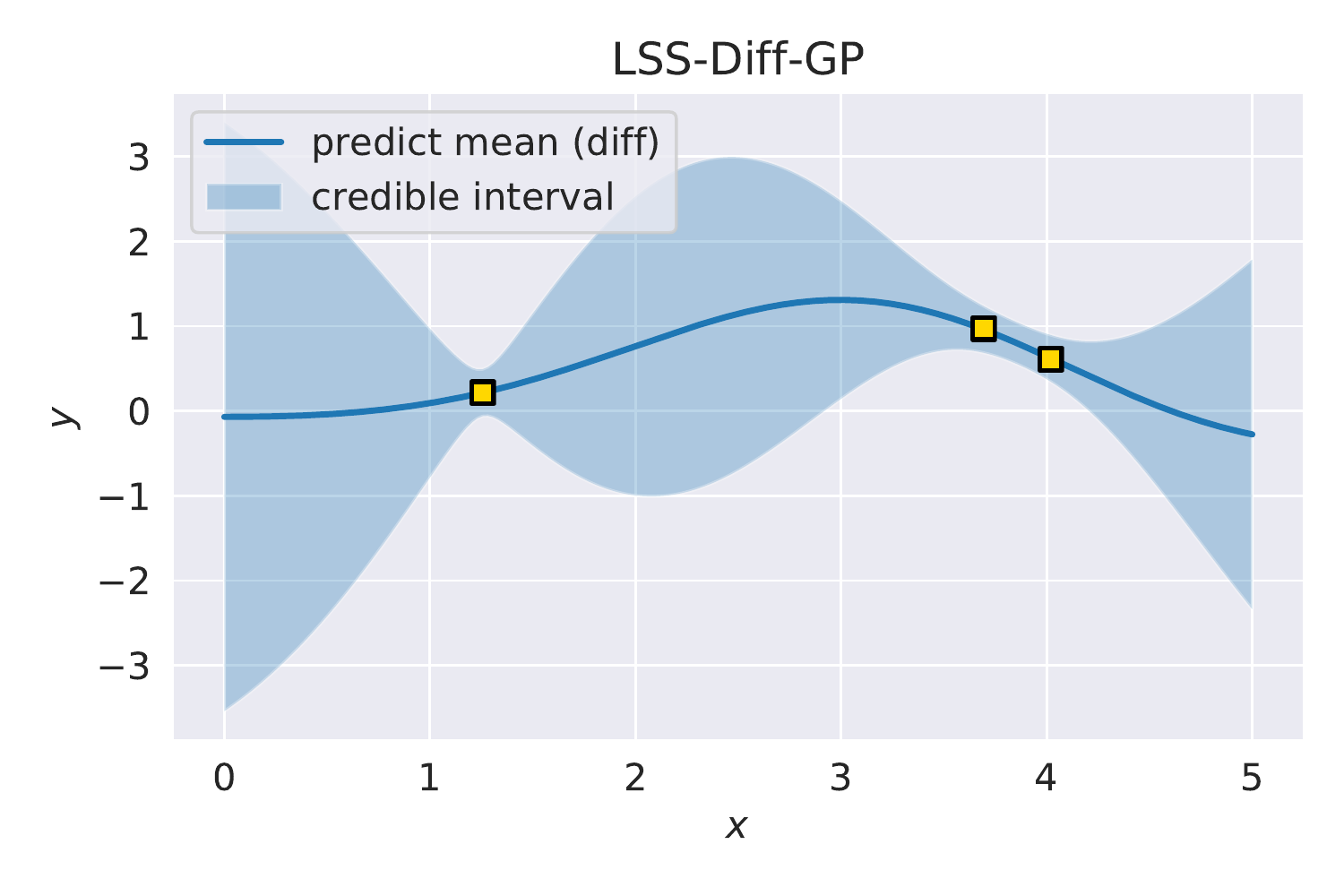} &
  \includegraphics[bb=0 0 432 288, clip, width=0.3\textwidth]{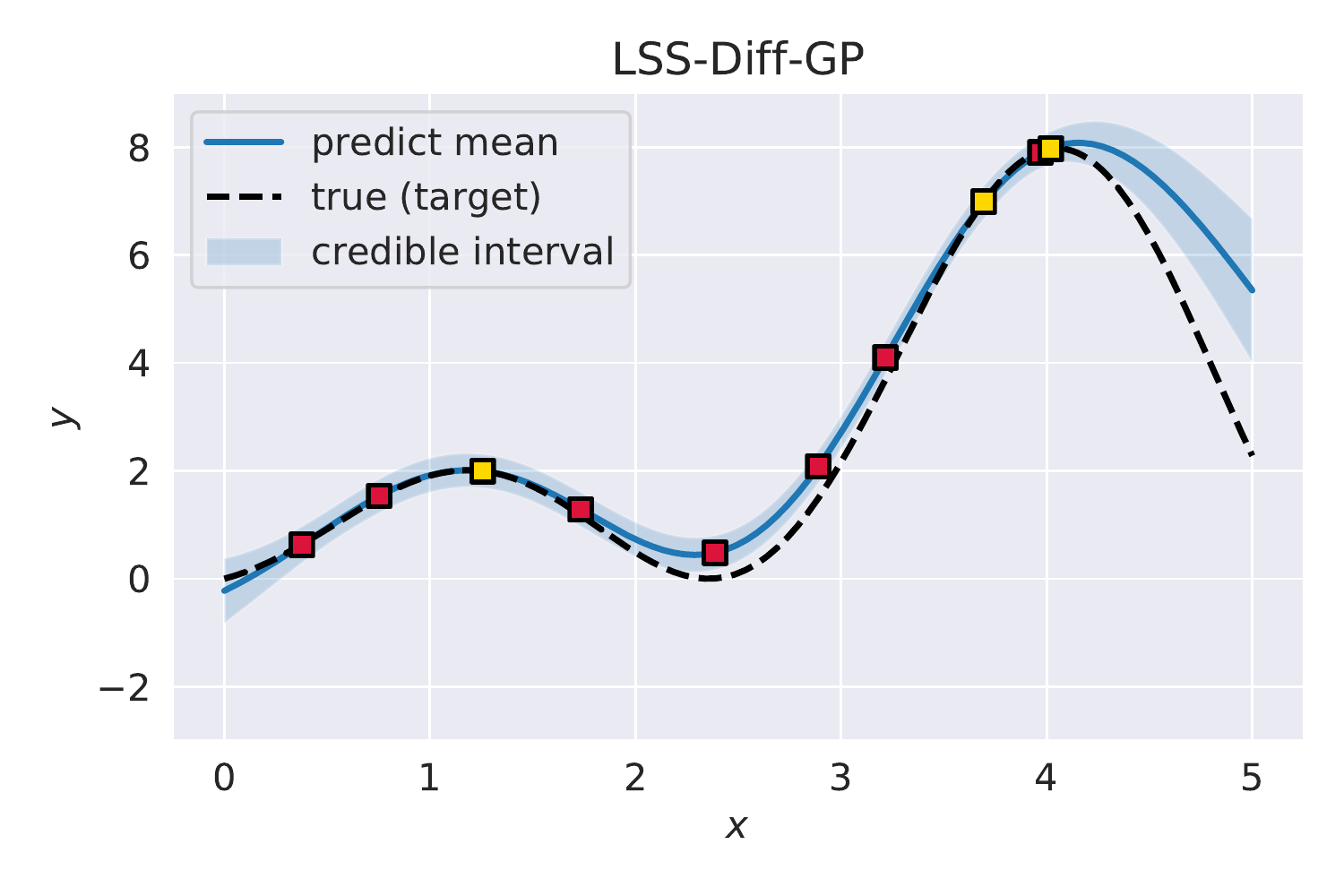} 
 \end{tabular}
\end{center}
\noindent
{\bf Figure 7}: 
Basic notions of ATL-based LSE methods ({\bf a}--{\bf c}) and examples of location-scale shifts ({\bf d}--{\bf f}). 
{\bf a}
Observations and individual GP model for the source domain (green).
{\bf b}
GP model for the difference between target and source functions.
{\bf c}
GP model for the target domain and knowledge transferred from the source domain. 
{\bf d}
Observations and individual GP model for the source domain (green), and function after location scale shift transformation (purple). 
{\bf e}
GP model for the difference between location-scale-shift target and source functions. 
{\bf f}
GP model for the target domain under the location-scale-shift, and knowledge transferred from the source domain. 

\clearpage

\begin{center}
 \begin{tabular}{lll}  
  {\bf 1} &
  {\bf 2} &
  {\bf 3} \\
  \includegraphics[bb=0 0 490 365, clip, width=0.3\textwidth]{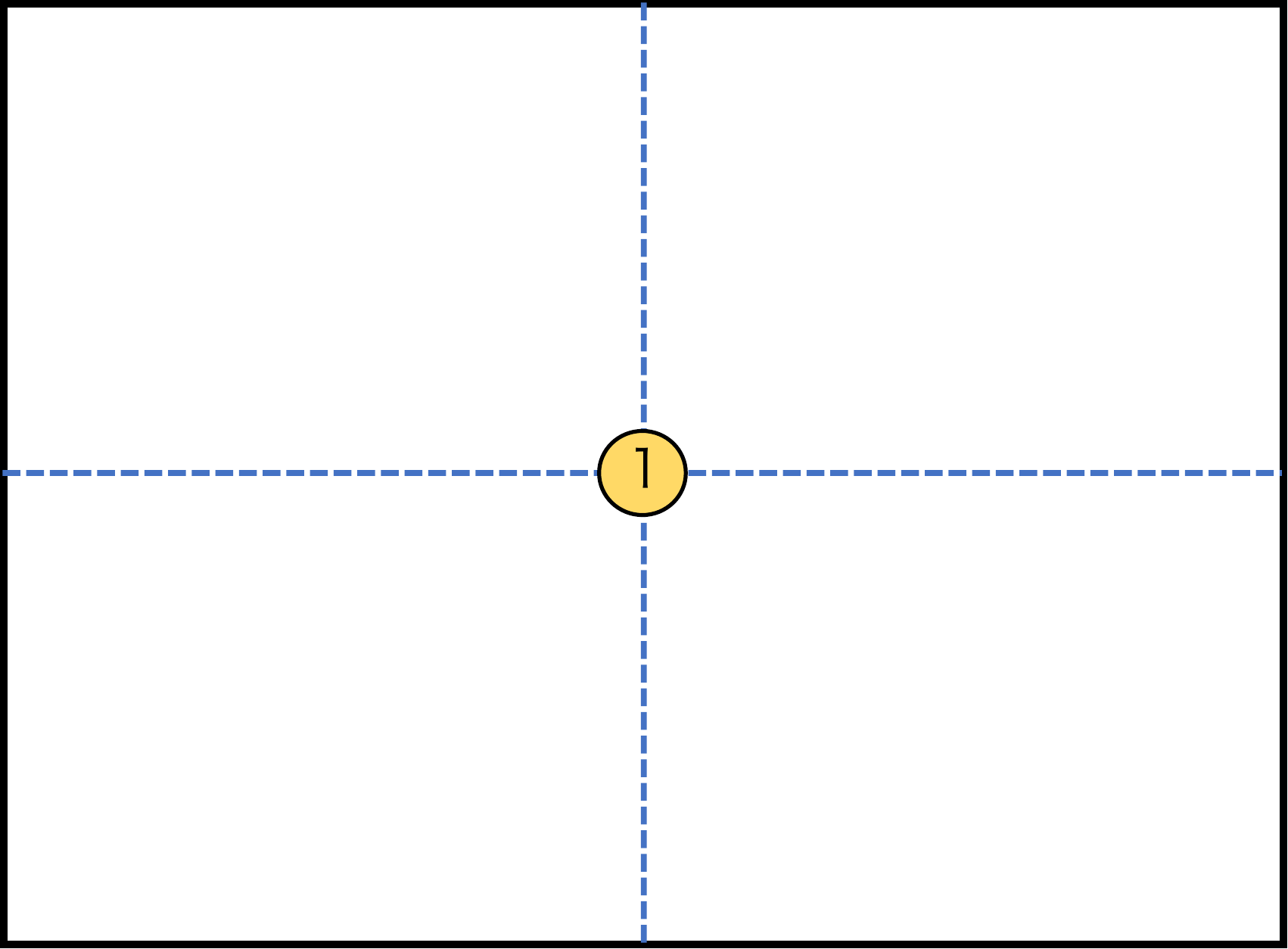} &
  \includegraphics[bb=0 0 490 365, clip, width=0.3\textwidth]{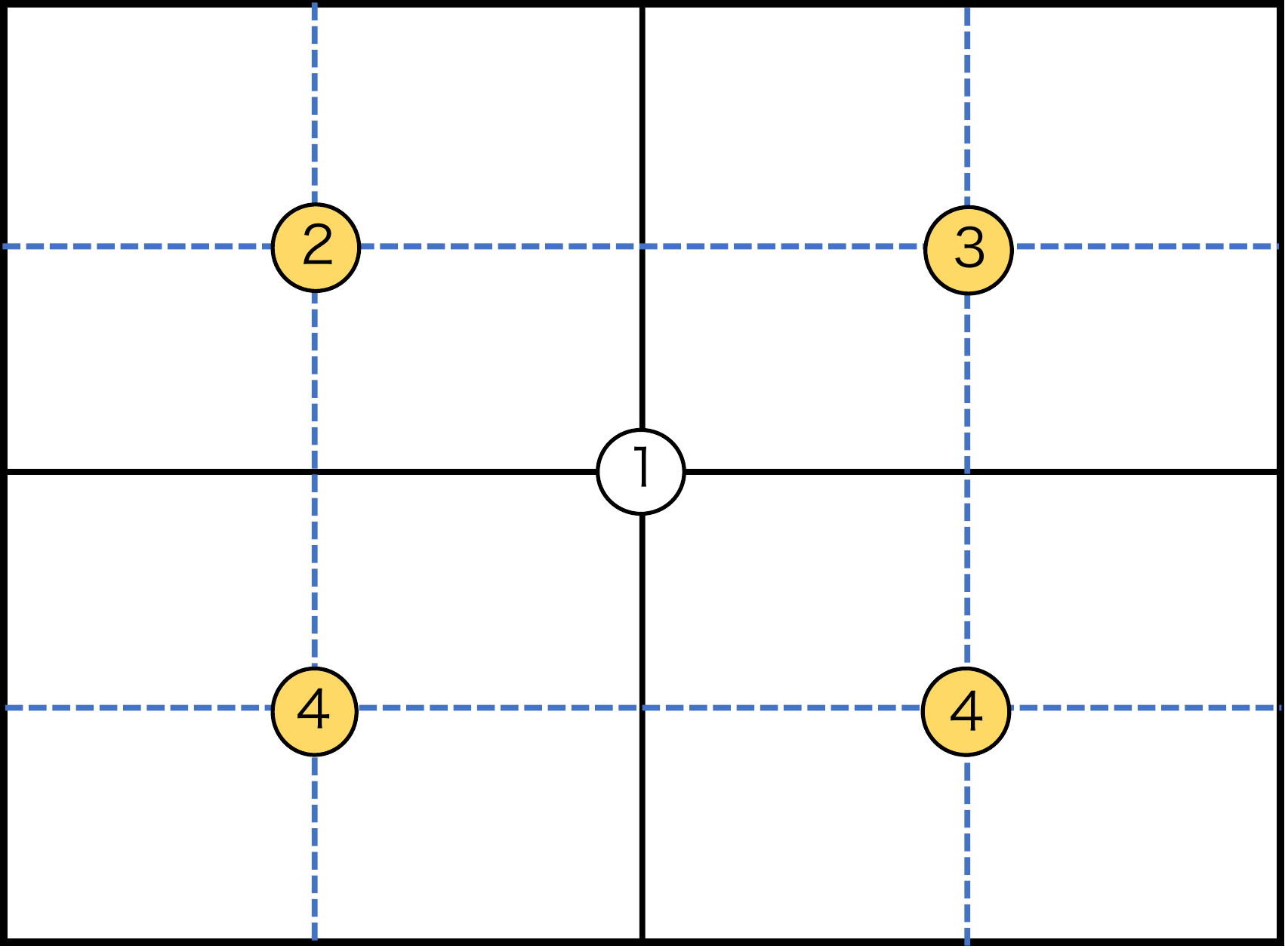} &
  \includegraphics[bb=0 0 490 365, clip, width=0.3\textwidth]{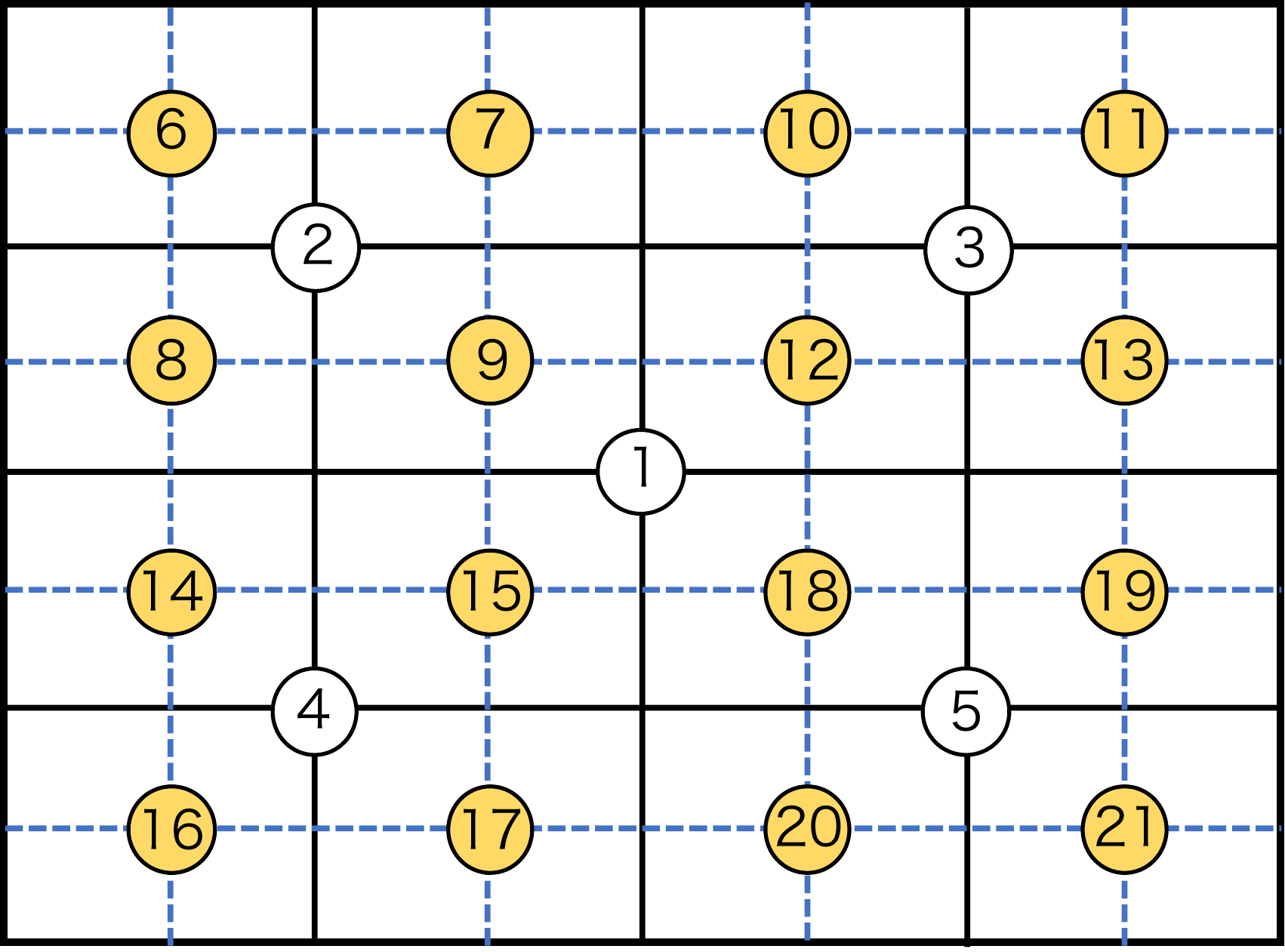} 
 \end{tabular}
\end{center}
\noindent
{\bf Supplementary Figure 1}
Illustration of NON-ADAPTIVE method. 


\clearpage
\appendix

\section*{Supplementary Information 1: Details of the Active Learning (AL) Algorithms for LSE}
\label{sec:Supp1}

\begin{algorithm}[t]
\caption{Active Learning-based Level Set Estimation (AL-based LSE)}
\label{alg}
\begin{algorithmic}
\REQUIRE observed data $\mathcal{D}_t$, GP prior, a set of candidate points $\mathcal{X}$, 
max iteration $T$
\ENSURE $\mathcal{U}_{\theta}, \mathcal{L}_{\theta}, \mathcal{C}_{\theta}$
\STATE {\bfseries Initialize:} $\mathcal{U}_{\theta}, \mathcal{L}_{\theta} \leftarrow \emptyset$, $\mathcal{C}_{\theta} \leftarrow \mathcal{X}$, $t \leftarrow 0$
\FOR{$t = 1, 2, ..., T$}
\STATE{\bfseries Step~1} Run the Gaussian process regression to obtain the predictive mean $\mu_t(\x)$ and predictive variance $\sigma_t(\x)$ defined in~(\ref{eq:pred_dist}). 
\STATE{\bfseries Step~2} Define the confidence region as $Q_t(\x) = [\mu_t(\x) - 1.96\sigma_t(\x), \mu_t(\x) + 1.96\sigma_t(\x)]$
\STATE{\bfseries Step~3} Classify all candidate points $\x \in \mathcal{X}$ as follows: 
				\begin{align*}
					\begin{cases}
						\mathcal{U}_{\theta} \leftarrow \mathcal{U}_{\theta} \cup \{\x\} & \mbox{ if }  \mu_t(\x) - 1.96 \sigma_t(\x) + \varepsilon \ge \theta, \\ 
						\mathcal{L}_{\theta} \leftarrow \mathcal{L}_{\theta} \cup \{\x\} & \mbox{ if } \mu_t(\x) + 1.96 \sigma_t(\x) - \varepsilon < \theta, 
					\end{cases}
				\end{align*}
				and update the uncertainty set as $\mathcal{C}_{\theta} \leftarrow \mathcal{X} \backslash \mathcal{U}_{\theta} \cup \mathcal{L}_{\theta}$. 
\STATE{\bfseries Step~4} Select the next evaluation point by maximizing the straddle acquisition function as follows: 
				\begin{align*}
					\x_{i(t)} = \argmax_{\x \in \mathcal{X}} 1.96 \sigma_t(\x) - |\mu_t(\x) - \theta|
				\end{align*}
\STATE{\bfseries Step~5} Update the dataset as $\mathcal{D}_t \leftarrow \mathcal{D}_t \cup \{(\x_{i(t)}, y_{i(t)})\}$ 
\ENDFOR
\end{algorithmic}
\end{algorithm}

In this section, we summarize the three proposed algorithms: AL-, ATL-, and LSS-ATL-based LSE. 
Let $\mathcal{D}_t = \{(x_i, y_i)\}_{i \in \mathcal{S}_t}$ be the observed dataset at step $t$, where $S_t \subset \{1, ..., N \}$.
Our goal is to classify all candidate points $\x \in \mathcal{X}$ into the super-level set $\mathcal{U}_{\theta}$, sub-level set $\mathcal{L}_{\theta}$, or 
undetermined set $\mathcal{C}_{\theta}$. 
Note that $\mathcal{C}_{\theta}$ is necessary to properly consider the uncertainty of the problem. 
Algorithm~\ref{alg} provides the pseudocode for the AL-based LSE algorithm. 
First, we perform a GP regression on the data already observed in Step~1 to obtain the predictive mean and variance functions. 
In Steps~2 and 3, we classify the candidate points using credible intervals consisting of the predictive mean and variance functions obtained in Step~1.
Then, we maximize the straddle acquisition function in Step~4 and determine the next observation point. 
Finally, the observed data are added to the dataset, and we return to Step~1.
This process loops until all candidate points are classified into $\mathcal{U}_{\theta}$ or $\mathcal{L}_{\theta}$ or until the maximum number of iterations is reached. 

\section*{Supplementary Information 2: : Details of the Active Transfer Learning (ATL) Algorithms for LSE}

For the ATL-based LSE method, the data transformation step shown in Algorithm~\ref{alg2} is inserted before Step~1 of Algorithm~\ref{alg}. 
Let $\mathcal{D}^{\prime} = \{(\x^{\prime}_j, y^{\prime}_j)\}_{j=1}^M$ be the observed dataset of the source domain.
Diff-GP model is an algorithm that transforms the data of the source domain into the data so that it can be regarded as of the target domain.
The output of Algorithm~\ref{alg2} (i.e., the transformed dataset $\hat{\mathcal{D}}$) is aggregated with the target dataset $\mathcal{D}_t$ and used as the dataset for Algorithm~\ref{alg}. 

\begin{algorithm}[t]
\caption{Data Transformation via Diff-GP Model}
\label{alg2}
\begin{algorithmic}
\REQUIRE observed source data $\mathcal{D}^{\prime}$, GP prior, a set of candidate points $\mathcal{X}$, 
max iteration $T$
\ENSURE transformed source data $\hat{\mathcal{D}} = \{(\x^{\prime}_j, \hat{y}_j) \}_{j=1}^M$
\STATE{\bfseries Step~1} Run the GP regression with $\mathcal{D}^{\prime}$ to obtain the predictive mean function $\mu^{\prime}(\x)$ and predictive variance function $\sigma^{\prime 2}(\x)$. 
\STATE{\bfseries Step~2} Compute the predictive mean and variance of the Diff-GP model as follows: 
					\begin{align*}
						\hat{\mu}(x) &=\k(\x)^{\top}\left(\K+(\sigma^{2} + \sigma^{\prime 2}(\x)) \I\right)^{-1} (\y - \boldsymbol{\mu}^{\prime}), \\ 
						\hat{\sigma}^{2}(x) &=k(\x, \x)+\k(\x)^{\top}\left(\K+(\sigma^{2} + \sigma^{\prime 2}(\x)) \I\right)^{-1} \k(\x),
					\end{align*}
					where $\k(\x)$ and $\K$ are the kernel vector and kernel matrix defined from the target data, respectively, $\sigma^2$ is the observation error 
					of the target data, and $\boldsymbol{\mu}^{\prime} = (\mu^{\prime}(\x_1), ..., \mu^{\prime}(\x_n))$. 
\STATE{\bfseries Step~3} Compute the transformed output as $\hat{y}_j = y^{\prime}_j + \hat{\mu}(\x^{\prime}_j)$, $j = 1, ..., M$. 
\end{algorithmic}
\end{algorithm}

\section*{Supplementary Information 3: Details of the Location-Scale Shift Active Transfer Learning (LSS-ATL) Algorithms for LSE}

For the LSS-ATL-based LSE method, the data transformation step shown in Algorithm~\ref{alg3} is inserted before Step~1 of Algorithm~\ref{alg}.
Algorithm~\ref{alg3} is a combination of the Diff-GP model and the location-scale-shift transformation. 
In this study, we assume that the location-scale shift is an affine transformation with two parameters: $\gamma$ and $\eta$. 
We transform the source data by estimating $\gamma$ and $\eta$ from the observed data. 

\begin{algorithm}[t]
\caption{Data Transformation via LSS-Diff-GP Model}
\label{alg3}
\begin{algorithmic}
\REQUIRE observed source data $\mathcal{D}^{\prime}$, GP prior, a set of candidate points $\mathcal{X}$, 
max iteration $T$
\ENSURE transformed source data $\hat{\mathcal{D}} = \{(\x^{\prime}_j, \hat{y}_j) \}_{j=1}^M$
\STATE{\bfseries Step~1} Run the GP regression with $\mathcal{D}^{\prime}$ to obtain the predictive mean function $\mu^{\prime}(\x)$ and the predictive variance function $\sigma^{\prime 2}(\x)$. 
\STATE{\bfseries Step~2} Solve the following least squares problem with respect to $\gamma$ and $\eta$:
					\begin{align*}
						\gamma^*, \eta^* = \argmin_{\gamma, \eta} \sum_{i=1}^n \left(y_i - (\gamma \mu^{\prime}(\x_i) + \eta) \right)^2.
					\end{align*}
\STATE{\bfseries Step~3} Compute the predictive mean and variance of the Diff-GP model as follows: 
					\begin{align*}
						\hat{\mu}(x) &=\k(\x)^{\top}\left(\K+(\sigma^{2} + \sigma^{\prime 2}(\x)) \I\right)^{-1} (\y - (\gamma^* \boldsymbol{\mu}^{\prime} + \eta^*)), \\ 
						\hat{\sigma}^{2}(x) &=k(\x, \x)+\k(\x)^{\top}\left(\K+(\sigma^{2} + \sigma^{\prime 2}(\x)) \I\right)^{-1} \k(\x),
					\end{align*}
					where $\k(\x)$ and $\K$ are the kernel vector and kernel matrix defined from the target data, respectively, $\sigma^2$ is the observation error 
					of the target data, and $\gamma^* \boldsymbol{\mu}^{\prime} + \eta^* = (\gamma^* \mu^{\prime}(\x_1) + \eta^*, ..., \gamma^* \mu^{\prime}(\x_n) + \eta^*)$. 
\STATE{\bfseries Step~4} Compute the transformed output as $\hat{y}_j = y^{\prime}_j + \hat{\mu}(\x^{\prime}_j)$, $j = 1, ..., M$. 
\end{algorithmic}
\end{algorithm}

\end{document}